\documentclass[11pt]{article} 

\usepackage[utf8]{inputenc} 

\usepackage{geometry} 
\usepackage{booktabs} 
\usepackage{array} 
\usepackage{paralist} 
\usepackage{verbatim} 

\usepackage{fancyhdr} 
\usepackage{sectsty}
\allsectionsfont{\sffamily\mdseries\upshape} 

\usepackage[nottoc,notlof,notlot]{tocbibind} 
\usepackage[titles,subfigure]{tocloft} 

\usepackage{multirow,array}
\usepackage{float} 
\usepackage{amsmath}
\usepackage{amsfonts}
\usepackage{amsthm}
\usepackage{amssymb}

\usepackage[hidelinks]{hyperref}

\let\origref \ref
\def \ref#1{\textbf{\origref{#1}}}

\usepackage{subcaption}

\newtheoremstyle{break}
  {\topsep}{\topsep}%
  {\itshape}{}%
  {\bfseries}{}%
  {\newline}{}%
\theoremstyle{break}

\usepackage[natbibapa]{apacite}
\setcitestyle{authoryear,open={(},close={)}}

\usepackage{comment}
\usepackage{xcolor}
\definecolor{darkorange}{rgb}{0.8, 0.4, 0.0} 

\usepackage{graphicx,multirow}

\usepackage[title,toc,titletoc]{appendix}

\usepackage{tikz}
\usetikzlibrary{positioning,arrows.meta,fit,calc,shadows}

\title{A Model of Understanding in Deep Learning Systems}
\author{David Peter Wallis Freeborn \\
\small Northeastern University London}
\date{} 

\begin{document}
\maketitle

\begin{abstract}
I propose a model of systematic understanding, suitable for machine learning systems. On this account, an agent understands a property of a target system when it contains an adequate internal model that tracks real regularities, is coupled to the target by stable bridge principles, and supports reliable prediction. I argue that contemporary deep learning systems often can and do achieve such understanding. However they generally fall short of the ideal of scientific understanding: the understanding is symbolically misaligned with the target system, not explicitly reductive, and only weakly unifying. I label this the Fractured Understanding Hypothesis.
\end{abstract}

\section{Introduction}
\label{sec:introduction}

John von Neumann reportedly claimed that ``With four parameters I can fit an elephant, and with five I can make him wiggle his trunk'' \citep{Dyson2004Fermi, Mayer2010Elephant}. Today, we are in the midst of a societal and scientific revolution, driven by deep learning, a powerful curve-fitting technology. Over the last few years it has driven major advances in natural language processing \citep{vaswani2017attention, devlin2018bert, Brown2020FewShot, touvron2023llama, openai2023gpt4}, computer vision \citep{He2016resnet, Kolesnikov2021vit}, image generation \citep{ramesh2021zero, rombach2022high, saharia2022photorealistic}, multimodal learning \citep{radford2021learning, ramesh2022hierarchical,alayrac2022flamingo}, game-playing \citep{Silver2016alphago, Vinyals2019alphastar, OpenAI_dota2019}, and protein structure prediction \citep{Jumper2021alphafold}. In each case, deep learning-based architectures with millions or billions of parameters are trained on vast datasets and then adapted to downstream tasks not explicitly specified in their training objectives. We will need to understand the epistemic capabilities of these systems if we are to deploy and govern them responsibly.

Yet the epistemic capacities of deep learning systems remain unclear. Much of the debate turns on whether these models genuinely \emph{understand} their target domains, or instead produce the mere appearance of understanding through interpolation, pattern-matching, and recombination of memorized fragments. Above all, their successes are strikingly uneven, forming a ``jagged frontier'' \citep{DellAcquaEtAl2023JaggedFrontier}: the best models can outperform humans on some difficult tasks, while failing badly on others that humans find comparatively easy. The use of deep learning models invites two contradictory reactions:

\begin{enumerate}
\item \emph{How could the models possibly achieve this, unless they possess genuine understanding?}
\item \emph{How could the models possibly fail at this, if they possess any genuine understanding?}
\end{enumerate} 

Skeptics argue that structures like deep learning systems are \emph{incapable} of understanding \citep{bishop2021artificial, Pearl2018, floridi2023ai}; in the case of large language models (LLMs), they are said to be ``stochastic parrots'' or ``blockheads'' that mimic competence without grasping meaning \citep{bender2021dangers, BenderKoller2020, blockheads, marcus2018deep}. Proponents counter that the \emph{best explanation} for the impressive capabilities exhibited by frontier deep learning systems is that they exhibit at least some degree of genuine understanding. They point to the striking ability for deep learning models to solve tasks they were never explicitly trained to do \citep{LeCunBengioHinton2015deep, Brown2020FewShot, Wei2022emergence, Bommasani2021FoundationModels, buckner1, buckner2, Grzankowski2025a, grzankowski2025b}. Following \citet{Sellars1962-PSIM}, we might distinguish two \emph{images} of deep learning systems. On the scientific image, they are best described as highly elaborate curve fitters, whereas on the manifest image, they appear as agentic systems with some capacity for understanding.

This academic debate has serious real-world consequences, because high-stakes decisions increasingly rely on the outputs of deep learning models. If they generate accurate predictions \emph{only} within the narrow support of their training data, they may fail catastrophically when deployed in novel settings, a phenomenon sometimes called ``brittleness under distribution shift'' \citep{Amodei2016concrete, Schneider2020robust}. Conversely, if they do possess forms of systematic understanding, then those forms need to be articulated, and perhaps made more legible to human stakeholders. The same question bears on engineering choices between scaling and incorporating explicit causal abstractions, hybrid neuro-symbolic modules, or mechanistic interpretability constraints \citep{Amodei2016concrete, Rudin2019Stop, DoshiVelezKim2017Rigorous, Kaplan2020Scaling, LeCun2022Path, Bereska2024MechInterp, Zhang2024CausalAbstraction}, and on forecasts about more general intelligence \citep{Wei2022emergence, Bubeck2023Sparks, Chollet2019Measure, LeCun2022Path}.

Unfortunately, the dialectic has been confused, with opponents largely talking past each other. The dialectic lacks an agreed conceptual framework for understanding ``understanding''. In this paper I propose a model of what I call \textbf{systematic understanding}, an explication of one kind of understanding that I hope is rigorous enough to give some foundation for the debate, while also being general and flexible enough to apply to machine learning systems, as well as humans and other kinds of epistemic agents. This notion of understanding is deliberately non-anthropocentric, so that it could, at least in principle, apply to non-human, mechanical systems.\footnote{Even for those who would prefer a different notion of understanding, hopefully providing this model will offer a starting point for alternatives to be explicated.} Then I assess whether deep learning systems in fact \emph{do} exhibit systematic understanding, of at least some salient properties, for some real-world systems.

Roughly speaking, for an agent to possess systematic understanding of some property of a target domain, they must have an internal model of the target system, capturing some of its salient regularities, and enabling them to make adequate predictions about this property systematically from that underlying model. That is, the predictions do not arise from luck, memorization, or blind interpolation, but rather because they have internalized some \emph{real patterns} in the target system. Following \citet{Dennett1991}, I define a pattern in a body of data as \emph{real} when there exists a finite, algorithmic description that (i) compresses the data more efficiently than brute‐force enumeration and (ii) supports predictions of the existing or future data with a reliability that materially exceeds chance. Crucially, understanding on this account comes in degrees. At one extreme, a system that merely memorizes its training data achieves no genuine compression of the target's regularities and thus possesses no systematic understanding. At the other, a system whose internal model compactly captures the target's dependency structure and supports successful predictions across a wide range of conditions possesses a high degree of understanding. Most real systems fall somewhere between these poles. I will use this notion of systematic understanding to argue that deep learning systems possess a real, but \textbf{fractured understanding}, in a way that should make sense of the two contradictory intuitions above.

I have two distinct aims, one primarily conceptual and one primarily diagnostic. The conceptual aim is to propose a deliberately \emph{thin}, non-anthropocentric model of \emph{systematic understanding}, intended as a basic framework for clarifying disputes about whether deep learning systems understand their target domains. The account is offered as a proposed explication: it introduces a regimented notion of understanding that can be applied to artificial systems, and that helps to locate disagreements more precisely, even for readers who ultimately prefer a different notion. The goal is to supply a tractable framework for assessing which epistemic capacities are present in current systems, what kinds of failures those capacities exhibit, and which engineering choices predictably shape the resulting form of understanding.\footnote{I do not claim to analyze the ordinary-language concept of understanding, nor to identify an underlying natural kind. Nor do I claim that satisfying the present criteria is sufficient for the strongest kinds of scientific understanding, especially those associated with causal-mechanistic explanation, although I hope it might provide a basis that could be adapted for such accounts.}

The diagnostic aim is to use that framework to characterize a recurrent pattern in contemporary deep learning. I argue that current systems can and often do satisfy the thin criteria for systematic understanding in restricted respects, yet that the form such understanding typically takes is not well captured by the traditional ideal of scientific understanding. In particular, learned representations are often misaligned with the target domain's natural variables, and competence is frequently distributed across locally reliable fragments rather than unified into a small stock of reusable principles. This motivates the \emph{Fractured Understanding Hypothesis} developed later.

Note that I will focus on neural network systems more generally, rather than natural language processing or large language models (LLMs) specifically. The reason is that applying this approach of systematic understanding to LLMs would require a further, specialized treatment. The framework here suggests that applying systematic understanding to LLMs requires answering two distinct questions in sequence: first, whether they acquire a systematic understanding of human language, and second, whether that linguistic understanding endows them with a wider understanding of the ``world'' (including a \textbf{world model}). These are genuinely different questions requiring different kinds of evidence and different bridge principles, and conflating them has been a source of confusion in the existing debate. By working through cases where the target system is well-defined and the bridge principles are (relatively) transparent, in geometry, algebra, game dynamics, I aim to secure the framework on firm ground before it is extended to the harder case. With that said, transformer models will provide two of the key case examples, in one case with highly structured linguistic data.

In section~\ref{sec:definition}, I offer a brief definition of what I mean by a deep learning system. In section~\ref{sec:muddles}, I give an overview of the existing dialectic around understanding in deep learning. In section~\ref{sec:structural-understanding}, I explicate a model of understanding, which I call \textbf{systematic understanding}. In section~\ref{sec:modelingmodeling}, I present a model of how deep learning systems model the world, with the aim of explaining the senses in which deep learning systems can or cannot achieve systematic understanding. I supplement this with four examples in section~\ref{sec:examples}, each designed to pump different intuitions about the kinds of systematic understanding that current deep learning systems are well suited for developing. In section~\ref{sec:fractured}, I propose a \textbf{fractured understanding hypothesis}, according to which deep learning systems often achieve systematic understanding, but that this understanding is highly fragmented and rarely completely general. I conclude (section~\ref{sec:conclusions}) with some lessons for both philosophers and artificial intelligence developers. 

All computational models and experiments discussed in this paper are implemented in a set of iPython Jupyter notebooks, available at: \url{https://github.com/DavidFreeborn/Model_of_Understanding_Deep_Learning}. The repository is intended as a transparent companion to the paper rather than as a general-purpose software library.

\section{Deep Learning Systems}
\label{sec:definition}

\textbf{Machine learning} systems are algorithms that improve their performance on a task by adapting to accumulated information, rather than through explicit programming of every rule. Through some specified learning technique, the systems are exposed to data, often in the form of input–output pairs, and inductively infer a mapping from inputs to outputs that can then be applied to new, unseen cases. At its most basic level, we can think of the task of a machine learning system as learning to approximate some real-world function, as an internal function $f_\theta$, by altering various \emph{parameter values}, $\theta$. Together, the function encodes all of the system's information about the target domain.

Typically, we divide the data into \textbf{training data} (examples that the learner actually \emph{sees} during the parameter‐adjustment phase) and \textbf{test data} (a previously unseen set of examples). Given training data ${(x^i,y^i)}_{i=1}^n$, where each $x^i \in \mathcal{X}$ is an input (often a vector in $\mathbb{R}^d$, for some dimensionality $d$) and each $y^i \in \mathcal{Y}$ is a desired output (a scalar, vector, or label), a machine learning system learns a function 

\begin{equation}
    f_\theta: \mathcal{X} \rightarrow \mathcal{Y}.
    \label{eq:deep_learning}
\end{equation}

\noindent The aim is for $f$ to correctly predict the outputs for new inputs (i.e. test data), suggesting that the machine is not merely memorizing the training pairs but has internalized genuine regularities \citep{Hastie2009, goodfellow2016deep, LeCunBengioHinton2015deep}.

\textbf{Deep learning} refers to a family of machine learning methods in which the model’s parameters are organized across many ($N$) layers, typically in a structure called a multilayer \textbf{neural network}. We call the layers between the input layer, which receives the data, and the output layer, which produces predictions, the \textbf{hidden layers}. In a feed-forward neural network\footnote{For simplicity, henceforth, I will discuss feed-forward neural networks, in which information flows strictly in one direction from input to output without recurrence or cycles. These are the most common kind, but general arguments of this paper apply more broadly.}, each layer $j$ transforms its input through a composition of an affine transformation $W_j (x_{j-1}) + b_j$ and a nonlinear activation function, $\sigma_j$, where $W_j, b_j \in \theta$ and $x_j$ refers to the inputs from layer $j$. So we can express the whole function as,

\begin{equation}
f_\theta(x_0) =
\sigma_N\big(
W_N\,\sigma_{N-1}\big(
\cdots
\sigma_1(W_1 x_0 + b_1)
\cdots
\big) + b_N
\big),
\label{eq:neural-net}
\end{equation}

\noindent where $x_0$ is the initial input data. The combination of depth (many layers) and scale (large numbers of parameters) allows these methods to approximate extremely complex functions. 

For example, take a neural network trained to classify images as depicting a cat or not. Each input $x_i$ might be a vectorized image (say, a $28 \times 28$ pixel image flattened into a 784-dimensional vector), and $y_i$ are the corresponding labels, either \emph{a cat} or \emph{not a cat}. The neural network learns a function that assigns a probability distribution over labels to each image.  We can think of the resulting function as specifying a \textbf{decision boundary} (e.g. $f_\theta = 0$), a surface defined over the input space, classifying what it thinks is or is not an image of a cat. More loosely, perhaps we might think of the function as giving us the machine learning system's model of \emph{catness}.  

What does this learned surface look like? For simplicity, let us suppose that the activation functions, $\sigma_i$, take the most common form, that of a \textbf{Rectified Linear Unit (ReLU)}\footnote{This greatly simplifies the arguments. However note that the general following arguments about spline interpolation will apply for any piecewise polynomial activation functions.}, defined as,

\begin{equation}
    \sigma_i (z) = \max(0,z).
\end{equation}

This function is continuous and piecewise affine: it is linear on each side of the origin but introduces a kink (a point of non-differentiability) at zero. The effect of combining these piecewise affine activations in each layer is to partition the input space $\mathbb{R}^d$ into convex polytope regions, within each of which the function behaves as an affine transformation,

\begin{equation}
    f_\theta (x) = A_k x + b_k \text{ for } x \in \mathcal{P}_k,
\end{equation}

\noindent where $\mathcal{P}_k$ is a polytope in the partition induced by the activation patterns of the ReLU units, and $A_k$, $b_k$ are matrices and vectors determined by the weights $\theta$. This piecewise-affine structure is a form of multivariate spline, and so this view is sometimes known as the \textbf{spline theory of neural networks} \citep{balestriero2018spline}.\footnote{The term ``spline'' originally referred to a flexible strip of wood, metal, or plastic used to draw smooth curves through a set of points. Mathematically, a spline is defined as a piecewise-polynomial function that is smooth at the boundaries where pieces connect. But here ``spline'' is meant in a looser sense; for ReLU networks, the resulting map is \emph{continuous piecewise-linear}, not a smooth function \citep{balestriero2018spline,raghu2017expressive}.}

\begin{figure}[ht]
\centering
\centering
\begin{subfigure}{0.48\textwidth}
  \centering
  \includegraphics[trim={7cm 5cm 8cm 7cm}, clip, width=\linewidth]{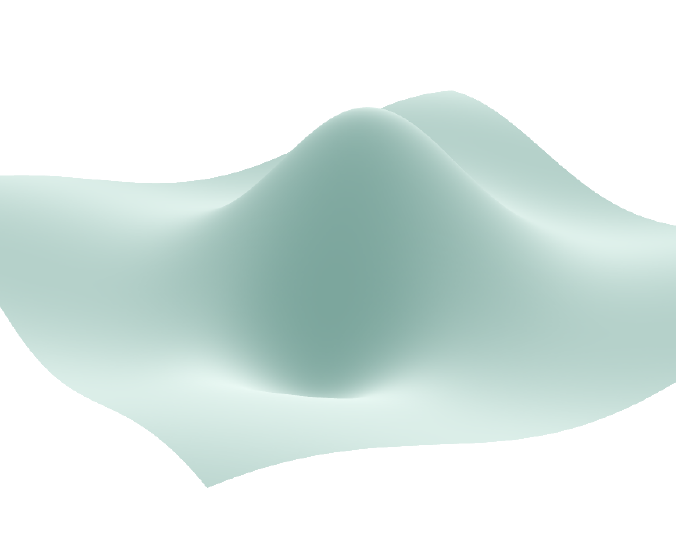}
  \caption{Ground truth surface}
\end{subfigure}
\hfill
\begin{subfigure}{0.48\textwidth}
  \centering
  \includegraphics[trim={7cm 5cm 8cm 7cm}, clip, width=\linewidth]{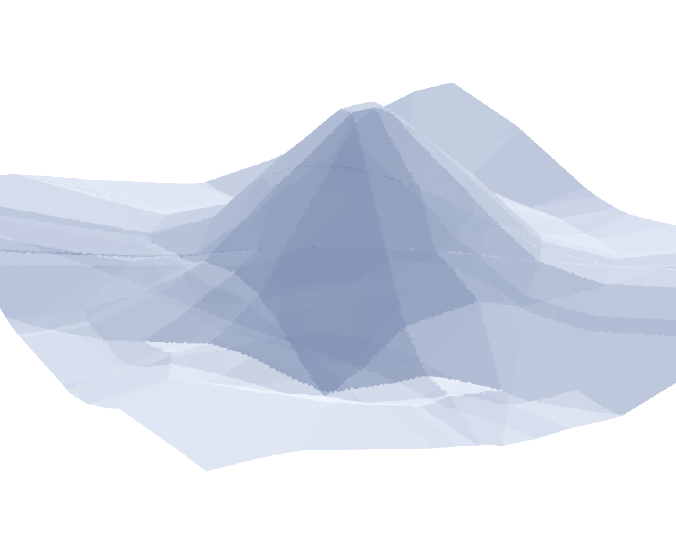}
  \caption{Learned isosurface}
\end{subfigure}
\caption{Comparison between a ground truth surface (left) and the neural network’s learned isosurface (right). Observe how the learned isosurface is made by stitching together piecewise-linear surfaces. The ReLU activation functions lead the network's output to be a continuous piecewise-linear function (a multivariate spline), approximating the smooth target curvature through a vast number of local planar regions.}
\label{fig:surface-learned}
\end{figure}

Thus, the learned surface of such a neural network consists of a complex arrangement of locally linear regions, stitched together into a continuous, but non-differentiable function. The network learns by adjusting how these linear patches are assembled to best approximate the training data. Figure \ref{fig:surface-learned} provides one visual example.\footnote{For this, I trained a multi-layer perceptron (MLP) neural network consisting of three linear layers with 24 neurons in each hidden layer ($3 \to 24 \to 24 \to 1$) and ReLU activation functions, totaling 697 trainable parameters. The network is trained to regress an implicit scalar field of the form $F(x,y,z) = z - f(x,y)$, where $f(x,y) = 0.5 \cos x \cos y + 2.0 \exp\!\left(-\frac{x^2 + y^2}{1.5}\right) - \exp\!\left(-\frac{(x-1.5)^2 + (y-1.5)^2}{0.5}\right)$, which defines a smooth height-field landscape with a central peak, oscillatory structure, and an off-center crater. The training data consist of 262,144 points sampled from a regular $64^3$ grid over the domain $[-3.5, 3.5]^3$, with target values given by the analytic implicit function. The model is optimized using Adam with learning rate $10^{-3}$ for 4{,}000 epochs, minimizing the Mean Squared Error between predicted and ground-truth scalar values. After training, the learned implicit surface $\hat{\Sigma} = \{(x,y,z) \mid f_\theta(x,y,z) = 0\}$ is extracted using the Marching Cubes \citep{lorensen1987marching} algorithm on a higher-resolution $200^3$ evaluation grid. See the accompanying repository for the implementation details.} So taken literally, neural networks really are sophisticated curve fitters; all the information that the neural network learns about the world can be understood as merely parameterizing this spline-like surface.

\section{Proxy Battles over Understanding}
\label{sec:muddles}

The debate over whether deep learning systems truly \emph{understand} has been intense, and often confused. One source of confusion is that ``understanding'' itself is rarely defined with enough precision to support productive disagreement. In practice, machine learning researchers often sidestep the term, treating it as too vague. Yet concerns about understanding continue to motivate much of the surrounding discourse. What has happened, I suggest, is that these concerns have often been displaced onto more tractable proxy concepts that capture only part of what is at stake. Two of the most important proxy battles concern whether machine learning systems genuinely generalize or merely memorize their training data, and whether they extrapolate beyond their training regime or merely interpolate within it. Both debates are important in their own right, but both can also be read as attempts to operationalize specific aspects of the deeper question about understanding.  However, as we shall see, these proxy debates have also been confused by a lack of common, clearly understood terminology. Making the connections explicit will help motivate the framework developed in the next section.

\subsection{Memorization and Generalization}
\label{sec:memorgeneral}

If a machine learning system genuinely \emph{understands} the target, by internalizing real patterns or regularities in the data, then we should expect it to make good predictions about new, relevantly similar data. This ability to extract underlying regularities from the training data and to apply them in order to make accurate predictions on previously unseen data is called \textbf{generalization}. On the other hand, \textbf{memorization} refers to the rote storage of specific training data. Memorization is often characterized by \emph{overfitting}, in which a system learns noise in the training data, resulting in poor performance when predicting on new, unseen data \citep{Vapnik1998Statistical, Hastie2009}. The operational definitions vary significantly \citep{wei2024memorizationdeeplearningsurvey}; however, the debate over whether a model genuinely generalizes, rather than merely memorizing, is often best read as a proxy battle about whether machine learning systems can acquire some level of genuine understanding.

Early theoretical arguments from statistical learning theory suggested that memorization might contribute significantly to the success of deep learning systems \citep{Arpit2017closer, Feldman2020DoesLearning, Zhang2017understanding}. Roughly, if a deep learning system has more free parameters than there are distinct training examples, then it is theoretically possible for the model to perfectly memorize each example by assigning each one a distinct degree of freedom. More precisely, \textbf{model capacity} can be thought of as the flexibility for a machine learning system to represent a wide variety of patterns.\footnote{This can be understood through Vapnik–Chervonenkis (VC) dimension or Rademacher complexity. The \emph{VC–dimension} of a model class is the size of the largest set of points it can shatter—that is, the largest dataset on which it can realize all possible labelings \citep{Vapnik1998Statistical}. Rademacher complexity measures the expected correlation between random noise and the model’s outputs, thereby quantifying the model’s ability to fit arbitrary labels \citep{Bartlett2002Rademacher}.} Low‐capacity (high‐bias) models generalize poorly because they cannot capture the target system's full complexity; high‐capacity (high‐variance) models risk overfitting by tailoring themselves too closely to the finite and noisy data. 

\begin{figure}[pht]
  \centering
  \begin{subfigure}[b]{0.81\textwidth}
    \centering
    \includegraphics[width=\textwidth]{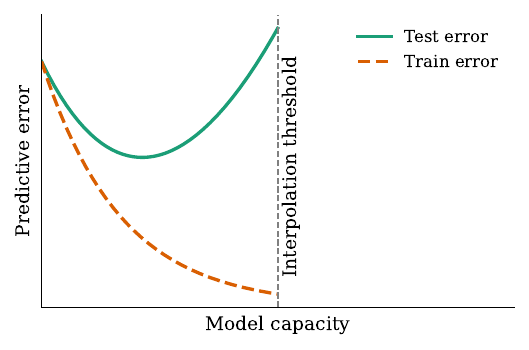}
    \caption{The bias–variance tradeoff: as deep learning system capacity increases from left to right, training error falls, but test error increases, indicating the system is overfitting due to memorizing more of the data.}
    \label{fig:biasvariance}
  \end{subfigure}

  \vspace{1em}

  \begin{subfigure}[b]{0.81\textwidth}
    \centering
    \includegraphics[width=\textwidth]{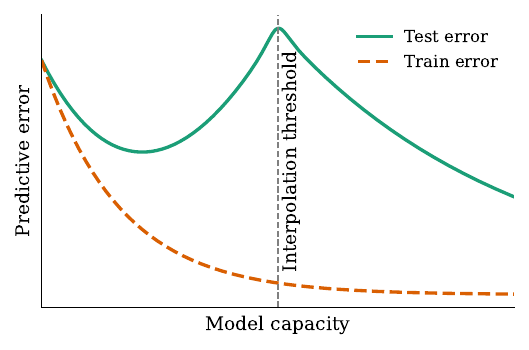}
    \caption{The double descent phenomenon: beyond the interpolation threshold, the test error continues to decrease, indicating the system is generalizing by learning real patterns in the training data.}
    \label{fig:doubledescent}
  \end{subfigure}

  \caption{Schematics showing the bias–variance tradeoff and the double descent behavior.}
  \label{fig:combined}
\end{figure}

Counting arguments suggested that memorization and generalization are in tension: not every training example can be memorized uniquely without harming the system’s ability to extract general patterns \citep{Hestness2017Deep, achille2017emergence, achille2020informationdeepneuralnetwork}. More generally, theoretical arguments predicted that once model capacity exceeds a certain threshold relative to the dataset size, memorization should dominate over generalization. When this happens, a model's predictive performance should deteriorate  when the model is deployed outside the memorized training data. This phenomenon is sometimes described as the bias-variance tradeoff, schematized in figure~\ref{fig:biasvariance}.

However, landmark experiments shattered these expectations. \citet{Zhang2017understanding} demonstrated that modern neural networks with millions of parameters can fit \emph{random} labels perfectly and yet still generalize well on \emph{real} labels, provided the data are structured. In other words, the same architecture that can \emph{memorize} unstructured noise to zero training error simultaneously \emph{generalizes} impressively on structured data. Further work revealed a more complex relationship between generalization and memorization. Rather than following a simple U‐shaped curve in figure~\ref{fig:biasvariance}, modern deep networks often exhibit a \textbf{double descent} phenomenon shown in figure~\ref{fig:doubledescent} \citep{Belkin2019reconciling,Nakkiran2020double, Rocks_Mehta_2022}. As the system's capacity grows, test error first decreases, then increases around the interpolation threshold (where the model first attains zero training error), and then decreases again. The second descent suggests that supposedly over‐parameterized models can re‐enter a generalization‐friendly regime, perhaps because the standard methods of optimization bias them towards seeking out simpler, more stable solutions. A distinct but related phenomenon is \emph{``grokking''}, in which a machine learning system undergoing continued training long after it achieves zero training error suddenly seems to acquire improved performance on test data \citep{Power2022grokking}. Generally, this is interpreted as a sudden shift from memorization to generalization. Once again, grokking is hypothesized to arise because contemporary learning algorithms are biased to favor simpler solutions \citep{Arpit2017closer,VallePerez2018deep,Liu2022towards}.

Another strand of evidence comes from attacks that extract verbatim training snippets from LLMs \citep{Carlini2021Extracting}. These studies demonstrate that, while models do indeed memorize \emph{some} verbatim snippets, this is vanishingly rare relative to their overall behavior. Contrary to early theoretical expectations, it seems that pockets of memorization can and do coexist with broad generalization. 

Thus, while debate continues \citep{recht2019imagenet, anagnostidis2022curious}, a general consensus has emerged. In current, successful deep learning models, only a small (but non-zero) fraction of model capacity is devoted to direct memorization, while many of the achievements arise due to generalization. At the very least, deep learning systems are able to identify and apply general patterns in the training data, an indication that these systems might possibly \emph{understand} some of the regularities in their target systems. We will explore memorization more precisely in section~\ref{sec:memorization}.

\subsection{Interpolation and Extrapolation}
\label{sec:interpextrap}

A second, closely related, proxy battle centers on whether these systems genuinely \textbf{extrapolate} or merely \textbf{interpolate}. Roughly speaking, a system interpolates when it generates successful, novel predictions from inputs within the envelope of its training data, whereas it extrapolates when it makes successful, novel predictions from inputs outside that envelope. A system that only memorizes the training data may have limited success in interpolating between those data points, but is unlikely to have \emph{systematic} success outside the envelope of the training regime. A system that generalizes should have at least some interpolative success. It \emph{may} succeed at extrapolation, \emph{if} the learned generalizations extend beyond the training regime. The appeal of extrapolation as a diagnostic comes from the thought that a system which has acquired deeper understanding should sometimes succeed beyond the immediate envelope of its training data, whereas a system that has learned only local fits will often fail outside of the training regime.

A prevailing folk-theory holds that deep learning systems are generally good at interpolation but poor at extrapolation. After all, deep learning systems work by learning to fit a function to the data only in the specific domain of their training data. As such, deep learning systems  can estimate intermediate values between known examples, but fail to pick up patterns that generalize outside of the training data. \citet{marcus2018deep} gives a simple example: he trained a three-layer model on the (seemingly trivial) identity function, $f(x) = x$ for a small set of even numbers. The model worked successfully on nearby even numbers but gave nonsensical results for odd numbers. Marcus takes these results to show that the model successfully interpolates to nearby even numbers, but fails at odd numbers, outside of the original training set.

\citet{balestriero2021learning} challenge the conventional wisdom, defining interpolation as predictions whose input lies inside the convex hull of the training points, whereas extrapolation applies whenever the input falls outside that. They demonstrate that in high ($\gtrsim 100$) dimensional spaces, characteristic of almost all real-world deep learning problems, the probability that any new input will fall inside the convex hull becomes very small. Consequently, they argue that almost all practical high-dimensional deep learning predictions are extrapolative.

However, \citet{Chollet2019Measure} suggests that the relevant criterion of \emph{interpolation} in many cases should be \emph{interpolation on the latent manifold}. The \textbf{manifold hypothesis} posits that many kinds of natural data occupy lower-dimensional structured subspaces (latent manifolds) rather than filling the surrounding volume.\footnote{See \citet{freeborn2025effective} for a philosophical overview of the manifold hypothesis.} Deep learning networks learn to unfold this manifold, turning it into a space in which local linear interpolation can suffice for generalization. As a result, most of the cases of extrapolation inside the convex hull can also be understood as interpolation on the manifold.

Both accounts are probably right: most deep learning predictions are both close to being interpolative on the manifold and extrapolative in the convex hull. It is likely that deep learning systems are poor extrapolators beyond the latent manifold of the data. They \emph{can} learn general patterns in the data. However, often, deep learning systems fail to pick up on the right kinds of inductive regularities: those that they identify often fail to extend far beyond the envelope of the training data. 

The two foregoing proxy disputes are best understood as partial attempts to operationalize constraints on genuine understanding. They arise because researchers often want to ask whether a model has learned real structure, but lack a shared and sufficiently precise account of what that would amount to. The memorization/generalization debate targets the thought that understanding must not depend essentially on sample-contingent idiosyncrasies of a particular training set. The interpolation/extrapolation debate targets the thought that understanding should involve some grip on structure that extends beyond local fit within the training regime. The model of systematic understanding developed in section~\ref{sec:structural-understanding} is intended to make these underlying concerns explicit and to show why these proxy debates capture something important, while also explaining why they are insufficient on their own.

\section{A Model of Understanding}
\label{sec:structural-understanding}

\emph{Understanding} goes beyond mere capacity to predict. We shall think of it as a relation between one kind of system (the \textbf{agent}) and certain \textbf{properties} of a second system (the \textbf{target system}). This is the intuition I want to capture: to say that an agent \emph{understands} some property of a system means that they have the capacity to make predictions about it that are not merely due to luck, and not due to rote memorization. Instead, they can systematically derive such predictions from an underlying model of the target system. Insofar as their understanding is good, then this model must track salient features of the target system, and the predictions should derive from the accurate tracking of these features.

This conception is intended to capture a single, reasonably coherent notion of understanding. Hopefully, the core intuition is familiar.  A human child who can reliably classify previously unseen four-legged pets as either dogs or cats plausibly understands something about ``dogness'' and ``catness'': they have internalized regularities that support correct generalization, even if they cannot articulate them. Likewise, an athlete who can consistently catch a ball has, in effect, internalized a model of its typical trajectories, and understands something about the dynamics of this physical system (even if they lack training in physics). And a scientist who constructs an explicit model of a system and uses it to generate novel predictions and explanations likewise exhibits understanding of the system in question (even if some parts of the model are false).

Understanding in this sense comes in degrees, and an agent may understand some properties of a target system better than others. Consider the case of planetary orbits as an example target system. These orbits have many properties; for instance their shape is not circular, but rather slightly anisotropic. In the early seventeenth century, Tycho Brahe’s meticulous observations provided accurate positional data for Mars and Venus. These data supported reliable short-term prediction: Brahe could list where Mars or Venus had been, and predict where they would be by interpolation, but he lacked any principled grasp of the underlying patterns behind the numbers. So, he already possessed some understanding, but it was very limited. Johannes Kepler went a step further. By recognizing that the Sun sits at a focus of an ellipse, Kepler could derive the planetary data from a simple geometrical model. His capacity to forecast future positions was therefore no longer a matter of lucky curve-fitting but followed systematically from the structure of the model itself  \citep{Thoren1990LordOfUraniborg,Kepler1992NewAstronomy, Voelkel2001CompositionAstronomiaNova}.
I contend that Kepler had a greater \emph{understanding} of the target system: he could predict the motion of the planets, deriving his predictions from an underlying elliptical model and three laws of motion.

Yet, Kepler's models could not explain all of the system's phenomena; the elliptical form of the orbits was treated as a basic geometrical fact, not itself explained. Newton achieved a deeper understanding by showing that elliptical orbits follow from an inverse-square force law. Yet, even Newton could not fully understand why the attraction took an inverse-square form. Gauss, in turn, developed a still deeper understanding using a further underlying model: a field emanating uniformly in all directions from a point source in three-dimensional space may naturally exhibit such an inverse-square dependence. So while Kepler had \emph{some} understanding of some properties of the target system, Newton understood \emph{more}, and Gauss understood even \emph{more}. At each stage, understanding deepened by embedding previously accepted regularities within a more encompassing model. Of course, none of these models captured all features of the target system. They idealized away tidal effects, relativistic corrections, and many other phenomena. Models of a target system are generally highly idealized, providing only an approximate understanding of some subset of properties, under some conditions, based on an incomplete representation of the system.

This account of understanding is intended to capture some core insights of an ability-based picture, towards which much of the recent philosophical literature has converged \citep{Woodward2003, Strevens2008, Salmon1984, Railton1978, Grimm2014, Grimm2017, Greco2014, deRegt2017}, especially the manipulationist or interventionist approaches, as well as the contention  that genuine understanding must be free of environmental luck \citep{Pritchard2010}.\footnote{However, see \citet{Kvanvig2009} for an alternative view.} With that said, my aim is to explicate a deliberately limited and philosophically \emph{thin} conception of understanding; for example, I will not consider causal mechanisms here. I make no claim that this is a general explication of the concept, nor that it is the best overall account. Rather, the purpose is to provide one useful starting point, to establish some common ground on which to assess and evaluate the kinds of understanding that deep learning system can, or cannot, exhibit.\footnote{To put it another way, I do not aim to establish a metaphysical thesis about ``understanding''. I wish to propose a model of ``understanding'' as a foundation for the debate, and thereby to endow the reader with some understanding of this concept.}

But I aim for an account of understanding that is suitable, at least in principle, for artificial intelligences. Therefore, I will avoid anthropocentric notions of understanding here, not because they are unhelpful, but they are, at least prima facie, poorly suited to machine learning systems. Thus my account will not aim to capture all of the key insights from explicitly psychological accounts of understanding \citep{deRegt2017}. After all, deep learning systems ostensibly lack precisely those psychological features.\footnote{With that said, the emphasis on an agent’s capacity to manipulate the model echoes many features de Regt’s contextual theory of scientific understanding. Likewise, this approach necessitates failing to capture some features of testimonial accounts \citep{Hills2016}.}

\subsection{Memorization as a Limiting Case}
\label{sec:memorization}

At this point, it is worth explicating the term ``memorization'' more clearly, and to explain why it constitutes a limiting case that falls short of systematic understanding.

Following \citet{Dennett1991}, a pattern in a body of data is \textbf{real} only if it admits a description that is shorter than brute-force enumeration, thereby supporting reliable prediction. The guiding contrast is between a lookup table, which guarantees accuracy by recording every idiosyncratic detail, and a compressed description that exploits stable regularities. Real patterns, in this sense, are precisely those that can be compressed while offering predictive power \citep{Dennett1991, Chaitin1975, LiVitanyi2008}.

Let $T$ be a target system, and let $p$ be a property of $T$ that an agent system $S$ is tasked with predicting. In standard learning settings, $S$ is trained on a finite dataset $D=\{(x^i,y^i)\}_{i=1}^n$ generated from $T$ under fixed bridge principles. At a first approximation, a memorizing system is one whose apparent success is essentially \emph{data-dependent}. Its performance relies on idiosyncratic features of the particular sample $D$ that happened to be encountered, rather than on extracting regularities that would support similar success across other datasets generated from the same target system.

This distinction can be made precise using the Minimum Description Length (MDL) framework. On an MDL approach, a learner extracts structure from $D$ by selecting a hypothesis $H$ that minimizes the total description length
\begin{equation}
    L(D,H) = L(H) + L(D\mid H),
\end{equation}
\noindent where $L(H)$ measures the complexity of the hypothesis and $L(D\mid H)$ measures the remaining description length of the data given that hypothesis \citep{Rissanen1978, Grunwald2007MDL}. A hypothesis offers genuine compression only if it improves upon brute listing:
\begin{equation}
    L(H) + L(D\mid H) < L(D).
\end{equation}

Here ``length'' refers to the length of an optimal code (e.g. in bits) for representing an object (see \citealp{Grunwald2007MDL, Rissanen1978, Chaitin1975}). A hypothesis with many free parameters, exceptions, or ad hoc clauses requires a longer description than a simple, structured hypothesis. Likewise, if a hypothesis captures genuine regularities in the data, fewer additional bits are needed to specify the remaining details of the dataset once the hypothesis is given. Thus MDL offers a formalization of the idea that learning consists in trading a modest increase in model complexity for a substantial reduction in the cost of describing the data.

Compression, however, is necessary but not sufficient for systematic understanding. What matters is not compression \textit{simpliciter}, but compression that captures stable regularities of the target system across the relevant perturbation family, rather than merely exploiting statistical artifacts of a particular training sample. To identify a real pattern, in Dennett's sense, is to distinguish \textbf{signal} from \textbf{noise}: to retain those regularities that persist across datasets generated from the same target system, while discarding sample-contingent idiosyncrasies that do not support reliable prediction under appropriate changes. \footnote{ This qualification is important because a system can achieve compression of the wrong kind. Shortcut learning, for example, compresses the training data by latching onto superficially predictive features, such as image texture or background context, that happen to be correlated with the target property in the training distribution but do not reflect the deeper structure of the target system. Such a system compresses, but it compresses noise rather than signal.} A memorizing system fails to make this distinction. Because it incorporates noise specific to $D$ into its representation, its success is fragile, not robust. When evaluated on a new dataset $D'$ drawn from the same target system, where the underlying signal is unchanged but stochastic details may differ, performance typically degrades. Overfitting provides a familiar illustration of this phenomenon.

In this sense, memorizing agent systems are those that build a trivial model of the target system. Model complexity scales with the amount of data rather than with the complexity of the regularities governing $p$. Memorization is the limiting case in which predictive success is secured primarily by retaining training-specific information, rather than by internalizing a model that tracks target-stable structure. In information-theoretic terms, memorization is the limit of zero genuine compression: the model's effective description length scales with the size of the dataset rather than with the complexity of the regularities governing $p$. Such a system may perform well within a narrow regime of familiar cases, but it does not thereby attain systematic understanding of the target property.

None of this implies that memorization is useless. Lookup tables and memory-based strategies can be valuable for reliable prediction in constrained domains. The point is rather that such success is compatible with a lack of systematic understanding. Moreover, real systems often occupy intermediate positions. A single agent may combine pockets of memorization with real understanding. This hybrid picture will be central to the diagnosis of \emph{fractured understanding} in later sections.

Of course, it is worth noting that compression itself admits of degrees and kinds. A system may achieve modest compression by encoding 
a few local regularities more efficiently than brute enumeration, while still falling far short of the deepest forms of compression, 
in which a small stock of reusable principles generates predictions across a wide range of phenomena. The distinction between these 
levels of compression will prove important when we consider the ideal of scientific understanding in section~\ref{sec:ideal}, 
and the characteristic ways in which deep learning systems fall short of it.

\subsection{Systematic Understanding}
\label{sec:systematic}

We are now in a position to state the core idea more precisely. I propose the following characterization of \textbf{systematic understanding}.

\begin{quote}
    An \textbf{agent system} $S$ \textbf{systematically understands} a property $p$ of a \textbf{target system} $T$ insofar as the following criteria are met.
    \begin{itemize}
         \item $S$ contains a subsystem $M \subset S$ that functions as an \emph{adequate model} of some aspect of $T$.
        \item The model $M$ systematically tracks the property $p$ of $T$ without memorizing it.
        \item There exist appropriate \textbf{bridge principles} connecting the terms of $M$ to those of $T$.
        \item The agent can use $M$ to (approximately) derive certain properties of $p$ of $T$.
    \end{itemize}
\end{quote}

\noindent Several of these terms require further explanation.

\paragraph{Structured Systems}
In practice, this notion of systematic understanding is fundamentally \emph{structural}. What matters is that the internal model instantiated by the agent mirrors, preserves, or tracks relevant patterns, relations, or invariants in the target system, and that the agent can exploit this structure to generate predictions. 

To make this more precise, I will treat both agent systems and target systems as structured systems. I leave open how best to represent such structures, whether in set-theoretic, algebraic, category-theoretic, or other terms.
\footnote{There is a parallel here with the free energy principle (FEP) \citep{Friston2010FEP}. On the FEP picture, an \emph{adaptive} system $S$ is a persisting self-organizing (stochastic) dynamical system that maintains its organization over time, typically by remaining within a bounded range of states in a relatively stable environment. Such a system can be described as maintaining internal states (often interpreted as encoding approximate posterior beliefs) that parameterize a generative model of how its sensory states depend on latent variables in an external system $T$. In our terms, this generative model functions as an internal model $\mathcal{M}$ whose structure approximately mirrors the dependency relations linking (some variables of) $T$ to the system’s observations. Under a Markov-blanket partition (a set of \emph{sensory} and \emph{active} interface variables such that internal and external states are conditionally independent given the blanket), internal and external states are coupled only via those interface states \citep{Friston2010FEP,Friston2013Life}. This partition plays a role akin to a family of bridge principles in our sense, since it fixes the interface variables through which $S$ and $T$ are coupled, and thereby the channel through which internal states can track external regularities. \emph{Active inference} is the associated account of perception and action: roughly, $S$ updates its internal states so that its sensory inputs become less surprising under its generative model (where ``surprise'' means the negative log probability of current sensations), and it selects actions expected to yield unsurprising, preferred, and uncertainty-reducing future sensory inputs \citep{FristonEtAl2017Process,ParrPezzuloFriston2022}. On this construal, paradigmatic active-inference cases are naturally interpreted as instances of \emph{structural} understanding (though not automatically as \emph{reductive} understanding).}

This is \emph{not} intended as an ontological commitment, that real-world systems admit structural representations.\footnote{Some (see \citealt{LadymanMustGo}) do contend that the world really does consist of structured systems, in which case, one could choose interpret these claims more literally.}  All I offer is a \emph{model} of understanding, to give us a way to study the concept more precisely. Representing systems as structured entities is intended as a convenient modeling choice, so that we can talk about how an agent's model system succeeds or fails in tracking properties of the target system.  Nonetheless, it is standard to describe real-world systems using structured systems (e.g. scientific theories). In the same spirit, I will treat both agent systems and target systems \emph{as if} they were structured systems, so that relations such as tracking, derivation, and misrepresentation can be meaningfully discussed.

Different kinds of tracking relationships could bestow different kinds of systematic understanding. In this paper, I will focus on \textbf{structural understanding}. However, it is also worth considering other kinds; I suggest one other, \textbf{reductive understanding}. Loosely speaking, structural understanding requires that the model represents the right patterns and relations in the target system, whereas reductive understanding requires that the model represents a higher level approximation of the target system.\footnote{We could also consider stronger kinds of tracking relation, which I will not discuss here for the sake of brevity. An obvious, and potentially highly illuminating extension would be a kind of \textbf{causal understanding}, with a requirement of a causal tracking relation. Plausibly, a causal tracking relation would require invariance under interventions: for appropriate variables $X$ and target property $p$, the model tracks $p$ causally only if it represents (perhaps implicitly) interventional dependencies of the form $P(p \mid do(X{=}x))$, not merely observational correlations $P(p \mid X{=}x)$, as captured by Pearl’s interventionist framework and do-calculus \citep{Pearl2018}.} I will discuss these specific kinds of systematic understanding in sections \ref{sec:struc} and \ref{sec:reduc} below.

\paragraph{Agent Systems}
The notion of an agent system is correspondingly broad. The agent may be a human, an artificial intelligence, a collective organization, an idealized Bayesian learner, or something else. In some cases, the agent system may even be embedded within the target system that it seeks to understand. What matters is not the boundaries of the agent, but whether there exists within it a subsystem capable of internalizing an adequate model of the relevant properties of the target.

This allows for cases in which understanding is distributed across multiple components. For instance, in Searle's \emph{Chinese Room} thought experiment \citep{Searle1980}, an English speaker follows a gigantic rule-book that pairs Chinese inputs with Chinese outputs. We may consider the target system to be the \emph{Chinese Language}. Searle contends that the individual in the room does not understand the Chinese language. A common response is to argue that the room-person system, taken as a whole nonetheless does understand the language. According to this response, the \emph{agent}, in our language,  would be the room-person system as a whole \citep{Harnad1989, Lycan1990, ChurchlandChurchland1990}.\footnote{To take another example, perhaps no individual within the intelligence services of Oceania understands the mechanisms behind the war with Eastasia \citep{Orwell1949}. Even so, the intelligence services as a whole may possess such an understanding, even though the requisite model is distributed across various files, algorithms and administrative clerks. As such we might think of the intelligence services as a whole as the \emph{agent}.}

\paragraph{Models of the System}

Now, models are often understood not as merely structures, but as intrinsically semantic, interpretive objects.\footnote{See \citet{Suppes1960, vanFraassen1980, daCosta1990} for semantic approaches for scientific models, and \citet{morgan1999models, Frigg2022, FriggNguyen2017} for additional discussion of the representative capacities that we might need to bestow upon models.} For instance, perhaps a formal structure within the agent only really counts as a model relative to additional interpretive scaffolding, including semantic assignments, idealizations, and background assumptions that enable the agent to use it representationally. For this reason, I distinguish between a bare structure $M$ and a model $\mathcal{M}$, where the latter includes whatever additional resources are required for the structure to play a representational role. However, nothing in my account, turns on this point: readers who prefer a purely syntactic conception of models may simply identify $M$ with $\mathcal{M}$ as needed.

By an \emph{adequate} model, I mean adequate for the purposes for which the agent will use the model. As \citet{parker2020model} emphasizes, what matters is not solely how accurately a model mirrors reality in general, but whether it is fit for the specific inferential or practical tasks for which it is employed. A model may idealize, distort, or even misrepresent aspects of a system, and yet remain adequate-for-purpose if it supports the kinds of epistemic or practical interventions the agent requires. However, at the very least, we require that an \textbf{adequate model} must allow the agent to make sufficiently accurate (approximate) predictions of property $p$, in order to endow our agent with understanding of $p$. Furthermore, such predictions must not be arbitrary, but derive from the fact that the model represents and tracks at least some features of the target system. This requires suitable bridge principles.

\paragraph{Bridge Principles}
\label{sec:bridge}

To construct a genuinely non-anthropocentric notion of systematic understanding, we must not demand that bridge principles rely on linguistic representations or something directly analogous to human mental states.  What matters is not that the agent \emph{interprets} its model in a reflective or semantic sense, but that there exists a stable interface through which the internal structure of the model can reliably track, and be used to derive claims about, properties of the target system. When such mappings are in place, the model’s internal variables can be said to be ``about'' the target system in a minimal but substantive sense: changes in the target systematically covary with systematic changes in the model, and the agent might exploit this relationship to generate predictions or guide action.

I do not intend to address the wider problem of intentionality, namely, what it is for a mental state, representation, or model component to be \emph{about} some external feature of the world. Instead, I will proceed with a modest naturalistic assumption: whatever intentionality consists in, it can be realized by physical systems (including engineered ones) in virtue of the roles their internal states play in reliable prediction, explanation, and action. Several possible accounts could suffice. First, on a broadly \emph{informational} or \emph{causal--covariational} view, internal states have content insofar as they carry information about external conditions: they stand in stable patterns of counterfactual dependence on features of the environment, and the agent can exploit those dependencies to guide successful prediction or action \citep{Dretske1981Flow, Dretske1988Explaining}. Second, on a \emph{teleofunctional} or etiological view, the relevant informational links are not mere accidental correlations but are stabilized by proper function, where correctness and error are fixed by the role that the state was selected, learned, or trained to perform \citep{Millikan1984LanguageThought, Neander1995Misrepresenting}. Third, on Dennett’s \emph{real-patterns} perspective, intentional characterizations are warranted at a coarser explanatory grain when they pick out objective patterns that support compression and reliable prediction across a range of counterfactual conditions \citep{Dennett1991RealPatterns}. On this view, treating some internal configuration as a representation is justified when doing so materially improves our explanatory and predictive grip.

These assumptions are intentionally modest. They do not require linguistic meaning, conscious awareness, or a special metaphysics of reference. They are, however, strong enough for what follows: they license the claim that, given suitable coupling and stability conditions, components of an agent's internal model can be \emph{about} features of a target system in a way that supports tracking, derivation, and the possibility of systematic error. This is exactly the sense of aboutness presupposed by the bridge principles.

Hence, by \textbf{bridge principles}, I mean any stable mappings, conventions, or interface relations that connect elements of the model $\mathcal{M}$ to elements of the target system $T$. As such, they are key to the semantic or interpretive component of understanding, fixing how the model is to be taken as being \emph{about} the target rather than about an arbitrary dataset or abstract mathematical structure.\footnote{Recent work on large language models suggests that representational aboutness may often be grounded in relatively stable regions of high-dimensional activation space, rather than in discrete symbolic vehicles \citep{Shea2007ContentVehicles, BallConcepts}.} These mappings determine which variables, states, or structures in the model correspond to which properties, quantities, or configurations of the target system, and thereby underwrite the model’s representational role. These need not take the form of explicit laws, identities, or semantic interpretations. For instance, in nonhuman agent systems, they may include procedures, tokenization or featurization schemes, conventions, evaluation protocols, or other engineering choices which serve to couple internal features of the model, $\mathcal{M}$, to features of the target system, $T$. In the context of machine learning systems bridge principles typically include the preprocessing steps that map target system states to data inputs, as well as the decoding rules that map internal weights or output scores to predictions, and the fixed routines by which those predictions are interpreted as claims about the target system. I will offer a more explicit unpacking of bridge principles in the deep learning case in section~\ref{sec:modelingmodeling}.

This raises a question about the \emph{boundaries} of the agent system. Bridge principles are not optional interpretive commentary supplied by an external observer, but an essential part of the agent system itself. When this framework is applied to deep learning systems (section~\ref{sec:modelingmodeling}), the natural locus of understanding is therefore often not the trained network in isolation, but the \emph{deployed computational pipeline} that couples the network to the target system and renders its internal states and outputs as determinate claims about $p$. As we will see, the deep learning agent system typically includes not only the neural network, but also rules of preprocessing, tokenization, sensor transduction, decoding, and other operations.

As such, I will treat bridge principles as part of the agent system only when they are (i) \emph{fixed at deployment}, rather than selected post hoc to rationalize success, (ii) \emph{executed automatically in normal operation}, rather than supplied ad hoc by an external evaluator, and (iii) \emph{available as part of the system’s competence}, in the sense that they participate in its closed-loop capacity to map target situations to task-relevant outputs. A further, stronger requirement is \emph{functional integration}: the bridging routines must be stably coupled to the learned component so that together they constitute a single, reliable end-to-end competence, rather than a mere attachment whose contribution is more appropriately credited to an external user. Without such constraints, claims about understanding would risk trivialization by tacitly importing additional and essential structure into the system.

\paragraph{Derivation}
The derivation condition is intended to exclude cases in which predictive success is merely read off by an external evaluator, rather than generated by a stable competence of the agent system itself. Accordingly, when I say that $S$ can use $M$ to (approximately) derive properties of $p$, I mean that there exists a \emph{fixed, agent-available procedure} by which $S$ maps situations of the target system (as presented under the bridge principles) into outputs that constitute claims about $p$. For machine learning systems, this procedure may include the trained forward pass together with the \emph{fixed} encoding, decoding, and evaluation conventions that are part of the deployed pipeline. By contrast, \emph{post hoc} scientific analyses performed only by the investigator (for example, interpretability probes trained after the fact, or external measurements that the deployed system does not itself run) may provide evidence that $M$ carries certain information, but they do not, by themselves, constitute the system's derivational competence unless those analyses are genuinely integrated into the agent system as deployed.

\paragraph{Non-Memorization}
Following section~\ref{sec:memorization}, we can operationalize non-memorization in several ways. One option is a \textbf{robustness test} of whether a system is merely memorizing, closely related to the tests performed by machine learning practitioners \citep{Vapnik1998Statistical, geirhos2020shortcut}. Fix a target system $T$, a property of interest $p$, and bridge principles that determine how data are generated and how predictions are evaluated. Consider a family of \emph{structure-preserving perturbations} $\Pi$ on the configurations of the target system $\Omega_T$, where each $\pi\in\Pi$ is designed to vary features that are \emph{irrelevant} to $p$ while preserving the salient structure that determines $p$.\footnote{I contend that what counts as ``structure-preserving'' is necessarily interest-relative and must be specified by the bridge principles. In image classification, for example, $\Pi$ might include changes in background or lighting; in a physical prediction task it might include perturbations that preserve the relevant conserved quantities; in a symbolic task it might include relabellings or symmetries that preserve the underlying algebraic relation.} Let $\widehat{p}(S,\omega)$ denote the system's predicted value for $p$ on a target situation $\omega\in\Omega_T$. Then a necessary condition for non-memorization is some degree of \emph{stability}:
\begin{equation}
\widehat{p}(S,\omega)\ \approx\ \widehat{p}(S,\pi(\omega))
\qquad\text{for all $\omega$ in the intended domain and for all $\pi\in\Pi$.}
\label{eq:stability-nonmem}
\end{equation}
Intuitively, a system that merely memorizes training-specific idiosyncrasies will tend to be fragile under such perturbations, because the idiosyncrasies do not persist across $\Pi$-variations. By contrast, a system that tracks a target-stable regularity can remain reliable across the perturbation family. In other words, a successful non-memorizing system should keep succeeding at predicting property $p$ when we \emph{change features that should not matter to $p$}. That is, if we intervene on the target situation or its presentation (for example, by altering irrelevant background factors, or by using a different but equivalent encoding or measurement setup) while holding fixed the underlying structure that makes $p$ true of the target, then the end-to-end pipeline should still make approximately correct predictions about $p$. 

We might supplement this robustness test with a \textbf{compression proxy}: holding the bridge principles fixed, a non-memorizing strategy is one whose effective description length (as captured by a suitable model class) does not scale linearly with the number of distinct training examples in the way a lookup strategy does. After all, robust structure-tracking typically permits a more compact description than sample-contingent encoding.

The robustness test and the compression proxy are complementary operationalizations of the same underlying requirement. A system that achieves genuine compression of target-stable regularities will tend to be robust under structure-preserving perturbations, because the compressed representation discards sample-contingent idiosyncrasies that do not persist across $\Pi$-variations. Conversely, a system that is robust across the perturbation family $\Pi$ has likely achieved some compression, since retaining all training-specific details would leave it fragile under perturbation. Together, these conditions make precise one sense in which understanding is a matter of degree: the greater the compression of target-stable structure, and the wider the family of perturbations across which predictions remain stable, the greater the degree of systematic understanding. On this picture, memorization and full systematic understanding are the two poles of a continuum, and most real systems occupy intermediate positions.

\subsection{Structural Understanding}
\label{sec:struc}

Structural understanding is one kind of systematic understanding, drawing from the contemporary approach to scientific structuralism (see \citealp{dewar2022structure, bokulich2011scientific}). We specify the tracking relation as follows:

\begin{quote}
    \begin{itemize}
        \item there exists a structure-preserving mapping between relevant parts of $\mathcal{M}$ and $T$, such that corresponding elements track the property $p$;
    \end{itemize}
\end{quote}

The structure-preserving relation need not be an isomorphism. In most cases it will be a homomorphism or other partial correspondence that preserves only those relations relevant to the property of interest.

As a simple example, let the agent system $S$ be a physics student equipped with a pen and paper and basic mathematical tools. Let the target system $T$ be a sample of unstable, decaying Barium-137m nuclei. Let the property of interest be the number of nuclei as a function of time,
\begin{equation}
    p \;:=\; \bigl\{\,N_T(t_T)\in\mathbb{R}_{\ge 0}\;\bigl|\;t_T\in\mathbb{R}_{\ge 0}\bigr\},
\end{equation}
\noindent where $N_T(t_T)$ denotes the number of undecayed nuclei at laboratory time $t_T$. To represent this system, the student employs an exponential decay model,
\begin{equation}
    \frac{dN_M}{dt_M}\;=\;-\lambda_M\,N_M(t_M),
\end{equation}
\noindent where $N_M$ and $t_M$ are model variables and $\lambda_M$ is a model parameter. Bridge principles identify $t_M$ with laboratory time $t_T$, interpret $N_M(t)$ as the expected number of undecayed nuclei at time $t$, and fix $\lambda_M=\lambda_T$ by an independent half-life measurement or by reference data in nuclear-data tables.

This defines a model structure
\begin{equation}
M=\bigl\langle
   \mathbb{R}_{\ge 0},\,
   \tfrac{d}{dt_M},\,
   \lambda_M
\bigr\rangle,
\end{equation}
\noindent interpreted with the usual rules of calculus and dimensional assignments, together with a structure-preserving mapping
\begin{equation}
h:\; N_T(\cdot)\;\longmapsto\; N_M(\cdot),
\end{equation}
\noindent where $h$ maps target trajectories to model trajectories under the identification $t_M=t_T$ and the interpretation of $N_M$ as an expectation value.

\noindent Because $h$ preserves the additive structure of the real numbers and commutes with the derivative operator under this identification, it is a homomorphism between the relevant dynamical substructure of $T$ (its mean-field decay law) and that of $\mathcal{M}$. In this sense, the same exponential pattern is instantiated in both the model and the target system.

The model is certainly idealized: the real decays are a probabilistic process. But if the model is successful, then our student can not only approximately \emph{predict} the number of Barium-137m nuclei over time. They also \textbf{understand} something about this property, that it arises from a decay process in which the rate of decay is proportional to the number of remaining nuclei. A deeper understanding might in turn derive the equations of this decay process from the nuclear structure of the particles.

\subsection{Reductive Understanding}
\label{sec:reduc}

Let us consider an alternative kind of systematic understanding, \textbf{reductive understanding}. This draws from the Nagelian-Schaffnerian tradition (see \citealp{nagel1961structure, schaffner1967approaches, dizadji2010afraid}). While structural understanding depends upon a homomorphism between model‐structure and target‐structure, reductive understanding is achieved when the agent possesses a model that \emph{derives} the relevant property of the target system from a more fundamental theoretical basis.

\begin{quote}
\begin{itemize}
        \item There is a reduction relation from part of the target system $T$ and the model $\mathcal{M}$, such that the properties of $\mathcal{M}$ can be derived from $T$, and we can approximately predict $p$ from $\mathcal{M}$.
\end{itemize}
\end{quote}

For present purposes, we do not need to specify this more fully. One possibility, following the generalized Nagel-Schaffner model, proposes three criteria for a reduction relation between $M$ and $T$: strong analogy, derivability, and connectability.

Perhaps both structural and reductive understanding may both be useful and distinct species of systematic understanding. Structural understanding is based on the requirement that the agent's model must track certain structural features of the target system. Reductive understanding is based on the requirement that the agent's model must be a reduction of the target system. It may be that the requirements for reductive understanding are as \emph{least as strong} as the requirements for structural understanding. That is, if the target system can be reduced to the model, then there must also be some structural homomorphism between the target system and the model.\footnote{See \citet{wallace2022stating} for a similar idea.} But it seems possible that we could sometimes have structural models without there being any \textbf{straightforward} reduction\footnote{I use the term ``\emph{straightforward}'' reduction judiciously. Perhaps, if the requirements for reductive understanding are sufficiently loose, any structural understanding could entail some kind of reductive understanding too.}. These could include empirical laws that successfully track structural regularities in the target system.\footnote{Possible examples, all contentious, include cases of Zipf's law \citep{Zipf1949human, Mitzenmacher2003generative, Piantadosi2014zipf, CanchoSole2003least}, the Tully-Fisher relations relating galaxy luminosities and rotational velocities \citep{Said2024TullyFisher} or effective chiral Lagrangians in low-energy Quantum Chromodynamics \citep{Georgi1993EFT}. }. Thus, I leave it open as to whether the requirements for reductive understanding are strictly stronger than the requirements for structural understanding.

\subsection{Causal Understanding}
\label{sec:causal}

Finally, let us consider one further kind of systematic understanding, \textbf{causal understanding}, in which the relevant tracking relation preserves not merely structural patterns but the target system's \emph{causal} dependency structure. As such, the model's variables correspond to features of the target system in a way that supports stable predictions under causal interventions, not merely under passive observation. Pearl's interventionist framework and do-calculus provide one natural way to make this precise \citep{Pearl2018}. In the Pearlian approach, the model tracks $p$ causally only if it represents interventional dependencies of the form $P(p \mid do(X{=}x))$, rather than merely observational correlations $P(p \mid X{=}x)$ \citep{Pearl2018}. Such an account might offer a richer conception of understanding than the structural variety developed here, and would connect naturally to recent work on causal discovery and causal representation learning \citep{scholkopf2021towards, peters2017elements}. I will not develop causal understanding further in this paper, but note it as a promising direction for extending the framework.

\subsection{Tacit and Symbolic Understanding}

We need to consider another distinction between kinds of systematic understanding. The examples I have discussed above were \textbf{symbolic}, expressed (somewhat) explicitly with mathematical terminology. But I suspect that most real understanding in human beings is not of this kind, but is rather more \textbf{tacit}, without the agent knowing how to express this in explicit symbolic form\footnote{Indeed, the ability to mentally, qualitatively manipulate a model without having to carry out exact mathematical calculations is central to the account of understanding  developed by \citep{deRegt2017}. I contend that at least some kinds of understanding are non-symbolic in this sense.}. This symbolic/tacit dichotomy does not perfectly map onto the structural/reductive distinction. However, \emph{generally speaking}, reductive understanding might be relatively more likely to be symbolic compared to structural understanding.

This symbolic--tacit dichotomy echoes Polanyi's distinction between tacit and explicit knowledge \citep{Polanyi1958}\footnote{Likewise, Ryle’s distinction between \emph{knowing‐that} and \emph{knowing‐how} \citep{ryle1949concept} marks the difference between possessing a catalog of explicit propositions (symbolic understanding) and possessing the trained disposition to act appropriately in concrete situations (implicit understanding).}. Polanyi’s \emph{tacit} component of personal knowledge encompasses precisely those sub-propositional skills and pattern recognitions. Let us apply this distinction not just to \emph{knowledge}, but also structural \emph{understanding}. 

Consider a physicist catching a ball.\footnote{Observational evidence suggests that some, though not all, physicists are capable of this.} Even the most committed theorist is unlikely to explicitly model the ball's trajectory mathematically, for example  by using Newton's equations of motion. Yet, they can adapt to subtleties in the ball's motion in a fraction of a second. The physicist's brain somewhere stores an implicit \emph{mental model} of the ball's trajectory, even if it is not an explicit one that they can symbolically describe. This mental model has clearly captured certain regularities about how the ball moves through space. And so the physicist has at least \emph{some} tacit, structural understanding of the ball's motion.

Alternatively, consider the ``waggle-dance'' communication of honey-bees. When a successful forager returns, she traces a figure-of-eight whose angle and duration encode the displacement vector to a nectar source. Recruits observe multiple dances, effectively average the encoded vectors, and then navigate to the advertised location. At the level of the colony, this implements vector integration in a two-dimensional metric space. Yet no single bee possesses the trigonometry capacity for such a computation. The spatial model is distributed across many individuals and implemented in sensorimotor routines rather than in explicit, symbolic form \citep{vonFrisch1967,Seeley1995}.

Finally, consider a chess grandmaster. Undoubtedly, they have a high-level understanding of the chess game, its rules and strategies. Yet they cannot rely on an explicit, symbolic traversal of an exponential game tree, but rather chess grandmasters excel due to superior recognition of structural patterns in the game \citet{deGroot1965} and \citet{GobetLane2005}.\footnote{Indeed, even Chess grandmasters typically only think a few moves ahead explicitly, during the mid-game. This can increase during early and late game phases.} Their skill reflects the internalization of a highly sophisticated model of the game, one that guides judgment and action without being fully expressible in symbolic form.

\subsection{The Ideal of Scientific Understanding}
\label{sec:ideal}

We might identify an epistemically privileged \textbf{Ideal of Scientific Understanding}, which has often historically been favored by both philosophers and scientists (although I do not claim that all or most scientific models meet this ideal). According to this ideal, understandings should be both \textbf{reductive}, \textbf{symbolic} and \textbf{unifying}.  There are good reasons for this. Symbolic representations are communicable, portable artifacts that can be inspected, critiqued, built upon and improved by other investigators, whereas an individual’s tacit grasp of a phenomenon is largely incommunicable. They can also be very directly manipulated through logical reasoning. Reductive models may offer especially suitable candidates for partial realist interpretations. In particular they may facilitate selective realism about the essential parts of the model (likely to survive under intertheoretic reduction) \citep{Psillos1999ScientificRealism, Worrall1989Structural, Freeborn2025-FRESMR-2}. 

Moreover reductive models may offer especially information-efficient, compressed representations of systems, in which a comparatively small stock of principles and derivational patterns can generate descriptions of a wide range of phenomena \citep{Friedman1974Explanation, Kitcher1981Unification, Grunwald2007MDL, nagel1961structure,Freeborn2025-FRESMR-2}. Such compression is sometimes considered an indicator that the model may have latched on to genuine structure in the world. 

The unification ideal connects to the compression framework already developed in section~\ref{sec:memorization}.On Kitcher's view, science advances understanding by deriving descriptions of many phenomena from the same patterns of reasoning, used again and again, thereby reducing the number of independent facts we must accept as brute. A unifying explanation does not merely represent a phenomenon symbolically, nor merely derive it from a more fundamental basis; it also brings many apparently distinct cases under a comparatively small stock of reusable principles, derivational patterns, or inferential templates. For instance, Newton's laws unify the motion of earthly cannonballs, and distant planets under a single framework. A maximally fragmented model, by contrast, may predict each phenomenon correctly without any of its components lending support to the others. Building on this \citet{Votsis2015} defines a unified hypothesis as one whose parts are \emph{confirmationally connected}, so that evidence for one part lends support to the others. A mere conjunction of independent claims, each supported only by its own datum, achieves no such connection, and is instead ``monstrous''. At least ideally, science offers unified theories, rather than monstrous patchwork models whose separate responses to different inputs are confirmationally disconnected.

The ideal of scientific understanding is plausibly strongest where these features coincide: where a representation is symbolic, reductive, and unifying. Nonetheless, it may be that \emph{much} (perhaps \emph{most}) of our actual understanding of real-world phenomena is most naturally thought of as tacit, structural and non-straightforwardly reductive. I suspect that, whatever implicit manipulations a physicist performs when they model a ball flying towards them, cannot be directly related to Newtonian mechanics through a reduction relation. Rather, their brain has latched onto, and modeled a mosaic of vague and overlapping structural regularities. Despite the advantages of symbolic, reductive and unified understanding, we should admit tacit, structural understanding as a genuine form of understanding.\footnote{True understanding should entail greater generalizability outside of the training data, but this is always limited. The physicist's heuristic model will succeed only in a limited domain. They may manage to catch a ball thrown slightly slower or faster, or at a different angle. But they will probably fail if asked to catch a ball in a different gravitational regime, such as on the moon. Newton's Laws generalize better, although of course they too would fail outside of a more general domain, for instance if the ball approaches the speed of light.}

In section~\ref{sec:fractured}, I will argue that contemporary deep learning systems often appear to fall short of this ideal, not necessarily because they fail to track any real regularities, but because the regularities they track are not always organized into a small set of globally coherent and reusable principles. The kinds of understanding that deep learning systems offer are typically symbolic (but using an unnatural symbolic representation), not clearly reductive, and not very unifying.

\section{Systematic Understanding in Deep Learning Systems}
\label{sec:modelingmodeling}

We are now in a position to assess whether, and in what sense, deep learning systems could ever satisfy the criteria for systematic understanding. For succinctness, we will focus on structural forms of understanding and on feed-forward ReLU neural network structures.

At first glance, it might seem that a neural network is precisely the type of system that is ill-suited for systematic understanding. After all, we have seen (section~\ref{sec:definition}), that they can be understood simply as curve fitters or \textbf{spline interpolators}. The system learns merely by systematically tweaking its vast array of parameters until the function approximately fits the data from the target system.  But let us keep an open mind for now. Insofar as this learned function approximates real regularities in the target system, then the deep learning system may have some systematic understanding of some properties in the target system. For example, if the learned function has some homomorphism to some features of the target system, then it may have a \textbf{structural understanding}; if it can reduce to a real function in the target system, then it may have some \textbf{reductive understanding}, assuming that the other criteria outlined in section~\ref{sec:systematic} are satisfied. 

Let the target system $T$ be characterized, relative to some property of interest $p$, by a structured domain
\begin{equation}
    \langle X_T, R_T \rangle,
\end{equation}
where $X_T$ is a space of relevant states, configurations, or conditions of the target system, and $R_T$ is a family of relations on $X_T$ that determine how variations in those states bear on $p$.\footnote{In typical supervised learning setups, $p$ is operationalized as an input--output relation (or conditional distribution) that generates labeled data. Nothing in what follows requires that $p_T$ be metaphysically fundamental; it may itself be an idealized or interest-relative property.} These relations may encode features like geometrical adjacency, algebraic composition, temporal ordering, or invariants that are salient for predicting or deriving the property in question.

We will need to treat $S$ as the full computational pipeline, which couples the target system to inputs and interprets outputs as claims about $T$. We can schematize this as
\begin{equation}
    \Omega_T \xrightarrow{\ \iota\ } \mathcal{X} \xrightarrow{\ f_\theta\ } \mathcal{Y} \xrightarrow{\ \delta\ } \widehat{\mathcal{Y}},
    \label{eq:pipeline}
\end{equation}
\noindent Here $\Omega_T$ is the space of \emph{target system situations}, for example configurations of a physical system, sequences of moves in a game, or ordered lists of integers in a mathematical task. The map $\iota : \Omega_T \rightarrow \mathcal{X}$ is an \emph{encoding or measurement interface}, which converts concrete target states into inputs the model can process. This includes sensor transduction (e.g.\ converting light into pixel values), featurization (e.g.\ extracting numerical descriptors from raw data), and tokenization (e.g.\ mapping words, board positions, or integers to discrete symbols). So, encoding is the point at which the target system is brought into contact with the learning architecture. The map $\delta : \mathcal{Y} \rightarrow \widehat{\mathcal{Y}}$ is a \emph{decoding or decision rule} that turns these raw outputs into interpreted predictions or actions. Examples include selecting the highest-scoring class (argmax classification), sampling from a probability distribution, extracting a geometric object from a learned scalar field, or issuing a control signal. The space $\widehat{\mathcal{Y}}$ consists of outputs as they are evaluated and acted upon. 

The learned function $f_\theta$ is specified by a neural network architecture and learned parameters, and captures all of the agent system's \emph{learned} information about $T$ (although of course some information will also be hardwired through the architecture or other inductive priors). Its outputs $\mathcal{Y}$ are typically uninterpreted numerical objects, such as vectors of logits, probability scores, or real-valued fields. The \emph{bridge principles}, including $\iota$ and $\delta$, allow subsets of the input space $\mathcal{X}$ and output space $\mathcal{Y}$ to be interpreted as representing elements of $X_T$, and thus allow $f_\theta$ to be \emph{about} the target system $T$, rather than as merely implementing an abstract transformation between datasets.\footnote{Recent work in philosophy of science has argued that, in some scientific contexts, machine-learned models can legitimately play theory-like roles, being both inferred from data and used to generate and test data-relevant predictions \citep{norelli2025}.}

In order to specify the syntactic component of the model, $M$, it is helpful to distinguish between the neural network's \emph{internal representational map} and a \emph{readout}. Concretely, we may factor the learned function $f_\theta : \mathcal{X} \rightarrow \mathcal{Y}$ as
\begin{equation}
    f_\theta = g_\theta \circ r_\theta,
\end{equation}
where $r_\theta : \mathcal{X} \rightarrow \mathcal{Z}$ maps inputs into an internal \emph{latent space} $\mathcal{Z}$, and $g_\theta : \mathcal{Z} \rightarrow \mathcal{Y}$ maps those internal representations to the system’s raw outputs. The latent space $\mathcal{Z}$ is specified by the neural network’s internal activation patterns, such as the activations of intermediate layers. These internal states are not directly observable at the level of inputs or outputs, but here they play an indispensable explanatory role.\footnote{This distinction is useful for two reasons. First, in most deep learning systems, the structure that supports generalization does not reside in the raw input space $\mathcal{X}$ itself, but in how inputs are mapped into and organized within the latent space $\mathcal{Z}$. Very different inputs may be mapped to nearby or systematically related points in $\mathcal{Z}$ if they are treated by the system as equivalent or similar for the purposes of the task. Second, the readout $g_\theta$ typically depends only on limited aspects of this internal organization, extracting task-relevant information while ignoring other variation present in the latent space.} As such, we can write the whole process as,

\begin{equation}
    \Omega_T \xrightarrow{\ \iota\ } \mathcal{X} \xrightarrow{\ r_\theta\ } \mathcal{Z} \xrightarrow{\ g_\theta\ } \mathcal{Y} \xrightarrow{\ \delta\ } \widehat{\mathcal{Y}},
    \label{eq:pipeline-full}
\end{equation}

 Let $X_M \subseteq \mathcal{Z}$ be the subset of latent states that the system actually occupies across its domain of application, and let $R_M$ be the family of relations on $X_M$ induced by the geometry and organization of $\mathcal{Z}$ together with the learned dynamics of the network. As we have already seen, in the case of feed-forward neural networks with ReLU activations, the learned function partitions $X_M$ into a finite collection of convex polytopal regions $\{\mathcal{P}_k\}$, within each of which $f_\theta$ is affine. The internal model component $M \subset S$ can therefore be treated as a structured object,
\begin{equation}
    \langle X_M, R_M \rangle,
\end{equation}
where $R_M$ includes relations such as adjacency between regions, linear dependence relations within regions, and continuity constraints across region boundaries. In this sense, the model instantiated by the network is a continuous, piecewise-linear structure, often described as a multivariate spline. On its own, $M$ is merely a piece of internal machinery. However, when coupled with the encoding and decoding interfaces $\iota$ and $\delta$, and with background task conventions that fix how internal states correspond to target-system properties, we can view $\mathcal{M}$ as a model of the target system $T$. Systematic understanding, on our account, requires that $\mathcal{M}$ stand in an appropriate relation to the relevant structure $\langle X_T, R_T \rangle$. For example, in the case of structural understanding, this should be a structure-preserving relation; in the case of reductive understanding it should be a reduction relation.

Seen in this light, spline interpolation is not inherently inimical to understanding. In principle, the models we have just described can track real patterns in the target system's data. Indeed, neural networks are extremely well suited for approximating a wide variety of functions. According to the \textbf{Universal Approximation Theorem}, a feed-forward neural network with a single hidden layer containing a finite number of neurons and a non-polynomial activation function $\sigma$ can approximate any continuous function on a compact subset of $\mathbb{R}^d$, to arbitrary precision, given sufficiently many parameters \citep{Cybenko1989Approximation}. Real neural networks are finite, but still show impressive expressive power. Deep learning systems with many such hidden layers increase the expressive capacity by composing many such approximations, yielding highly flexible families of piecewise-defined functions.

Yet, this raises an obvious worry. Most real-world systems are probably not governed by piecewise affine functions. It seems likely that these learning systems could learn to approximate the functions that describe such systems, at least in the target data, without approximating the true underlying regularities of the system. In that case, the learned model may succeed at prediction while remaining epistemically shallow. A memorizing system would be the extreme case of this. It may fit the training data by allocating idiosyncratic representational states (or, in ReLU terms, idiosyncratic affine regions) to specific inputs in the training data, while failing to preserve the relevant relations $R_T$ under systematic variation.

Yet this worry is not decisive. First, there is no requirement that a successful representational map must mirror the target’s underlying dynamics in its own native mathematical idiom. What matters for systematic understanding is whether some internal structure in the agent systematically approximately tracks a structure in the target. Crucially, a piecewise affine function can, in principle, encode and preserve non-piecewise regularities via a change of variables, or by approximating a smooth dependence closely enough on the relevant domain that the salient invariances are preserved. The key question, then, is not whether the learned function is literally piecewise affine, but whether its internal organization implements a stable, compressive, and generalizable dependence on the target property of interest.

Second, the depth of the concern depends on what we take approximating the true regularities to amount to. In many scientific contexts, we do routinely accept models whose functional form is non-reductive, piecemeal, or otherwise mathematically alien to the underlying microdynamics, so long as they capture the right dependence structure at the level of description that matters. Likewise, a network can be epistemically shallow if its fit is achieved by a patchwork of idiosyncratic regions that fail to extend beyond the training distribution, but it can also be epistemically deep if its learned representation induces a small set of robust relations that continue to govern its behavior under systematic variation. From the above account, we cannot rule out the possibility of deep learning systems exhibiting genuine structural understanding, even of highly complex target systems.

To get a sense of whether and when this kind of systematic understanding can arise in practice, we will next look at some simple examples. In each case, we will ask what a deep learning model does, and does not, learn about some salient properties of a target system. Each example is meant to be simple enough to be clearly interpretable. In each case, a machine learning agent system will learn to approximate a function through a spline-like interpolation. We will analyze to what extent this kind of curve-fitting endows our agent system with systematic understanding of the target system.



\section{Illustrative Examples}
\label{sec:examples}

\subsection{The Genus of a Torus}
\label{sec:Torus}

Our first example demonstrates that structural understanding of a global property can emerge from purely local supervision, without any explicit encoding of the target property. We consider a relatively trivial case: a neural network trained to learn an implicit representation of a toroidal surface embedded in three-dimensional space. We shall see that, by training the neural network on individual data points from the torus surface, the network gains a ``structural understanding'' of the genus-one topology of the torus.

The target system $T$ is a continuous scalar field on $\mathbb{R}^3$, defined by the function,

\begin{equation}
F(x, y, z) = \left(R - \sqrt{x^2 + y^2} \right)^2 + z^2 - r^2.
\end{equation}

Let $D=[-3.5,3.5]^3 \subset \mathbb{R}^3$. We define the target surface $\Sigma \;:=\; \{x\in D : F(x)=0\}.$  This is a toroidal surface of major radius $R = 2.0$ and minor radius $r = 0.7$. The training data are obtained by uniformly sampling data from $D$ and assigning each point its scalar value under $F(x, y, z)$, i.e. ${(x^i, y^i, z^i, F^i)}_{i=1}^{n}$. 

The agent system consists of a trained three-layer feed-forward network, together with a fixed post-processing pipeline that extracts an isosurface from the learned scalar field and computes topological invariants of that surface. The neural network is trained to minimize the mean squared error between the predicted and true scalar field values.\footnote{The network is a fully connected feed-forward neural network with ReLU activations (as defined in equation~\ref{eq:neural-net}), specifically, a 3-layer multilayer perceptron with architecture \texttt{Linear(3$\rightarrow$64) $\rightarrow$ ReLU $\rightarrow$ Linear(64$\rightarrow$64) $\rightarrow$ ReLU $\rightarrow$ Linear(64$\rightarrow$1)}. The model has a total of 4,481 learnable parameters. It is trained to regress the scalar field $F(x,y,z)$ whose zero-level set defines the torus, using 262{,}144 points sampled on a $64^3$ grid over $D$. The loss function is mean squared error (MSE), optimized using Adam with a learning rate of $10^{-3}$ for 5000 epochs. See the accompanying repository for the implementation details.} 

\begin{figure}[ht]
\centering
\centering
\begin{subfigure}{0.48\textwidth}
  \centering
  \includegraphics[trim={8cm 6cm 8cm 6cm}, clip, width=\linewidth]{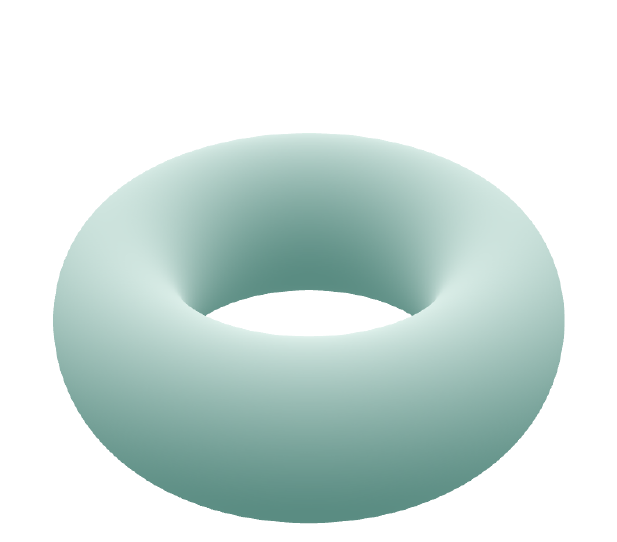}
  \caption{Ground truth surface}
\end{subfigure}
\hfill
\begin{subfigure}{0.48\textwidth}
  \centering
  \includegraphics[trim={8cm 6cm 8cm 6cm}, clip, width=\linewidth]{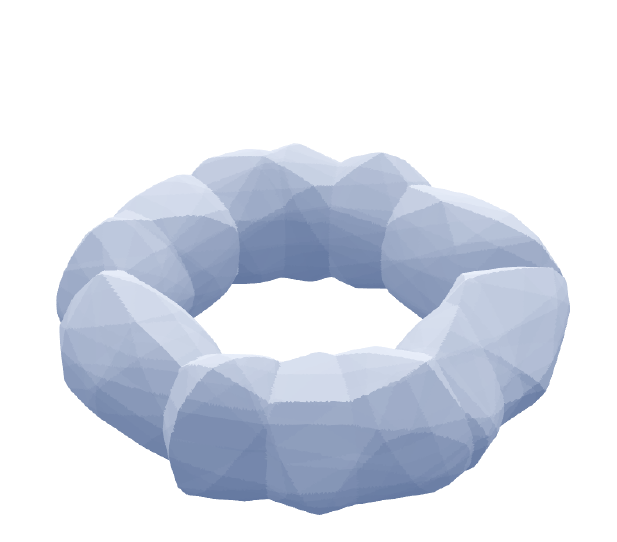}
  \caption{Learned isosurface}
\end{subfigure}
\caption{Comparison between the ground truth torus surface (left) and the neural network’s learned isosurface (right). Observe how the learned isosurface is made by stitching together piecewise-linear surfaces. Nonetheless, it forms a coherent genus-one surface. }
\label{fig:torus-learned}
\end{figure}

This architecture defines a function $f_\theta : \mathbb{R}^3 \rightarrow \mathbb{R}$ parameterized by weights $\theta$. The model $\mathcal{M}$ consists of this learned function, together with the fixed reconstruction and measurement procedures that extract a surface from $f_\theta$ and compute its topological properties. The system's goal is to learn an approximation of the function $F$, from which a learned surface, $\hat \Sigma \;:=\; \{x\in D : f_\theta(x)=0\}$ can be extracted. The property of interest, $p$, is the \emph{genus} of the target surface i.e. the property that the surface $F = 0$ is a connected, orientable surface of genus one. That is, the torus forms a single, continuous surface, with a consistent notion of inside and outside, and exactly one hole running through it (see figure~\ref{fig:torus-learned}).\footnote{Note that I extract the learned surface using the marching cubes algorithm \citep{lorensen1987marching} on a fixed grid. The resulting triangulation is therefore an artifact of the extraction procedure rather than a feature of the learned model itself. What matters for present purposes is that the reconstructed surface is globally coherent and piecewise planar, allowing its large-scale geometric and topological structure to be inspected. In this sense, marching cubes functions as a tool used by the human investigator, rather than as part of the agent system, to test hypotheses about the topology of the learned surface. The agent system’s internal model, understood as a learned scalar field, exists independently of this visualization procedure.}

The bridge principles are straightforward in this setting. Inputs in $\mathcal{X}=\mathbb{R}^3$ are interpreted as spatial locations in the domain $D$, and the real-valued output is interpreted as an estimate of the underlying scalar field at that location.\footnote{We might specify further bridge principles, mapping from the scalar field to a geometric object: the $0$-level set is interpreted as a surface. Furthermore, the agent system derives properties of $p$ by applying a fixed extraction-and-measurement procedure (marching cubes followed by Euler characteristic, hence genus) to the learned field. On this construal, the derivation step is performed by $S$ as a whole, namely the trained network taken together with this fixed post-processing pipeline.} 
It is natural to treat $\Sigma$ and $\hat \Sigma$ as topological spaces with the subspace topology inherited from $\mathbb{R}^3$, and to treat the relevant structure-preserving map as a homeomorphism.\footnote{Here, \emph{homeomorphic} means that there exists a continuous bijection between the reconstructed surface and the standard torus whose inverse is also continuous; equivalently, the two surfaces have the same topology (in particular, the same pattern of connectedness and holes), even if their geometries differ.} Concretely, the model tracks the structural property $p$ insofar as the learned surface $\hat \Sigma$ is topologically equivalent to $\Sigma$, for example insofar as there exists a homeomorphism\footnote{Note that, in practice we do not construct $h$ explicitly. Instead, we test the intended structural agreement by comparing topological invariants of the reconstructed surface.}

\begin{equation}
h: \Sigma \rightarrow \hat \Sigma.
\end{equation}

The trained model produces a surface with the correct \textbf{topological genus} (1), capturing the defining property of a torus: the presence of a single hole, and hence is homeomorphic to a torus.\footnote{As a check, let $\hat T_{h}$ be the triangulated $0$-isosurface extracted from $f_\theta$ by marching cubes on an $h$-spaced grid. In our reconstruction $\hat T_h$ is a single closed connected surface with Euler characteristic $\chi=0$, hence genus $g=(2-\chi)/2=1$.} In this sense, the learned model reproduces the salient topological structure of the target system, even though the training signal contains no explicit topological supervision. Consequently, according to our criteria, the network exhibits a form of structural understanding of the topological property $p =$ \textit{genus = 1}.  This is not a trivial accomplishment, as no global topological information about the genus was supplied to the model during training. The model was trained entirely on local scalar values, and yet the learned function globally reconstructs the correct topological structure (figure~\ref{fig:torus-learned}). 

This case illustrates several key points. First, structural understanding does not require explicit symbolic encoding of the target property. The network lacks any concept of genus, holes, or topology in the formal sense, and yet it implicitly represents those features through the structure of the learned function. In this sense, it may be somewhat analogous to cases of tacit understanding in humans. Second, this understanding emerges from local supervision alone: the system infers global structure from local data. Finally, the ReLU network's piecewise-linear function class is expressive enough to model a globally nonlinear manifold (a toroidal surface) by stitching together many locally affine patches, a realization of the spline interpretation discussed in section~\ref{sec:modelingmodeling}.

Thus, the neural network achieves a form of structural understanding of the torus: it creates a model that tracks and reproduces a salient topological property of the target system through its internal functional representation, without any explicit symbolic manipulation. The example illustrates that under appropriate conditions, deep learning systems can internalize structural regularities like topological genus. 



\subsection{Kepler's Elliptical Orbits}
\label{sec:Kepler}

Our second example concerns a different kind of global structure: not a topological invariant of a surface, but \emph{geometrical} regularities in a dynamical system. Unlike our previous example, these geometric quantities will only be learned \emph{approximately}. Recall, in section~\ref{sec:structural-understanding} I argued that Kepler achieved a greater understanding of the planetary orbits by realizing he could derive Brahe's descriptive ephemerides with a compact geometric model, in which the Sun occupies a focus of an ellipse \citep{Kepler1992NewAstronomy}. Instead, let us suppose that, in a moment of divine insight, an inventor (perhaps Wilhelm Schickard) presented Emperor Rudolf II with the \emph{Denkrechnuhr}, a mechanical neural network, to tackle the problem of planetary orbits.\footnote{Historically, Schickard described a calculating clock (or \emph{Rechenuhr}) in letters to Kepler in 1623--1624 (see \citealt{Kistermann1985AbridgedMultiplication,Seck2005NDBSchickard}).} Could such a system have achieved the same understanding as Kepler?

The target system $T$ is the orbit of Mars relative to the Earth and Sun. The properties of interest, $p$, include the basic observational regularities about the orbits (e.g. the synodic periodicity, and the profile of apparent angular speed), as well as the property that Mars's heliocentric orbit is well-approximated by an ellipse with eccentricity $e$, one of Kepler's notable findings being that the orbit was non-circular, $e \neq 0$. We will take the Denkrechnuhr to be the agent system, $S$. Concretely, $S$ consists of (i) a fixed neural network architecture, (ii) the learned parameter values obtained by training on the historical observations, and (iii) a fixed procedure for producing predictions and evaluating error. At prediction time, $S$ receives a date (represented as the number of days since the first observation) and outputs a predicted sky position for Mars in ecliptic coordinates (longitude and latitude). We will take a subset of Tycho Brahe's observations for Mars as the dataset \citep{Thoren1990LordOfUraniborg, BraheDreyer1913OperaOmnia}.\footnote{These consist of $12$ recorded oppositions. Each observation provides Mars's geocentric ecliptic longitude and latitude at a date $(\mathrm{day},\mathrm{month},\mathrm{year})$. In addition, the dataset includes the mean ecliptic longitude of the Sun at those epochs. A second table provides $10$ triangulation-style constraints, giving Earth heliocentric longitude and Mars geocentric longitude at additional dates. We represent each time-stamp by the number of days elapsed since the earliest observation, $t\in\mathbb{R}_{\ge 0}$, and represent angles in radians.} The system receives a date (represented as days since the first observation) and predicts the ecliptic longitude and latitude for Mars. 

We can consider two versions of the Denkrechnuhr, differing in model and bridge principles.\footnote{See the accompanying repository for the implementation details.} Let us call them the Baseline version, $S_B$, and the Keplerian version, $S_K$.  In the baseline version, the output is produced by directly regressing angles on time. The learned model $\mathcal{M}_B$ consists of two continuous, piecewise-linear, \emph{spline-like} maps from time to longitude and latitude, treating the observed angular trajectories as primitive objects to be fit. The bridge principles consist of the mapping from dates to temporal inputs, and the interpretation of the model’s outputs as sky angles measured in fixed coordinates. These mappings fix how the learned curves are to be taken as representing observational regularities, without directly positing any underlying spatial orbit structure. As such, the Baseline version treats Mars's observed sky-position as a time-indexed curve to be learned directly. We can think of it as drawing a smooth path through the recorded points on two graphs: one graph for longitude-versus-time and one for latitude-versus-time. Training amounts to adjusting the shape of those two curves until, at the recorded dates, the curves pass close to Brahe's measurements.

In the Keplerian version, the prediction procedure maps dates to angles \emph{indirectly} by first computing where Earth and Mars would lie under a simple orbital model with a small number of adjustable parameters, and then converting that direction into the predicted longitude and latitude. The model, $\mathcal{M}_K$, consists of a geometric representation of heliocentric motion, parameterized by orbital elements such as period, eccentricity, and orientation, together with deterministic rules that derive predicted sky angles from these spatial variables. As such, the learned spline encodes the set of global orbital parameters that pick out a specific Keplerian orbit (an ellipse with a particular eccentricity, orientation, and period).  The system then computes the planet’s position along that orbit over time and derives the corresponding sky angles by fixed geometric relations. Consequently, the observed angle–time curves are treated as consequences of the underlying geometric model. The bridge principles consist of the temporal mapping from dates to model times, the geometric mapping from internal spatial variables to heliocentric positions, and the coordinate transformations that map those positions to observable angular quantities. These mappings ensure that the observed angular trajectories are treated as consequences of an underlying spatial model rather than as primitive curves.

Necessarily, the Keplerian version bakes in some of Kepler's own assumptions: that planets move around the Sun on simple geometrical paths, and that the apparent motion of Mars arises from the \emph{relative motion} of Mars and Earth. Historically, Kepler's decisive move was to abandon ad hoc combinations of circles and epicycles and to search for a single simple curve that could generate the observations \citep{Voelkel2001CompositionAstronomiaNova}. In modern terms, his hypothesis was that the orbit is a \emph{conic section}, and in the bound case an \emph{ellipse}, with the Sun at one focus \citep{Kepler1992NewAstronomy}.\footnote{Kepler's assumptions function as strong inductive biases here. The model represents Mars's heliocentric path as an ellipse described by a small number of adjustable quantities (including its eccentricity, overall scale, period, and orientation). It then uses Kepler's area law (equal areas in equal times) to determine where Mars lies on that ellipse at each date. It treats Earth's motion in a simplified way and subtracts Earth's heliocentric position from Mars's to obtain the Earth-to-Mars direction. Finally, it converts that direction into the predicted ecliptic longitude and latitude that correspond to the observational data.} Instead of treating the angles as curves to be drawn directly, the Keplerian version treats them as consequences of an underlying spatial motion.

Both systems achieve a degree of predictive success. The Baseline system accurately reproduces the observed positions at the training dates and interpolates between them. It captures several genuine regularities of the target system, including the synodic period of Mars relative to Earth and the qualitative profile of apparent angular speed, including retrograde loops. In this sense, the Baseline model tracks real patterns in the data and so satisfies a very minimal condition for systematic understanding. What the baseline does \emph{not} provide on its own is a bridge from these observational quantities to heliocentric orbital elements like eccentricity or semi-major axis; extracting those would require imposing an additional spatial orbit model, which is precisely what the Keplerian version builds in from the start. 

The Keplerian system can track additional geometric properties of the orbits. In our results, it learns a Mars eccentricity value $e_M \approx 0.096$ and a period $P_M \approx 687.05$ days, both close to modern estimates (Mars has eccentricity $\approx 0.093$ and a sidereal period of $\approx 687$ days) \citep{JPLMarsPhysPar}. So, once the hypothesis class is restricted to Keplerian conic-section motion, the model can approximately track a global geometric property of the target system, namely non-circularity as quantified by $e_M$. What the model most clearly tracks here is a stable, nonzero eccentricity consistent with Kepler's elliptical hypothesis, rather than a fully faithful reconstruction of all geometric and dynamical details. So the system does build a rather accurate model of the orbital system, when we allow it to make certain assumptions, similar to those of Kepler. A more general system might be able to optimize to select conic sections from a wider space of models, starting with fewer initial assumptions.

The Keplerian system achieves a richer homomorphism between a low-dimensional geometric model and the observed data, mediated by bridge principles that connect spatial positions to angular observations. As such, it more closely resembles Kepler’s own achievement: not merely fitting the data, but showing how a wide range of observed phenomena arise from a single, simple underlying structure. At the same time, an important caveat applies. It is also plausible to regard the Keplerian system as exhibiting a genuine, though limited, form of reductive understanding. The target of reduction here is not planetary dynamics in full generality, but the structured body of observational regularities encoded in the angular ephemerides themselves. In the Keplerian model, those regularities are instead derived from a lower-dimensional spatial-geometric model of relative motion, together with explicit bridge principles linking heliocentric positions to observed angles.

It is perhaps unsurprising that the Keplerian system needs some greater input assumptions in order to achieve a greater understanding of the target system. One might reasonably question how much epistemic credit should be assigned to the learned parameters, as opposed to the hand-built scaffold in which learning occurs. Crucially, the Keplerian system’s success depends on inductive biases that arguably encode some of Kepler’s insight a priori. The restriction to conic sections, the assumption of heliocentric motion, and the use of area-preserving dynamics imposes a narrow hypothesis space for the deep learning model. From one perspective, this weakens the epistemic significance of the result: the system does not discover ellipticity from a neutral starting point, but rather selects parameter values within a family already known to be appropriate. From another perspective, however, this mirrors scientific practice. Kepler himself did not search the space of all possible curves, but worked within a constrained space of geometrically motivated hypotheses. On this interpretation, the Keplerian system exhibits a genuine, though conditional, form of structural understanding: conditional on the correctness of the imposed modeling assumptions, it successfully tracks a real and explanatorily salient property of the target system.

Ultimately, it is debatable as to the degree to which the Keplerian system is really a case of understanding in deep learning systems, as opposed to understanding in extended systems with inbuilt structures and biases (although it does fit the guidelines I suggest in section~\ref{sec:bridge}). Nonetheless, it is plausible that a more powerful deep learning model could feasibly operate without such stringent structures and biases. I leave this as an open question. Regardless, the Keplerian system demonstrates how the choices of representation and bridge principles can strongly influence  the degree of understanding that any given system can develop.

\subsection{Modulo Addition}
\label{sec:Modulo}

Our third example provides the clearest illustration of a system transitioning from memorization to genuine systematic understanding. We study a transformer-architecture neural network trained to perform \emph{modular addition}, closely following a model introduced by \citet{Power2022grokking} and analyzed by \citet{nanda2023progress}. We use this setup for a different purpose. Modular addition serves as a particularly clean test case for the idea that a system can move from merely \emph{memorizing} observed input-output pairs to internalizing a \emph{systematic} representation of the target structure. In the first possibility, the system internalizes a distributed lookup table, storing enough information about the particular training pairs it has seen to output the right labels for those pairs, without encoding any general rule. In the second possibility, the system internalizes a  procedure that effectively computes the remainder of $a+b$ for \emph{any} input pair, in other words it builds a \emph{model} of modulo addition. These two possibilities come apart sharply once we withhold a substantial fraction of the $m^2$ possible pairs for testing.  A memorizing solution has no particular reason to succeed on withheld pairs, since those pairs are not ``near'' the training pairs in a straightforward sense (unless there is some kind of internalization of the modular nature of the task).  By contrast, a genuinely rule-like solution should succeed uniformly across all pairs, because the same procedure applies everywhere.

The target system is the finite abelian group $\mathbb{Z}_m \times \mathbb{Z}_m$, where $\mathbb{Z}_m = {0,1,\dots,m-1}$ denotes the integers modulo $m$, equipped with componentwise addition modulo $m$. We choose to fix $m = 113$.\footnote{We choose $m$ to be prime because then $\mathbb{Z}_m$ has especially clean algebraic structure: every non-zero element has a multiplicative inverse, and (equivalently) $\mathbb{Z}_m$ forms a field.  This avoids certain extra symmetries and degeneracies that can appear for composite moduli (where $\mathbb{Z}_m$ has zero divisors).} In modular addition, we add two numbers as usual, and then keep only the remainder after dividing by $m$.  For example, modulo $113$ we have $(110+5)\bmod 113 = 115 \bmod 113 = 2$. We train on a random fraction of all $m^2$ pairs (in our main training run, $60\%$), and we test on the remaining held-out pairs. The property of interest is which output, $y$, is correct for each input pair of integers $(a,b)$.  

The agent system $S$ is a trained network, together with the fixed architectural choices, training procedure, and decoding conventions that define its end-to-end input–output behavior. Following the spirit of Nanda et al., we use a small transformer architecture, trained with gradient descent and weight decay.\footnote{The model takes a two-token input sequence $[a,b]$, embeds each token into $\mathbb{R}^{128}$, adds learned positional embeddings, applies a stack of transformer encoder blocks (in our implementation, two layers with $4$ attention heads and a $512$-dimensional feed-forward sublayer), and then produces logits over the $p$ possible outputs from the representation at the second token position. Our implementation is intentionally minimal. In particular, we do not include an explicit ``$=$'' token as in \citet{nanda2023progress}, and we use a standard PyTorch encoder block. These differences are not philosophically important here, since our aim is not a fine-grained mechanistic reconstruction of the internal circuit, but rather a demonstration that the learned input-output map can become both compact and systematically correct. We train with cross-entropy loss on the training pairs, and we evaluate accuracy on \emph{all} held-out pairs. See the accompanying repository for the implementation details.} The model, $\mathcal{M}$, is the learned input–output function implemented by the network, understood as a score surface over the discrete grid of inputs, together with the internal representations that support that mapping. The bridge principles include the representation of each input integer by a discrete token supplied to the network, the fixed association between output positions and the possible remainders $0,1,\ldots,m-1$, and a decoding rule that selects the output with the highest score as the system’s answer. Taken together, these mappings fix how the network’s internal activity is to be understood as implementing addition modulo $m$ rather than some other input--output transformation.

\begin{figure}[ht]
    \centering
    \includegraphics[width=0.8\linewidth]{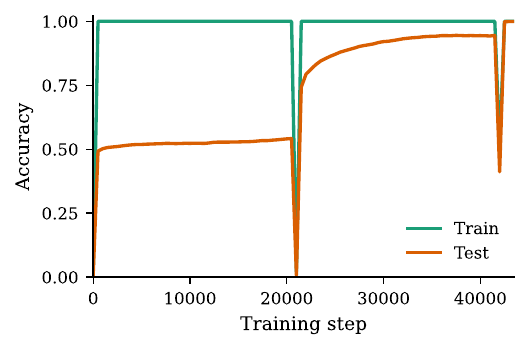}
    \caption{Training and test accuracy as a function of optimization step in the modular-addition experiment.  Training accuracy rises rapidly to near-perfect performance on the seen input pairs, while test accuracy remains low for an extended period.  After a period of continued training, the model \emph{groks} the target system, with test accuracy suddenly increases sharply, nearing the training accuracy. The sharp spikes at around steps 22,000 and 42,000 are consistent with the ``slingshot'' grokking mechanism identified by \citet{thilak2022slingshot}.}
    \label{fig:accuracy}
\end{figure}

Upon training and testing, we see a clear case of the \emph{grokking} phenomenon described in section~\ref{sec:memorgeneral}. As shown in figure~\ref{fig:accuracy}, the system attains near-perfect \emph{training} accuracy quite quickly, while \emph{test} accuracy remains low for a long time, and then, after many further training steps, test accuracy rises sharply to near-perfect values. Following the analysis of \citet{nanda2023progress}, it seems that the training initially settles into a solution that fits the training pairs by a form of partial memorization, but which fails to generalize to new, non-memorized pairs. However, later the system seems to internalize a (highly compressed) generalizing model of modulo addition, one which also succeeds at new test pairs. 

We gain further insight by analyzing the structure of the system's learned surface, as with our previous two examples. Suppose that the system receives an input, $a,b$; then let $S_y(a,b)$ be the model's score assigned to each possible answer $y$. A higher score means the model weighs $y$ more highly as an answer. For example, if our system has succeeded in learning additional modulo 113, we would expect $S_2 (110, 5)$ to be high, and $S_y (110, 5)$ to be low for every $y \neq 2$.\footnote{Many neural-network classifiers convert these scores into probabilities by applying a \emph{softmax} transformation.  Nothing in the present analysis depends on that step.  What matters is that higher scores indicate that the model ``prefers'' that answer, and the model’s predicted answer is the $y$ with the highest score.} Thus, for each fixed output class $y$, the assignment $(a,b)\mapsto S(a,b,y)$ is simply a real-valued function on the finite grid $\mathbb{Z}_m\times\mathbb{Z}_m$.  We can therefore view it as a numerical table with $m\times m$ entries, one entry for each possible input pair.

\begin{figure}[ht]
  \centering
  \includegraphics[width=0.8\textwidth]{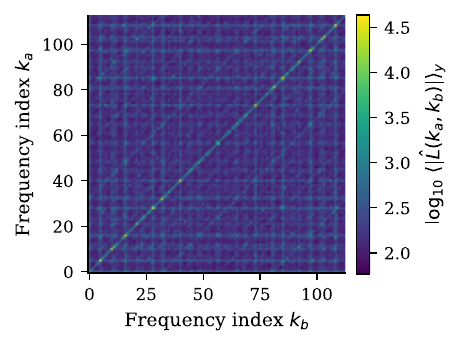}
  \caption{Average magnitude of the 2D discrete Fourier transform of the system's logits over the $(a,b)$ input grid (averaged over output classes). The learned input-output map exhibits strong concentration in Fourier space, consistent with a compact, structured dependence on the inputs.}
  \label{fig:modulo-fourier}
\end{figure}

Any such function on a finite grid admits a discrete Fourier decomposition.\footnote{Concretely, one expands in the characters $\exp(2\pi i(k_a a + k_b b)/m)$, where $(k_a,k_b)\in \mathbb{Z}_m^2$.  The discrete Fourier transform returns the coefficients of this expansion.}  This gives a clear way to extract the structure of the learned dependence on $(a,b)$.  Roughly, if the model’s score table is complicated in an idiosyncratic, example-by-example way (as one expects from a memorizing lookup strategy), then its Fourier representation tends to be spread broadly across many frequencies.  By contrast, if the score table is governed by a simple underlying regularity, then its Fourier representation tends to be \emph{concentrated} on a small, geometrically meaningful subset of frequencies. In particular, the rule should depend only on the sum $a+b$ modulo $m$.\footnote{This imposes a strong symmetry: inputs with the same value of $a+b \bmod m$ lie on diagonals of the $(a,b)$ grid, and the correct answer is constant along each such diagonal.  When a learned function is primarily ``a function of $a+b$'', this symmetry sharply constrains which Fourier components can carry substantial weight, and one expects the Fourier magnitude to concentrate on a low-dimensional locus rather than being diffuse across the plane.} Empirically, this is exactly what we observe.  Figure~\ref{fig:modulo-fourier} visualizes the average Fourier magnitude (averaged over output classes $y$), and it exhibits a pronounced concentration pattern rather than a high-entropy spectrum.  This provides further, strong evidence that the system really is internalizing a model of modular addition.\footnote{\citet{nanda2023progress} find that their grokked solution can be described as a ``Fourier multiplication'' algorithm that converts addition into composition of rotations, implemented by a small circuit inside the transformer.}

The evidence strongly suggests that the model $\mathcal{M}$ does not achieve success in an arbitrary way.\footnote{However, It is worth noting what enabled this transition. The training included weight decay, a form of regularization that penalizes large parameter values and thereby biases the system toward simpler solutions. Without it, the system remains in the memorization regime indefinitely \citep{Power2022grokking}. Additionally, the target domain possesses a genuine simple regularity, the group-theoretic structure of modular addition, which falls within what the architecture can represent.} The patterns of the learned score surface are organized around the same symmetry that organizes the target rule, namely dependence on the sum $a+b \bmod m$. There is a structure-preserving alignment between the organization of the model’s internal map from inputs to preferred outputs and the organization of the target rule itself.  The model is not merely correct, it is correct \emph{for the right kind of reason}, namely because its internal representation tracks a genuine regularity of $T$ rather than a patchwork of training-specific contingencies.\footnote{Nonetheless, the bridge principles that make the task well-defined (the tokenization of residues, the fixed decoding rule, the fixed modulus $m$) are doing substantial work, and nothing here suggests that the system has a general grasp of arithmetic beyond this carefully delimited setting.} This seems to be an especially clear-cut case of the system acquiring structural understanding.

\subsection{The Game Othello}
\label{sec:othello}

Our final example occupies a middle ground between the highly restricted modular-arithmetic task of section~\ref{sec:Modulo} and the rich, open-ended domains usually associated with natural language, and demonstrates that a system trained on sequential data can learn a hidden internal model of a complex state space. The system is trained on the board game Othello, developed and analyzed by \citet{li_othello}. We shall see that an agent system trained on data about game moves can nonetheless learn a hidden internal model of the game rules and board state. As Othello noted of Iago, this system ``\emph{doubtless Sees and knows more, much more, than he unfolds}'' \citep[Act~III, Scene~III]{ShakespeareOthello}.

\begin{figure}[ht]
\centering
\resizebox{\linewidth}{!}{%
\begin{tikzpicture}[
    board/.style={
        draw=gray!40,
        fill=gray!15,
        minimum size=0.6cm
    },
    blackdisc/.style={
        draw=black,
        fill=black,
        circle,
        minimum size=0.32cm
    },
    whitedisc/.style={
        draw=gray,
        fill=white,
        line width=0.8pt,
        circle,
        minimum size=0.32cm
    },
    label/.style={
        font=\small
    }
]

\newcommand{\drawboard}[2]{
    \foreach \x in {1,...,8} {
        \foreach \y in {1,...,8} {
            \node[board] at (#1+\x*0.6, #2+\y*0.6) {};
        }
    }
    \foreach \x in {1,...,8} {
        \node[label] at (#1+\x*0.6, #2+8*0.6+0.45) {\x};
    }
    \foreach \y/\l in {1/H,2/G,3/F,4/E,5/D,6/C,7/B,8/A} {
        \node[label] at (#1-0.45, #2+\y*0.6) {\l};
    }
}

\drawboard{0}{0}
\node at (2.4,5.9) {\large (A)};

\node[blackdisc] at (0+4*0.6,0+5*0.6) {};
\node[blackdisc] at (0+5*0.6,0+4*0.6) {};
\node[whitedisc] at (0+4*0.6,0+4*0.6) {};
\node[whitedisc] at (0+5*0.6,0+5*0.6) {};

\drawboard{6.2}{0}
\node at (8.6,5.9) {\large (B)};

\node[blackdisc] at (6.2+3*0.6,0+4*0.6) {};
\node[blackdisc] at (6.2+4*0.6,0+4*0.6) {};
\node[blackdisc] at (6.2+5*0.6,0+4*0.6) {};
\node[blackdisc] at (6.2+4*0.6,0+5*0.6) {};
\node[whitedisc] at (6.2+5*0.6,0+5*0.6) {};

\drawboard{12.4}{0}
\node at (14.8,5.9) {\large (C)};

\node[whitedisc] at (12.4+3*0.6,0+5*0.6) {};
\node[whitedisc] at (12.4+4*0.6,0+5*0.6) {};
\node[whitedisc] at (12.4+5*0.6,0+5*0.6) {};
\node[blackdisc] at (12.4+3*0.6,0+4*0.6) {};
\node[blackdisc] at (12.4+4*0.6,0+4*0.6) {};
\node[blackdisc] at (12.4+5*0.6,0+4*0.6) {};

\end{tikzpicture}
}
\caption{A visual demonstration of the Othello rules. From left to right. (A) The board is initialized with four discs placed at the center. (B) Black moves first and must outflank one or more opponent discs. (C) The opponent repeats this process. The game ends when no legal moves remain.}
\label{fig:othello}
\end{figure}
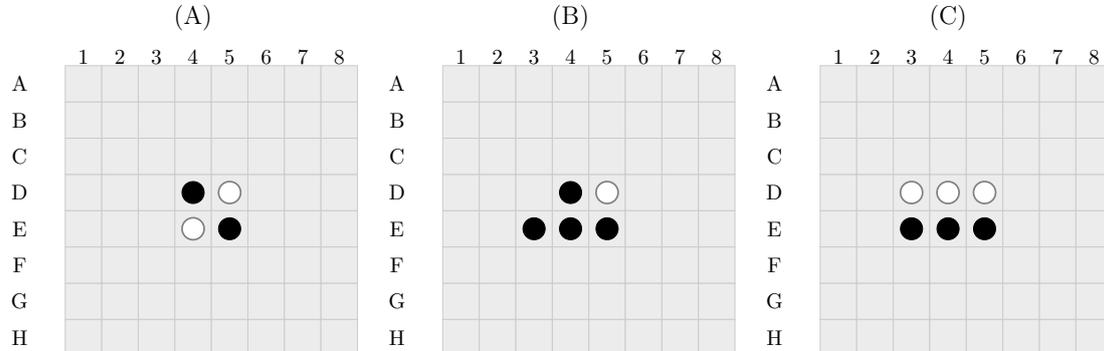

So our target system $T$ will be the Othello game dynamics (see figure~\ref{fig:othello}), understood as a deterministic state-update process over board configurations, together with a legality constraint. Othello is played on an $8\times 8$ square grid. Players alternate placing discs of their color on empty squares. A move is legal only if the placed disc \emph{outflanks} (sandwiches) one or more contiguous lines of the opponent’s discs in any of the eight directions; those outflanked discs are then flipped. The game begins from a fixed four-disc configuration in the center, and ends when no legal moves remain. Although the rules are simple to state, the game tree is large. Moreover, the legality of a move depends on a global property of the game state, namely the configuration of discs across the entire board, rather than on any simple local feature of the immediately preceding move.

Formally, we may represent the board state after $t$ moves as a function
\begin{equation}
B_t:\ \mathcal{X}\rightarrow{\text{black},\text{white},\text{empty}},
\end{equation}
where $\mathcal{X}$ is the set of board squares, and where $(B_{t+1},m_{t+1})$ is determined from $(B_t,m_t)$ by the rules of play (with $m_t$ the side to move at time $t$). One property of interest $p_l$ is the set of legal next moves. We will also focus on another particular property of interest, $B_t$, the state of the board at a given time, $t$. That is, we will ask whether the agent system's internal model contains a representation of the board state.

\citet{li_othello} train a General Purpose Transformer (GPT)-style language model, Othello-GPT, which we will take to be our agent system, together with a fixed inference procedure for next-move prediction. The model, $\mathcal{M}$ involves the internal activation patterns of the trained network, together with the pathways by which those activations influence next-move predictions.\footnote{See the accompanying repository an implementation of a small-scale replication.}

At each step, the system receives the sequence of ``move tokens'' corresponding to the moves played so far, and it produces as output a probability distribution over possible next move tokens. These move tokens are discrete symbols drawn from a fixed vocabulary. Each token corresponds to one of the $64$ squares on the $8\times 8$ board, and represents the act of placing a piece on that square.  A sequence of move tokens therefore represents a sequence of moves in a game, and, together with the fixed rules of Othello, determines a unique board configuration at each time step. The bridge principles include the convention that maps each move token to a specific square on the Othello board, and a fixed update rule that determines how a sequence of such moves generates a board configuration.

As such, the system's sole training objective is next-token prediction: given a partial game transcript, it must predict the next move made by a (synthetic or human) player. Crucially, the agent system is never given the rules of Othello and never shown the board state. All structural information about the game must therefore be inferred from statistical regularities in the move sequences themselves. Nonetheless, the trained model predicts legal moves with very high accuracy on held-out games. At a behavioral level, this already goes beyond straightforward memorization. \citet{li_othello} show that performance remains high even when large portions of the game tree are systematically removed from the training distribution, ruling out the hypothesis that the model simply recalls previously seen continuations. Instead, it must have internalized some regularities in the game rules. This can be tested further with probes and interventions.\footnote{It is important to distinguish two claims that are easily conflated. First, \emph{decodability} results show that some information about $B_t$ is present in internal activations, in the sense that an auxiliary predictor can recover it. Second, \emph{causal-use} results (for example, activation interventions that systematically shift predicted moves in line with counterfactual legality constraints) provide evidence that the model’s behavior depends on that internal information. Probes primarily support the first claim; interventions primarily support the second. Note that the probe and intervention analyses are epistemic tools used by the human (not the agent system) to test hypotheses about the network’s internal structure. The agent system's model exists independently of them, but is illegible to humans. These probes and interventions are not part of the agent system’s deployed pipeline, and are not needed for the agent system's understanding.}

A probe is an auxiliary model trained \emph{post hoc}, after the main agent system has been trained and frozen. They are not part of the original agent system and do not affect the system's behavior. The probes receive as input a hidden activation vector $h(t)$ (representing the weights for a particular inner layer and time step $t$) from the Othello-GPT network and are trained to predict the underlying game state, here the board configuration $B_t$. Concretely, probes are trained on pairs of the form $( h(t), B_t )$. Probes therefore provide a direct test of whether the internal states of the model provide sufficient information to reconstruct the state of the board. \citet{li_othello} demonstrate that nonlinear probes can predict the state of the board from this information with low error rates.\footnote{In particular, for models trained on synthetic game data, a simple two-layer nonlinear multilayer perceptron (MLP) probe recovers the color or emptiness of individual board squares with error rates as low as roughly 2 percent in mid network layers, corresponding to correct reconstruction of over 98 percent of the board state. Importantly, analogous probes trained on a randomly initialized network fail, indicating that the success of the nonlinear probes reflects learned structure rather than hardwired properties of the architecture.} The natural conclusion is that the Othello-GPT system learns an internalized, accurate-but-imperfect model of the board state.\footnote{A subsequent analysis by \cite{Nanda_othello} takes this further. From the model’s internal perspective, it is often more natural to represent each square not as ``black versus white'', but as ``\emph{mine} versus \emph{theirs}''. Once one makes this change of variables, much of the representation becomes linearly accessible! They show that, once the representation is viewed in an appropriate coordinate system, even \emph{linear} probes can accurately recover the board state. Linearity is significant here because a linear probe can only succeed if the relevant information is already organized in the model’s internal space as a separable variable, rather than being diffusely encoded across complex nonlinear interactions (see \citealp{AlainBengio2016Probes, ElhageEtAl2021TransformerCircuits, ElhageEtAl2022Superposition, Anthropic2023Monosemanticity, CunninghamEtAl2023SparseAutoencoders, HewittManning2019StructuralProbe}).}

Second, \citet{li_othello} test whether the decodable board information plays a causal role in next-move prediction. The basic idea is to take an internal activation state, decode a baseline board $B$, and then \emph{modify} the activation so that the probe instead decodes a counterfactual board $B'$, differing from $B$ at a chosen square. One then asks whether the model’s predicted legal moves shift in the manner one would expect \emph{if} the model were consulting the internal board representation in order to generate its output distribution. If changing only the internal state (without changing the input transcript) yields systematically different predicted moves aligned with the \emph{counterfactual} legality constraints, that is strong evidence that the model computes and uses an internal representation that functions as a proxy for the board state. They test deliberately unnatural board states that are far from the training data distribution, including states that cannot arise from any legal sequence of moves. They find that intervening on the internal representation produces predictions aligned with the legal moves of the intervened board even in these stringent settings \citep{li_othello}. This is difficult to reconcile with a picture on which the model merely memorizes a catalogue of local transcript correlations. Once again the natural conclusion is that the model has learned a model that represents the board state. 

Thus, there is strong evidence that our agent system has learned an \emph{adequate model component} $M\subset S$ that encodes information sufficient to reconstruct the board state. Second, there is a plausible tracking relation: the internal variable covaries with, and is causally implicated in, the model’s next-move distribution in a manner aligned with the target’s legality constraints. Third, bridge principles connect move tokens to board updates and connect internal activations to decoded board variables (whether in black/white or mine/theirs form). Fourth, the system uses this internal structure in its predictions. By our previous definitions, it seems reasonable to conclude that the system has achieved a systematic understanding of the Othello game and the state of the board.

\section{The Fractured Understanding Hypothesis}
\label{sec:fractured}

The four preceding examples each show cases in which deep learning systems \emph{can} acquire a degree of systematic understanding. In the torus case, the system learned exact, discrete topological properties of the target system, despite being trained only on local scalar data and despite implementing its hypothesis as a piecewise linear spline. In the Keplerian case, it learned to approximate continuous geometric properties, such as orbital eccentricity. In the modular-addition case, the transformer eventually internalizes a symmetry-respecting rule that generalizes correctly across the target domain. In the Othello case, the system learns an internal state variable that tracks the evolving board configuration and is used to guide the next-move predictions. 

However, the forms of understanding that deep learning systems typically acquire rarely resemble the \textbf{ideal of scientific understanding} proposed in section~\ref{sec:ideal}. In general, they are not straightforwardly reductive, they are symbolic but use unnatural symbolic representations, and they are only weakly unified. 

\subsection{Non-Reductive Understanding}
As spline-based function approximators, such systems are optimized to fit observed input--output relations as flexibly as possible, but these spline-like surfaces may be unsuitable in practice for capturing the salient structures of many target systems. This makes them extremely effective at capturing \emph{local} regularities, but comparatively ill-suited to discovering global regularities. Often, the target system may be well suited for a simple description in one coordinate system, or one hypothesis class, but the deep learning system can only represent a large patchwork of local facets, which fail to latch onto the target system's more general structures. In that case the learned map can be accurate on the training data while failing to preserve the right structure under the perturbations that matter, because small changes can move inputs across region boundaries of the learned surface. As such, the understanding is usually \textbf{structural} but rarely \textbf{reductive}.

Recall that deep learning systems are trained to minimize some loss quantity over samples drawn from a particular data-generating process. The training objective is rarely to discover a generalizable, reductive model of the underlying target structure $T$; it is to achieve low error on the observed input--output behavior induced by the bridge principles. Consequently, the easiest route to low loss is often to capture \emph{whatever} dependence structure is stable \emph{in the training distribution}, even if that dependence does not coincide with the intended property $p$ of the target system. As a result, even when the system achieves impressive generalization, the resulting internal model $\mathcal{M}$ need not be reductive, unified, or cleanly factorized into reusable variables.

This matches more general observations of deep learning systems. \citet{geirhos2020shortcut} discover that many apparent successes of deep neural networks can be traced to \emph{shortcut learning}: the acquisition of decision rules that perform well when training and test data are drawn from the same underlying distribution, yet fail to transfer under modest distribution shifts or more demanding test conditions. Shortcut solutions are not mere memorization, but they often encode \emph{regularities} that do not fully generalize. Deep learning systems are sensitive to superficially stable cues in the distribution of the training data but may miss deeper regularities in the target system.

Arithmetic in large language models might offer a further example. Contemporary LLMs succeed at many basic arithmetic tasks, such as addition and multiplication, but are increasingly likely to fail as numbers become larger \citep{Saxton2019MathematicalReasoning}. In a mechanistic study, \citet{Nikankin2025ArithmeticWithoutAlgorithms} find that transformer models are neither memorizing arithmetic patterns, nor learning the true underlying rules. Rather (perhaps like many students), they learn a \emph{bag of heuristics}.\footnote{Their experimental setup itself makes vivid one reason performance need not scale smoothly: each model tokenizes numbers as single tokens only up to a finite limit, and their analysis focuses on operand and result ranges within that regime.} The upshot is that they acquire a partial understanding: the system has latched onto genuine patterns in arithmetic, but these regularities do not fully generalize.

\subsection{Symbolic Misalignment}

Likewise, taken literally, a trained neural network does implement a symbolic object, namely a function defined by a finite parameter vector. But the symbols it employs, the coordinates of a high-dimensional activation space and the boundaries of affine regions, do not usually correspond to the natural symbolic structure of the target domain. Instead, the neural network systems usually seem to learn a patchwork of locally reliable regularities, stitched together into a large continuous function (or, in discrete settings, a large score table).

Recall that, in the Kepler \emph{baseline} system (section~\ref{sec:Kepler}), regressing longitude and latitude directly on time encourages precisely such a patchwork fit to the observed data: it captures real periodicities, but it cannot build a deeper general orbital model. By contrast, the Keplerian variant achieves a more robust kind of tracking mainly because we \emph{change the representational interface}: we forced the learned degrees of freedom to live in orbital-element space, so that the spline is fitting a small number of global parameters rather than stitching together an angle-time curve. But this involved hardwiring inductive biases into the system.

An analogous pattern appears in modular addition (section~\ref{sec:Modulo}): before grokking, the model can fit the training pairs using a diffuse, example-specific surface over $(a,b)$ that fails to extend uniformly to withheld pairs, and only later does training discover a representation aligned with the group structure, at which point generalization becomes global. Even the Othello case (section~\ref{sec:othello}) reflects a milder version of the same phenomenon: the internal state is present, but in a coordinate system (mine/theirs) that better matches the task’s invariances than our initial black/white description, which is why some structure is initially ``hidden'' until we adopt the right variables. The general lesson is that spline-like function classes are expressive enough to \emph{approximate} many regularities, but they do not automatically discover the right variables in which those regularities become simple; when they do not, understanding can remain locally adequate yet globally brittle.

\subsection{Weak Unification and Representational Fragmentation}

In section~\ref{sec:ideal}, I argued that the ideal of scientific understanding is not only reductive and symbolic, but also unifying: it organizes many phenomena under a compact set of reusable principles. Deep learning systems often fall short of this ideal. Their internal successes may be real, but they are frequently not organized into a small stock of globally reusable rules.

Recall the distinction between \emph{organic} and \emph{monstrous} theories \citep{Votsis2015}. A unified hypothesis is one whose parts are \emph{confirmationally connected}: evidence for one part lends support to the others, because they are bound together by shared structure. A ``monstrous'' theory, by contrast, is a mere conjunction of independent claims, each supported only by its own datum and confirming nothing about the rest. This notion of monstrosity gives us a precise way to characterize the patchwork understanding that deep learning systems often exhibit. A system whose separate responses to different inputs are only weakly confirmationally connected, each supported mainly by its own region of the training data, is monstrous in something like Votsis's sense, even if many individual responses happen to be correct. The modular-addition system \emph{before} grokking is a particularly clear example: it can fit individual input--output pairs without yet organizing them under a single general rule. After grokking, the same responses become confirmationally connected, because the same algorithmic structure governs all of them.

This vocabulary speaks directly to a recent debate about whether artificial neural networks are amenable to unifying explanation. \citet{Prasetya2022} argues that, on \citeauthor{Kitcher1981Unification}'s unificationist account, neural networks resist genuine explanation, because each trained network requires its own enormously complex derivation rather than instantiating a small number of reusable reasoning patterns. \citet{Erasmus2022} reply that this confuses being maximally unifying with being explanatory at all: even if neural-network derivations are less unifying than Newtonian mechanics, they do not thereby cease to be explanatory. Both sides capture something important. Prasetya is right that a typical trained network, with millions of individually tuned parameters, does not implement a small stock of reusable principles. However, Erasmus and Brunet are also right that some networks \emph{do} discover simple, general internal structures that apply uniformly across their whole domain.

Related concerns also arise in the machine learning literature. \citet{bengio2013representation} argue that the success of machine learning depends crucially on learning the right internal representations of target systems, in particular representations that disentangle the underlying explanatory factors of the data, rather than operating directly on raw inputs. Learned representations often capture only those aspects of structure that are useful for the training objective, leaving other salient factors unrepresented or only implicitly encoded. Similarly, \citet{kumar2025fer} argue that the objective-driven training of deep learning systems can yield internal representations in which the system's abilities are split into disconnected, redundant fragments, which then become entangled with other fragments in ways that hinder robustness and independent adaptation. Even in cases when output behavior is excellent, the internal strategy may therefore resemble an unruly patchwork rather than a unified, factored decomposition of the target structure. They call this the \emph{Fractured Entangled Representation} hypothesis.

More broadly, a growing philosophical literature has begun to address these questions directly. \citet{Sullivan2022Understanding} identifies ``link uncertainty''(i.e. the evidential gap between model and target) as key to determining whether machine learning supports genuine understanding. \citet{Tamir2023} propose indicators of machine understanding that naturally come apart across different systems and tasks. \citet{Beckmann2026} offer a tiered framework distinguishing levels of machine understanding, and observe that large language models rely on ``parallel, heterogeneous mechanisms'': different tasks within a single model are handled by different internal processes, some understanding-like and others not. These contributions characterize \emph{what kinds} of understanding deep learning systems exhibit. The question that remains is \emph{why} fragmented understanding is so common, and when, if ever, it gives way to genuine unification.

\subsection{Why Fractured Understanding Is Common}
\label{sec:why-fractured}

Why is fractured understanding so common? The answer is partly structural. Deep learning hypothesis classes are highly permissive: they admit both relatively unified and highly patchwork solutions. Yet patchwork solutions are often easier to find and easier to preserve.

First, they are \textbf{available}. A sufficiently expressive model can fit training data in ways that need not cohere into a single globally reusable scheme. As \citet{Zhang2017understanding} famously showed, the same architecture that generalizes impressively on structured data can also memorize pure noise to zero training error. The capacity that enables genuine pattern extraction therefore also enables rote memorization, shortcut dependence, and many intermediate forms of only partial or local adequacy. Such permissive hypothesis classes leave ample room for patchwork solutions.

Second, patchwork solutions are often more \textbf{discoverable}. Gradient-based optimization proceeds by local improvement, and so may settle on strategies that are adequate across many cases before it discovers a more globally unified organization. This helps explain the prevalence of \emph{shortcut learning}: models often latch onto features that are predictively useful in the training distribution, even when those features do not reflect the deeper structure of the task \citep{geirhos2020shortcut}. Such strategies are not mere memorization, but neither do they amount to a robust grasp of the target system's more general regularities.

Third, and critically, patchwork solutions are often \textbf{stable}. Once a locally adequate solution drives training loss sufficiently low, the objective typically supplies little pressure toward further reorganization. From the perspective of the loss function, a patchwork solution that performs well enough may be just as acceptable as a more unified one. Moving toward unification therefore usually requires some additional pressure toward simplicity, compression, or architectural constraint that the basic training objective does not itself provide.

This helps explain why Prasetya's observation is often empirically apt: many trained networks do not implement a small stock of globally reusable principles. Erasmus and Brunet are right, however, that this is not universal. Some systems do achieve more unified internal organization, but typically only when specific enabling conditions are in place.

The modular-addition case from section~\ref{sec:Modulo} illustrates the point. In that example, unification emerges when training includes sustained pressure toward compression, via weight decay, and when the target domain contains a simple discoverable regularity that the architecture can represent. Where such conditions are absent, as the LLM arithmetic case discussed above illustrates, fragmentation is more likely to persist. Conversely, some architectures can reduce the scope for patchwork by building stronger structural constraints directly into the model, for instance by encoding symmetries or conservation principles. The claim, then, is not that deep learning systems can never achieve unification, but that permissive hypothesis classes create a standing bias toward fractured understanding unless that bias is actively overcome.

\subsection{Statement of the Hypothesis}
\label{sec:fuh-statement}

These considerations motivate a \textbf{Fractured Understanding Hypothesis} (FUH). Deep learning systems are well-suited for acquiring \emph{some}, \emph{real} systematic understanding. But the understanding they acquire characteristically falls short of the unifying ideal, and does so in specific, predictable ways. The understanding often exhibits the following characteristic forms of fragmentation:

\begin{itemize}
\item \textbf{Symbolic Misalignment.} A trained network implements a precise symbolic function, but the internal ``coordinates'' in which it carries out the computation, such as activations, learned features, and affine-region boundaries, typically do not coincide with the target domain's natural variables, nor with variables that are easily legible to humans.
\item \textbf{Non-reducibility.} Instead of representing the phenomenon by a small set of general principles that unify many cases, the model often implements a patchwork of locally effective mappings. Even when each component tracks a real regularity, the overall representation need not amount to a small set of reusable principles that generate the behavior across the whole domain.
\item \textbf{Local fracturing.} Because the learned map is assembled from locally reliable pieces, its competence can depend on remaining within the same region of its internal partition of input space. Small, seemingly innocuous changes that push an input across such a boundary can trigger qualitatively different behavior, even when the corresponding change in the target system preserves the salient structure we care about.
\item \textbf{Proxy-dependent fragmentation.} Training often rewards any dependence that is predictively useful in the training distribution. As a result, the model may successfully identify regularities that exist only in the training data, while missing deeper regularities in the target system.
\item \textbf{Representational fragmentation.} Even when the model tracks the intended property, it may not represent the underlying factors as separable components. Instead, they can be distributed across many units and mixed together, so that the model cannot easily be decomposed into parts that can be cleanly recombined or reused.
\end{itemize}

\noindent These five characteristics are predictable consequences of the structural forces identified above. Symbolic misalignment arises because optimization finds whatever internal coordinates minimize loss, not the target's natural variables. Non-reducibility arises because patchwork is a stable resting point once predictive adequacy is achieved. Local fracturing arises from the piecewise nature of the learned function. Proxy-dependent fragmentation arises because training rewards any predictively useful dependence in the training distribution. Representational fragmentation arises because there is no exogenous pressure to factorize internal representations cleanly. In Votsis's terms, the resulting model may be at least partly monstrous: its components are only weakly confirmationally connected, each locally adequate but collectively lacking the organic unity of a genuinely unified model.

None of this undermines the claim that deep learning can yield genuine systematic understanding. Very plausibly, a deep learning system can track properties of interest under certain kinds of variation while failing under others, because different regions of the input space, or different kinds of perturbation, are governed by different local regularities. The result is that models often look like a mosaic or patchwork of partially overlapping structures, each reliable within a particular domain. As such, the learned model will often achieve generalization in some regimes and not in others. This entails a partial understanding, in which some real, often local, regularities in the target system are learned. This may be one cause of the ``jagged frontier'' pattern introduced in section~\ref{sec:introduction}.\footnote{Although more fundamentally, given their difference in architecture and training regimes from humans and our evolutionary environment, it is unsurprising that artificial neural networks would have greatly different capabilities from us.}

Of course, as we have discussed, human understanding may often be very similar to this. As I stressed in section~\ref{sec:ideal}, much, perhaps most, human understanding is itself tacit, non-reductive, and probably piecemeal. When humans walk, catch a ball, navigate a landscape, play chess, or learn to use language, we typically do not deploy explicit symbolic models or derive predictions from a small set of general principles. I contend that these \emph{are} real cases of human understanding, despite falling short of the ideal of scientific understanding. In many ways, the systematic understanding that deep learning systems acquire is, at least superficially, analogous to these cases of human understanding: it is usually non-reductive and technically symbolic, albeit in a non-natural way.

Taken together, these features help explain why the same system can appear, from one perspective, to display impressive understanding, and from another, to be epistemically shallow. The understanding it achieves is often real, but often partial: it tracks genuine regularities, often with remarkable precision, yet does so in a way that is locally constrained, distribution-sensitive, and poorly aligned with the explanatory structure of the target system.

With this being said, I do not claim that deep learning systems necessarily only exhibit fractured understanding. The examples we have seen suggest that they can and do develop a deeper understanding of many systems.\footnote{Indeed, it may be instructive to note that just as human researchers may develop successively improved scientific theories, so too can machine learning systems quite suddenly coalesce around new and improved models, under the appropriate circumstances. The double descent and grokking phenomena described in section~\ref{sec:memorgeneral} illustrate this in different ways: performance can appear to plateau or even degrade in one regime, yet later improve sharply as training or capacity pushes the system into a qualitatively different solution. This invites a loose analogy with the frameworks of \citet{Kuhn1962, Lakatos1978}. In Lakatosian terms, a research programme may maintain empirical success for a time through increasingly protective adjustments before being displaced by a rival programme. Likewise, in the double descent phenomenon, phases of overfitting can give way to new models that better track the underlying structure of the target system.}

\section{Conclusions}
\label{sec:conclusions}

I have proposed a minimal, naturalistic model of \textbf{systematic understanding}, suitable for deep learning systems. On this account, an agent system systematically understands a property of a target system when it contains an adequate internal model that tracks the relevant property without mere memorization, is coupled to the target by stable bridge principles and can be used to derive (approximately) correct predictions. This model might offer a clearer shared framework for the debate around understanding in deep learning systems. I hope that both proponents and skeptics of understanding in deep learning systems might locate their disagreements more precisely, or dissolve them, either by using this model of understanding, or by specifying alternatives to it. 

I have argued that deep learning systems often do achieve real systematic understanding, on the basis of this model. This understanding is substantial and goes far beyond mere compression. However, such understanding often fails to resemble the classical ideal of scientific understanding. This motivates the \textbf{Fractured Understanding Hypothesis}. Even when deep learning systems track genuine regularities, their understanding is often symbolically misaligned with the target domain’s natural variables, non-reductive and non-unifying, in the sense of being implemented as a patchwork of locally effective mappings rather than a small stock of reusable principles, and brittle, with competence that can change sharply across internal boundaries or distributional shifts.

One clear future step would be to apply this model to understanding in transformer-based large language models. Applying this framework would have two distinct steps. The first is to assess whether language models can and do systematically understand \emph{language} itself. Namely, do they build an internal model that captures the relevant structural distinctions in human language (whether those are syntactic features, compositional relations, or other contextual facets of linguistic meaning). The second is to assess whether such linguistic understanding, endows the models with understanding of the \emph{world}. Here the target properties are not purely intra-linguistic regularities but extra-linguistic structures of the world that language describes, for example potentially including causal, spatial and other relations. An LLM would count as having a \textbf{world model}, in the present sense, only if some internal variables reliably track those extra-linguistic regularities and can be used to derive predictions that remain stable under interventions that preserve the underlying structure \citep{ha2018worldmodels, li2023worldmodel}. 

Furthermore, the basic formulation of systematic understanding here might naturally invite other variants, beyond structural and reductive understanding. One natural and potentially promising contender would be a form of \textbf{causal understanding}. On such a variant, the relevant tracking relation would not merely preserve patterns, but would also capture the target’s \emph{causal} dependency structure: the model’s variables would correspond to features of the target system in a way that supports stable predictions under causal interventions, for example through a do-calculus framework \citep{Pearl2018}. This might well offer a richer conception of understanding than that offered here.

The framework also suggests some lessons for artificial intelligence development. First, deep learning can in principle achieve, and support, scientific and other kinds of understanding. However, this is greatly facilitated when the training, architecture, and interfaces of the system lead the learned representation to correspond with the target system’s salient invariances. Importantly, the bridge principles and representational interfaces are not neutral conveniences; they are major determinants of what kinds of understanding are even \emph{possible} to the system. The Keplerian contrast makes this explicit: changing the space in which learning occurs can convert a patchwork fit into a compact, globally meaningful model. More generally, if we care about scientific understanding, then we should expect to do substantive work in the design of measurement, featurization, tokenization, and decoding, not merely in scaling parameters.

Second, if fractured understanding is widespread, then it suggests a corresponding research program. We should expect progress to come from methods that encourage factorization, variable discovery, and robust structure-preservation. One natural approach would be hybrid architectures. Neuro-symbolic systems, broadly construed, promise a way to combine the expressive pattern-learning strengths of neural models with symbolic resources that support explicit variable-binding, compositional derivations, and re-use across contexts \citep{Votsis2024, LeCun2022Path, garcez2002neural, besold2018neural, rocktaschel2017neural}. Importantly, this should not be read as a return to a strict dichotomy between ``mere'' pattern recognition and ``real'' reasoning. Human cognition itself plausibly combines tacit, model-based competence with more explicit symbolic scaffolding. There is no principled reason to rule out a comparable division of labor in artificial systems, whether by integrating symbolic modules, by training neural components to implement more explicit algorithmic subroutines, or by coupling learned representations to external tools and verification procedures. 

Symbolic regression methods \citep{schmidt_lipson_2009, udrescu_tegmark_2020, cranmer_etal_2020, petersen_etal_2021, kamienny_etal_2022} offer an alternative, but related, route forwards. Whereas standard deep learning typically yields an implicit model whose structure is difficult to inspect, symbolic regression aims to extract an explicit, human-legible functional form (often a compact equation) that approximately governs the dependence between variables in the learned representation or in the target data. When it succeeds, this can convert tacit structural understanding into something closer to the scientific ideal.

Other promising approaches to solving related problems could be explicitly causal models \citep{Pearl2018, spirtes2000causation, scholkopf2021towards, peters2017elements}. Recent work on causal discovery and causal representation learning makes this idea precise, aiming to recover latent variables and structural equations that are invariant across heterogeneous data or experimental settings. Alternatively, scientific machine learning approaches, such as equivariant architectures and Hamiltonian or Lagrangian neural networks, suggest that imposing the right structural biases can turn locally effective fits into compact, globally meaningful models that more closely resemble the scientific ideal \citep{greydanus2019hamiltonian, cranmer2020lagrangian}.

The framework also motivates the growing body of work on mechanistic interpretability \citep{olah2020zoom, olah2018interpretability, elhage2021framework}. If systematic understanding requires that internal variables track target-relevant structure and support stable predictions under intervention, then interpretability is not merely a transparency add-on but an epistemic tool. Mechanistic analyses aim to identify internal components and circuits that play determinate computational roles, allowing us to test whether competence depends on robust, reusable structure or on brittle, locally effective shortcuts.

If the general argument of this paper is correct, then we can make progress in the understanding debate without either inflating deep learning into a human analogue or dismissing it as mere mimicry. Deep learning systems can acquire real systematic understanding, but the default form of that understanding is often fractured. The philosophical and engineering challenge is therefore not to decide, in the abstract, whether deep learning ``can understand'', but to identify the conditions under which it does, to diagnose the characteristic failure modes, and to develop methods that turn locally successful tracking into more robust, communicable, and generalizable models of the world.

\bibliographystyle{apalike}
\bibliography{references}

@article{Votsis2015,
  author  = {Votsis, Ioannis},
  title   = {Unification: Not Just a Thing of Beauty},
  journal = {Theoria},
  volume  = {30},
  number  = {1},
  pages   = {97--114},
  year    = {2015},
  doi     = {10.1387/theoria.12695}
}

@article{Prasetya2022,
  author    = {Prasetya, Yunus},
  title     = {{ANNs} and Unifying Explanations: Reply to {Erasmus}, {Brunet}, and {Fisher}},
  journal   = {Philosophy \& Technology},
  volume    = {35},
  number    = {2},
  articleno = {43},
  year      = {2022},
  doi       = {10.1007/s13347-022-00540-4}
}

@article{Erasmus2022,
  author    = {Erasmus, Adrian and Brunet, Tyler D. P.},
  title     = {Interpretability and Unification},
  journal   = {Philosophy \& Technology},
  volume    = {35},
  number    = {2},
  articleno = {42},
  year      = {2022},
  doi       = {10.1007/s13347-022-00537-z}
}

@article{Sullivan2022Understanding,
  author  = {Sullivan, Emily},
  title   = {Understanding from Machine Learning Models},
  journal = {The British Journal for the Philosophy of Science},
  volume  = {73},
  number  = {1},
  pages   = {109--133},
  year    = {2022},
  doi     = {10.1093/bjps/axz035}
}

@article{Tamir2023,
  author    = {Tamir, Michael and Shech, Elay},
  title     = {Machine Understanding and Deep Learning Representation},
  journal   = {Synthese},
  volume    = {201},
  articleno = {51},
  year      = {2023},
  doi       = {10.1007/s11229-022-03999-y}
}

@article{Beckmann2026,
  author  = {Beckmann, Pierre and Queloz, Matthieu},
  title   = {Mechanistic Indicators of Understanding in Large Language Models},
  journal = {arXiv preprint arXiv:2507.08017},
  year    = {2025},
  doi     = {10.48550/arXiv.2507.08017}
}

@article{Saxton2019MathematicalReasoning,
  title   = {Analysing Mathematical Reasoning Abilities of Neural Models},
  author  = {Saxton, David and Grefenstette, Edward and Hill, Felix and Kohli, Pushmeet},
  journal = {International Conference on Learning Representations},
  year    = {2019}
}

@article{Nikankin2025ArithmeticWithoutAlgorithms,
  title   = {Arithmetic Without Algorithms: Learning Exact Computation with Neural Networks},
  author  = {Nikankin, Ivan and Mikolov, Tomas and others},
  journal = {arXiv preprint arXiv:2501.XXXXX},
  year    = {2025}
}

@article{bengio2013representation,
  title   = {Representation Learning: A Review and New Perspectives},
  author  = {Bengio, Yoshua and Courville, Aaron and Vincent, Pascal},
  journal = {IEEE Transactions on Pattern Analysis and Machine Intelligence},
  volume  = {35},
  number  = {8},
  pages   = {1798--1828},
  year    = {2013}
}

@misc{kumar2025fer,
      title={Questioning Representational Optimism in Deep Learning: The Fractured Entangled Representation Hypothesis}, 
      author={Akarsh Kumar and Jeff Clune and Joel Lehman and Kenneth O. Stanley},
      year={2025},
      eprint={2505.11581},
      archivePrefix={arXiv},
      primaryClass={cs.CV},
      url={https://arxiv.org/abs/2505.11581}, 
}

@article{norelli2025,
	author = {Maria Federica Norelli and Ioannis Votsis and Jon Williamson},
	doi = {10.1017/psa.2025.10161},
	journal = {Philosophy of Science},
	pages = {1--16},
	title = {The Interplay of Data, Models, and Theories in Machine Learning},
	year = {forthcoming}
}

@InProceedings{Votsis2024,
author="Votsis, Ioannis",
editor="Ippoliti, Emiliano
and Magnani, Lorenzo
and Arfini, Selene",
title="A Neuro-symbolic Approach to the Logic of Scientific Discovery",
booktitle="Model-Based Reasoning, Abductive Cognition, Creativity",
year="2024",
publisher="Springer Nature Switzerland",
address="Cham",
pages="306--330",
isbn="978-3-031-69300-7"
}

@article{ha2018worldmodels,
  title         = {World Models},
  author        = {Ha, David and Schmidhuber, J{\"u}rgen},
  journal       = {arXiv preprint arXiv:1803.10122},
  year          = {2018},
  url           = {https://arxiv.org/abs/1803.10122}
}

@article{li2023worldmodel,
  title         = {Do Large Language Models Have a World Model?},
  author        = {Li, Kenneth and others},
  journal       = {arXiv preprint arXiv:2303.15447},
  year          = {2023},
  url           = {https://arxiv.org/abs/2303.15447}
}

@book{garcez2002neural,
  title     = {Neural-Symbolic Learning Systems: Foundations and Applications},
  author    = {Garcez, Artur d'Avila and Gabbay, Dov M. and Lamb, Luis C.},
  year      = {2002},
  publisher = {Springer},
  series    = {Perspectives in Neural Computing}
}

@article{besold2018neural,
  title   = {Neural-Symbolic Learning and Reasoning: A Survey and Interpretation},
  author  = {Besold, Tarek R. and d'Avila Garcez, Artur and Bader, Sebastian and Bowman, Howard and Domingos, Pedro and Hitzler, Pascal and Lamb, Luis C. and Lowd, Daniel and de Penning, Leo and Pinkas, Gadi and Poon, Hoifung and Zaverucha, Gerson},
  journal = {Cognitive Systems Research},
  volume  = {52},
  pages   = {1--22},
  year    = {2018}
}

@inproceedings{rocktaschel2017neural,
  title     = {End-to-End Differentiable Proving},
  author    = {Rockt{\"a}schel, Tim and Riedel, Sebastian},
  booktitle = {Advances in Neural Information Processing Systems},
  volume    = {30},
  year      = {2017}
}

@article{schmidt_lipson_2009,
  author    = {Michael Schmidt and Hod Lipson},
  title     = {Distilling Free-Form Natural Laws from Experimental Data},
  journal   = {Science},
  year      = {2009},
  volume    = {324},
  number    = {5923},
  pages     = {81--85},
  doi       = {10.1126/science.1165893},
  url       = {https://www.science.org/doi/10.1126/science.1165893}
}

@article{udrescu_tegmark_2020,
  author    = {Silviu-Marian Udrescu and Max Tegmark},
  title     = {AI Feynman: A Physics-Inspired Method for Symbolic Regression},
  journal   = {Science Advances},
  year      = {2020},
  volume    = {6},
  number    = {16},
  pages     = {eaay2631},
  doi       = {10.1126/sciadv.aay2631},
  eprint    = {1905.11481},
  archivePrefix = {arXiv},
  primaryClass = {cs.LG},
  url       = {https://www.science.org/doi/10.1126/sciadv.aay2631}
}

@inproceedings{cranmer_etal_2020,
author = {Cranmer, Miles and Sanchez-Gonzalez, Alvaro and Battaglia, Peter and Xu, Rui and Cranmer, Kyle and Spergel, David and Ho, Shirley},
title = {Discovering Symbolic Models from Deep Learning with Inductive Biases},
year = {2020},
isbn = {9781713829546},
publisher = {Curran Associates Inc.},
address = {Red Hook, NY, USA},
booktitle = {Proceedings of the 34th International Conference on Neural Information Processing Systems},
articleno = {1462},
numpages = {14},
location = {Vancouver, BC, Canada},
series = {NIPS '20}
}

@inproceedings{petersen_etal_2021,
  author    = {Brenden K. Petersen and Mikel Landajuela Larma and Terrell N. Mundhenk and Claudio Prata Santiago and Soo Kyung Kim and Joanne Taery Kim},
  title     = {Deep Symbolic Regression: Recovering Mathematical Expressions from Data via Risk-Seeking Policy Gradients},
  booktitle = {International Conference on Learning Representations (ICLR)},
  year      = {2021},
  url       = {https://openreview.net/forum?id=m5Qsh0kBQG}
}

@inproceedings{kamienny_etal_2022,
author = {Kamienny, Pierre-Alexandre and d'Ascoli, St\'{e}phane and Lample, Guillaume and Charton, Fran\c{c}ois},
title = {End-to-end symbolic regression with transformers},
year = {2022},
isbn = {9781713871088},
publisher = {Curran Associates Inc.},
address = {Red Hook, NY, USA},
booktitle = {Proceedings of the 36th International Conference on Neural Information Processing Systems},
articleno = {746},
numpages = {13},
location = {New Orleans, LA, USA},
series = {NIPS '22}
}

@book{spirtes2000causation,
  title     = {Causation, Prediction, and Search},
  author    = {Spirtes, Peter and Glymour, Clark and Scheines, Richard},
  year      = {2000},
  edition   = {2},
  publisher = {MIT Press}
}

@article{scholkopf2021towards,
  title   = {Toward Causal Representation Learning},
  author  = {Sch{\"o}lkopf, Bernhard and Locatello, Francesco and Bauer, Stefan and Ke, Nan Rosemary and Kalchbrenner, Nal and Goyal, Anirudh and Bengio, Yoshua},
  journal = {Proceedings of the IEEE},
  year    = {2021},
  volume  = {109},
  number  = {5},
  pages   = {612--634}
}

@book{peters2017elements,
  title     = {Elements of Causal Inference: Foundations and Learning Algorithms},
  author    = {Peters, Jonas and Janzing, Dominik and Sch{\"o}lkopf, Bernhard},
  year      = {2017},
  publisher = {MIT Press},
  address   = {Cambridge, MA}
}

@inproceedings{greydanus2019hamiltonian,
author = {Greydanus, Sam and Dzamba, Misko and Yosinski, Jason},
title = {Hamiltonian neural networks},
year = {2019},
publisher = {Curran Associates Inc.},
address = {Red Hook, NY, USA},
booktitle = {Proceedings of the 33rd International Conference on Neural Information Processing Systems},
articleno = {1378},
numpages = {11}
}

@article{cranmer2020lagrangian,
  title   = {Lagrangian Neural Networks},
  author  = {Cranmer, Miles and Greydanus, Samuel and Hoyer, Stephan and Battaglia, Peter and Spergel, David and Ho, Shirley},
  journal = {arXiv preprint arXiv:2003.04630},
  year    = {2020}
}

@article{olah2020zoom,
  title   = {Zoom In: An Introduction to Circuits},
  author  = {Olah, Chris and Cammarata, Nick and Schubert, Ludwig and Goh, Gabriel and Petrov, Michael},
  journal = {Distill},
  year    = {2020}
}

@article{olah2018interpretability,
  title   = {The Building Blocks of Interpretability},
  author  = {Olah, Chris and Mordvintsev, Alexander and Schubert, Ludwig},
  journal = {Distill},
  year    = {2018}
}

@misc{elhage2021framework,
  author       = {Elhage, Nelson and Hume, Tristan and Olsson, Catherine and Nanda, Neel and Henighan, Tom and Joseph, Nicholas and Mann, Ben and Askell, Amanda and Bai, Yuntao and Chen, Anna and Conerly, Tom and DasSarma, Nova and Drain, Dawn and Ganguli, Deep and Lovitt, Liane and Hatfield-Dodds, Zac and Kernion, Jackson and Jones, Andy and Brown, Tom and Clark, Jack and Kaplan, Jared and McCandlish, Sam and Amodei, Dario and Olah, Christopher},
  title        = {A Mathematical Framework for Transformer Circuits},
  year         = {2021},
  howpublished = {Transformer Circuits Thread},
  note         = {Available at \url{https://transformer-circuits.pub/2021/framework/index.html}},
}

@inproceedings{lorensen1987marching,
  title     = {Marching Cubes: A High Resolution 3D Surface Construction Algorithm},
  author    = {Lorensen, William E. and Cline, Harvey E.},
  booktitle = {Proceedings of the 14th Annual Conference on Computer Graphics and Interactive Techniques (SIGGRAPH '87)},
  year      = {1987},
  pages     = {163--169},
  publisher = {ACM}
}

@article{Shea2007ContentVehicles,
  author  = {Shea, Nicholas},
  title   = {Content and Its Vehicles in Connectionist Systems},
  journal = {Mind \& Language},
  volume  = {22},
  number  = {3},
  pages   = {246--269},
  year    = {2007},
  doi     = {10.1111/j.1468-0017.2007.00310.x}
}

@unpublished{BallConcepts,
  author = {Ball, Brian and Freeborn, David and Helliwell, Alice and Loi-Heng, Kevin},
  title  = {Concepts and Classification Algorithms: A Case Study Involving a Large Language Model},
  note   = {Unpublished manuscript},
  year   = {2025}
}

@book{Dretske1981Flow,
  author    = {Dretske, Fred I.},
  title     = {Knowledge and the Flow of Information},
  year      = {1981},
  publisher = {MIT Press},
  address   = {Cambridge, MA}
}

@book{Dretske1988Explaining,
  author    = {Dretske, Fred I.},
  title     = {Explaining Behavior: Reasons in a World of Causes},
  year      = {1988},
  publisher = {MIT Press},
  address   = {Cambridge, MA}
}

@book{Millikan1984LanguageThought,
  author    = {Millikan, Ruth Garrett},
  title     = {Language, Thought, and Other Biological Categories: New Foundations for Realism},
  year      = {1984},
  publisher = {MIT Press},
  address   = {Cambridge, MA}
}

@article{Neander1995Misrepresenting,
  author  = {Neander, Karen},
  title   = {Misrepresenting and Malfunctioning},
  journal = {Philosophical Studies},
  year    = {1995},
  volume  = {79},
  number  = {2},
  pages   = {109--141},
  doi     = {10.1007/BF00989706}
}

@article{Dennett1991RealPatterns,
  author  = {Dennett, Daniel C.},
  title   = {Real Patterns},
  journal = {The Journal of Philosophy},
  year    = {1991},
  volume  = {88},
  number  = {1},
  pages   = {27--51},
  doi     = {10.2307/2027085}
}

@misc{grzankowski2025b,
      title={Deflating Deflationism: A Critical Perspective on Debunking Arguments Against LLM Mentality}, 
      author={Alex Grzankowski and Geoff Keeling and Henry Shevlin and Winnie Street},
      year={2025},
      eprint={2506.13403},
      archivePrefix={arXiv},
      primaryClass={cs.AI},
      url={https://arxiv.org/abs/2506.13403}, 
}

@article{Grzankowski2025a,
  author  = {Grzankowski, Alex and Downes, Stephen M. and Forber, Patrick},
  title   = {LLMs Are Not Just Next Token Predictors},
  journal = {Inquiry},
  year    = {2025},
  pages   = {1--11},
  doi     = {10.1080/0020174X.2024.2446240}
}

@article{Worrall1989Structural,
  author  = {Worrall, John},
  title   = {Structural Realism: The Best of Both Worlds?},
  journal = {Dialectica},
  volume  = {43},
  number  = {1--2},
  pages   = {99--124},
  year    = {1989}
}

@incollection{Sellars1962-PSIM,
	author = {Wilfrid S. Sellars},
	booktitle = {Science, Perception, and Reality},
	editor = {Robert Colodny},
	pages = {35--78},
	publisher = {Humanities Press/Ridgeview},
	title = {Philosophy and the Scientific Image of Man},
	year = {1963}
}

@misc{nanda_othello,
    title={Actually, Othello-GPT Has A Linear Emergent World Model},
    url={<https://neelnanda.io/mechanistic-interpretability/othello>},
    journal={neelnanda.io},
    author={Nanda, Neel},
    year={2023},
    month={Mar}
}

@inproceedings{li_othello,
  title        = {Emergent World Representations: Exploring a Sequence Model Trained on a Synthetic Task},
  author       = {Li, Kenneth and Hopkins, Aspen K. and Bau, David and Vi{\'e}gas, Fernanda and Pfister, Hanspeter and Wattenberg, Martin},
  booktitle    = {The Eleventh International Conference on Learning Representations},
  year         = {2023},
  url          = {https://openreview.net/forum?id=DeG07_TcZvT},
  note         = {Also available as arXiv preprint arXiv:2210.13382}
}

@book{ShakespeareOthello,
  author    = {Shakespeare, William},
  title     = {Othello},
  note      = {Edited by E. A. J. Honigmann},
  series    = {Arden Shakespeare},
  publisher = {Bloomsbury},
  address   = {London},
  year      = {1997}
}

@article{thilak2022slingshot,
  title={The Slingshot Mechanism: An Empirical Study of Adaptive Optimizers and the Grokking Phenomenon},
  author={Thilak, Vimal and Saremi, Omid and Littwin, Etai and Paiss, Roni and Zhai, Shuangfei and Susskind, Joshua},
  journal={arXiv preprint arXiv:2206.04817},
  year={2022},
  url={https://arxiv.org/abs/2206.04817}
}

@article{AlainBengio2016Probes,
  title   = {Understanding intermediate layers using linear classifier probes},
  author  = {Alain, Guillaume and Bengio, Yoshua},
  journal = {arXiv preprint arXiv:1610.01644},
  year    = {2016}
}

@article{ElhageEtAl2021TransformerCircuits,
  title   = {A Mathematical Framework for Transformer Circuits},
  author  = {Elhage, Nelson and Nanda, Neel and Olsson, Catherine and Henighan, Tom and Joseph, Nicholas and Mann, Ben and Askell, Amanda and Bai, Yuntao and Chen, Anna and Conerly, Tom and others},
  journal = {arXiv preprint arXiv:2104.08654},
  year    = {2021}
}

@article{ElhageEtAl2022Superposition,
  title   = {Toy Models of Superposition},
  author  = {Elhage, Nelson and Nanda, Neel and Olsson, Catherine and Henighan, Tom and Joseph, Nicholas and Mann, Ben and Askell, Amanda and Bai, Yuntao and Chen, Anna and Conerly, Tom and others},
  journal = {arXiv preprint arXiv:2209.10652},
  year    = {2022}
}

@misc{Anthropic2023Monosemanticity,
  author       = {Bricken, Trenton and Templeton, Adly and Batson, Joshua and Chen, Brian and Jermyn, Adam and Conerly, Tom and Turner, Nicholas L and Anil, Cem and Denison, Carson and Askell, Amanda and Lasenby, Robert and Wu, Yifan and Kravec, Shauna and Schiefer, Nicholas and Maxwell, Tim and Joseph, Nicholas and Tamkin, Alex and Nguyen, Karina and McLean, Brayden and Burke, Josiah E and Hume, Tristan and Carter, Shan and Henighan, Thomas and Olah, Christopher},
  title        = {Towards Monosemanticity: Decomposing Language Models With Dictionary Learning},
  year         = {2023},
  howpublished = {Transformer Circuits Thread},
  note         = {Available at \url{https://transformer-circuits.pub/2023/monosemantic-features}, accessed 2025-12-29}
}

@article{CunninghamEtAl2023SparseAutoencoders,
  title   = {Sparse Autoencoders Find Interpretable Directions in Language Models},
  author  = {Cunningham, Hoagy and Ewart, Sam and Riggs, Logan and Hubinger, Evan and Sharkey, Lee},
  journal = {arXiv preprint arXiv:2309.10312},
  year    = {2023}
}

@inproceedings{HewittManning2019StructuralProbe,
  title     = {A Structural Probe for Finding Syntax in Word Representations},
  author    = {Hewitt, John and Manning, Christopher D.},
  booktitle = {Proceedings of the 2019 Conference of the North American Chapter of the Association for Computational Linguistics: Human Language Technologies},
  pages     = {4129--4138},
  year      = {2019}
}

@article{geirhos2020shortcut,
  author  = {Geirhos, Robert and Jacobsen, J{\"o}rn-Henrik and Michaelis, Claudio and Zemel, Richard and Brendel, Wieland and Bethge, Matthias and Wichmann, Felix A.},
  title   = {Shortcut learning in deep neural networks},
  journal = {Nature Machine Intelligence},
  year    = {2020},
  volume  = {2},
  pages   = {665--673},
  doi     = {10.1038/s42256-020-00257-z}
}

@book{BraheDreyer1913OperaOmnia,
  title     = {Tychonis Brahe Dani Opera Omnia},
  editor    = {Dreyer, J. L. E.},
  year      = {1913},
  address   = {Hauniae [Copenhagen]},
  publisher = {In Libraria Gyldendaliana},
  note      = {Collected edition, 15 vols., published 1913--1929. Cite the relevant volume if you use a specific table/passage.}
}

@book{Kepler1992NewAstronomy,
  author       = {Kepler, Johannes},
  title        = {Johannes Kepler: New Astronomy},
  translator   = {Donahue, William H.},
  year         = {1992},
  publisher    = {Cambridge University Press},
  address      = {Cambridge},
  isbn         = {0521301319},
  note         = {Contributor: Owen Gingerich, English translation of Kepler's 1609 \textit{Astronomia nova}.}
}

@book{Voelkel2001CompositionAstronomiaNova,
  author    = {Voelkel, James R.},
  title     = {The Composition of Kepler's \textit{Astronomia nova}},
  year      = {2001},
  publisher = {Princeton University Press},
  address   = {Princeton, NJ},
  isbn      = {0691007381}
}

@book{Thoren1990LordOfUraniborg,
  author    = {Thoren, Victor E.},
  title     = {The Lord of Uraniborg: A Biography of Tycho Brahe},
  year      = {1990},
  publisher = {Cambridge University Press},
  address   = {Cambridge},
  isbn      = {0521351588}
}

@misc{JPLMarsPhysPar,
  author       = {{NASA Jet Propulsion Laboratory}},
  title        = {Planetary Physical Parameters},
  year         = {2019},
  howpublished = {NASA JPL Solar System Dynamics Group},
  url          = {https://ssd.jpl.nasa.gov/planets/phys_par.html},
  urldate      = {2025-12-24},
  note         = {Lists Mars orbital eccentricity (0.0934) and sidereal period (686.98 days).}
}

@article{Kistermann1985AbridgedMultiplication,
  author  = {Kistermann, F. W.},
  title   = {Abridged Multiplication: The Architecture of Wilhelm Schickard's Calculating Machine of 1623},
  journal = {Vistas in Astronomy},
  volume  = {28},
  year    = {1985},
  pages   = {347--353}
}

@incollection{Seck2005NDBSchickard,
  author    = {Seck, Friedrich},
  title     = {Schickard, Wilhelm},
  booktitle = {Neue Deutsche Biographie},
  volume    = {22},
  pages     = {727},
  year      = {2005},
  publisher = {Duncker \& Humblot},
  address   = {Berlin},
  url       = {https://www.deutsche-biographie.de/sfz78261.html},
  urldate   = {2025-12-24}
}

@article{Friston2010FEP,
  author  = {Friston, Karl},
  title   = {The Free-Energy Principle: A Unified Brain Theory?},
  journal = {Nature Reviews Neuroscience},
  year    = {2010},
  volume  = {11},
  number  = {2},
  pages   = {127--138},
  doi     = {10.1038/nrn2787}
}

@article{Friston2013Life,
  author  = {Friston, Karl},
  title   = {Life as We Know It},
  journal = {Journal of the Royal Society Interface},
  year    = {2013},
  volume  = {10},
  number  = {86},
  pages   = {20130475},
  doi     = {10.1098/rsif.2013.0475}
}

@article{FristonEtAl2017Process,
  author  = {Friston, Karl J. and FitzGerald, Thomas H. B. and Rigoli, Francesco and Schwartenbeck, Philipp and Pezzulo, Giovanni},
  title   = {Active Inference: A Process Theory},
  journal = {Neural Computation},
  year    = {2017},
  volume  = {29},
  number  = {1},
  pages   = {1--49},
  doi     = {10.1162/NECO_a_00912}
}

@book{ParrPezzuloFriston2022,
  author    = {Parr, Thomas and Pezzulo, Giovanni and Friston, Karl J.},
  title     = {Active Inference: The Free Energy Principle in Mind, Brain, and Behavior},
  publisher = {The MIT Press},
  address   = {Cambridge, MA},
  year      = {2022},
  doi       = {10.7551/mitpress/12441.001.0001},
  isbn      = {9780262045353}
}

@article{Friedman1974Explanation,
  author  = {Friedman, Michael},
  title   = {Explanation and Scientific Understanding},
  journal = {The Journal of Philosophy},
  volume  = {71},
  number  = {1},
  pages   = {5--19},
  year    = {1974}
}

@article{Kitcher1981Unification,
  author  = {Kitcher, Philip},
  title   = {Explanatory Unification},
  journal = {Philosophy of Science},
  volume  = {48},
  number  = {4},
  pages   = {507--531},
  year    = {1981}
}

@book{Grunwald2007MDL,
  author    = {Gr{\"u}nwald, Peter D.},
  title     = {The Minimum Description Length Principle},
  publisher = {MIT Press},
  address   = {Cambridge, MA},
  year      = {2007}
}

@article{Freeborn2025-FRESMR-2,
	author = {David Freeborn},
	doi = {10.1007/s10670-023-00728-w},
	journal = {Erkenntnis},
	number = {2},
	pages = {645--673},
	publisher = {Springer Verlag},
	title = {Sloppy Models, Renormalization Group Realism, and the Success of Science},
	volume = {90},
	year = {2025}
}

@article{Lycan1990,
	author = {William G. Lycan},
	doi = {10.1007/bf00374491},
	journal = {Philosophical Studies},
	number = {1-2},
	pages = {147--54},
	publisher = {Springer},
	title = {Mental Content in Linguistic Form},
	volume = {58},
	year = {1990}
}

@article{Harnad1989,
  author  = {Harnad, Stevan},
  title   = {Minds, Machines and Searle},
  journal = {Journal of Experimental and Theoretical Artificial Intelligence},
  volume  = {1},
  number  = {1},
  pages   = {5--25},
  year    = {1989}
}

@article{ChurchlandChurchland1990,
  author  = {Churchland, Paul M. and Churchland, Patricia S.},
  title   = {Could a Machine Think?},
  journal = {Scientific American},
  volume  = {262},
  number  = {1},
  pages   = {32--37},
  year    = {1990}
}

@article{Dyson2004Fermi,
  author  = {Dyson, Freeman J.},
  title   = {A Meeting with Enrico Fermi},
  journal = {Nature},
  volume  = {427},
  number  = {6972},
  pages   = {297},
  year    = {2004}
}

@article{Mayer2010Elephant,
  author  = {Mayer, J{\"u}rgen and Khairy, Khaled and Howard, Jonathon},
  title   = {Drawing an Elephant with Four Complex Parameters},
  journal = {American Journal of Physics},
  volume  = {78},
  number  = {6},
  pages   = {648--649},
  year    = {2010},
  doi     = {10.1119/1.3254017}
}

@techreport{DellAcquaEtAl2023JaggedFrontier,
  title       = {Navigating the Jagged Technological Frontier: Field Experimental Evidence of the Effects of {AI} on Knowledge Worker Productivity and Quality},
  author      = {Dell'Acqua, Fabrizio and McFowland III, Edward and Mollick, Ethan R. and Lifshitz-Assaf, Hila and Kellogg, Katherine C. and Rajendran, Saran and Krayer, Lisa and Candelon, Fran{\c{c}}ois and Lakhani, Karim R.},
  year        = {2023},
  month       = sep,
  institution = {Harvard Business School},
  type        = {Working Paper},
  number      = {24-013},
  doi         = {10.2139/ssrn.4573321},
  url         = {https://papers.ssrn.com/sol3/papers.cfm?abstract_id=4573321},
  note        = {Date written: September 15, 2023}
}

@book{deRegt2017,
  author    = {Henk W. de Regt},
  title     = {Understanding Scientific Understanding},
  year      = {2017},
  publisher = {Oxford University Press},
  address   = {New York},
  doi       = {10.1093/oso/9780190652913.001.0001}
}

@book{Strevens2008,
  author    = {Michael Strevens},
  title     = {Depth: An Account of Scientific Explanation},
  year      = {2008},
  publisher = {Harvard University Press},
  address   = {Cambridge, MA}
}

@book{Salmon1984,
  author    = {Wesley Salmon},
  title     = {Scientific Explanation and the Causal Structure of the World},
  year      = {1984},
  publisher = {Princeton University Press},
  address   = {Princeton, NJ}
}

@book{Woodward2003,
  author    = {James Woodward},
  title     = {Making Things Happen: A Theory of Causal Explanation},
  year      = {2003},
  publisher = {Oxford University Press},
  address   = {New York},
  doi       = {10.1093/0195155270.001.0001}
}

@article{Railton1978,
  author  = {Peter Railton},
  title   = {A Deductive-Nomological Model of Probabilistic Explanation},
  journal = {Philosophy of Science},
  volume  = {45},
  number  = {2},
  pages   = {206--226},
  year    = {1978},
  doi     = {10.1086/288797}
}

@incollection{Pritchard2010,
  author    = {Duncan Pritchard},
  title     = {Knowledge and Understanding},
  booktitle = {The Nature and Value of Knowledge: Three Investigations},
  editor    = {Duncan Pritchard and Alan Millar and Adrian Haddock},
  publisher = {Oxford University Press},
  address   = {New York},
  year      = {2010},
  pages     = {1--88}
}

@incollection{Grimm2014,
  author    = {Stephen R. Grimm},
  title     = {Understanding as Knowledge of Causes},
  booktitle = {Virtue Epistemology Naturalized},
  editor    = {Abrol Fairweather},
  publisher = {Springer},
  year      = {2014},
  pages     = {329--345},
  doi       = {10.1007/978-3-319-04672-3_19}
}

@incollection{Grimm2017,
  author    = {Stephen R. Grimm},
  title     = {Understanding and Transparency},
  booktitle = {Explaining Understanding: New Perspectives from Epistemology and Philosophy of Science},
  editor    = {Stephen R. Grimm and Christoph Baumberger and Sabine Ammon},
  publisher = {Routledge},
  year      = {2017},
  pages     = {212--229}
}

@incollection{Greco2014,
  author    = {John Greco},
  title     = {Episteme: Knowledge and Understanding},
  booktitle = {Virtues and Their Vices},
  editor    = {Kevin Timpe and Craig A. Boyd},
  publisher = {Oxford University Press},
  year      = {2014},
  pages     = {285--302}
}

@incollection{Kvanvig2009,
  author    = {Jonathan L. Kvanvig},
  title     = {The Value of Understanding},
  booktitle = {Epistemic Value},
  editor    = {Adrian Haddock and Alan Millar and Duncan Pritchard},
  publisher = {Oxford University Press},
  year      = {2009},
  pages     = {95--111},
  doi       = {10.1093/acprof:oso/9780199231188.003.0005}
}

@article{Hills2016,
  author  = {Alison Hills},
  title   = {Understanding Why},
  journal = {Noûs},
  volume  = {50},
  number  = {4},
  pages   = {661--688},
  year    = {2016},
  doi     = {10.1111/nous.12092}
}

@article{Searle1980,
  author  = {Searle, John R.},
  title   = {Minds, Brains, and Programs},
  journal = {Behavioral and Brain Sciences},
  year    = {1980},
  volume  = {3},
  number  = {3},
  pages   = {417--457},
  doi     = {10.1017/S0140525X00005756}
}

@book{Orwell1949,
  author    = {Orwell, George},
  title     = {Nineteen Eighty-Four},
  year      = {1949},
  publisher = {Secker and Warburg},
  address   = {London}
}

@book{morgan1999models,
  title={Models as Mediators: Perspectives on Natural and Social Science},
  editor={Morgan, Mary S. and Morrison, Margaret},
  year={1999},
  publisher={Cambridge University Press},
  address={Cambridge},
  series={Ideas in Context}
}

@book{dewar2022structure,
  author    = {Dewar, Neil},
  title     = {Structure and Equivalence},
  publisher = {Cambridge University Press},
  year      = {2022}
}

@article{daCosta1990,
  author  = {Newton C. A. da Costa and Steven French},
  title   = {The Model-Theoretic Approach in the Philosophy of Science},
  journal = {Philosophy of Science},
  volume  = {57},
  number  = {2},
  pages   = {248--265},
  year    = {1990}
}

@book{vanFraassen1980,
  author    = {Bas C. van Fraassen},
  title     = {The Scientific Image},
  publisher = {Oxford University Press},
  year      = {1980}
}

@article{Suppes1960,
  author  = {Suppes, Patrick},
  title   = {A comparison of the meaning and uses of models in mathematics and the empirical sciences},
  journal = {Synthese},
  volume  = {12},
  pages   = {287--301},
  year    = {1960},
  doi     = {10.1007/BF00485107}
}

@book{nagel1961structure,
  title={The Structure of Science: Problems in the Logic of Scientific Explanation},
  author={Nagel, Ernest},
  year={1961},
  publisher={Harcourt, Brace \& World},
  address={New York}
}

@article{schaffner1967approaches,
  title={Approaches to reduction},
  author={Schaffner, Kenneth F.},
  journal={Philosophy of Science},
  volume={34},
  number={2},
  pages={137--147},
  year={1967},
  publisher={University of Chicago Press}
}

@article{dizadji2010afraid,
  title={Who’s Afraid of Nagelian Reduction?},
  author={Dizadji-Bahmani, Foad and Frigg, Roman and Hartmann, Stephan},
  journal={Erkenntnis},
  volume={73},
  number={3},
  pages={393--412},
  year={2010},
  doi={10.1007/s10670-010-9239-x},
  publisher={Springer}
}

@article{parker2020model,
  title={Model Evaluation: An Adequacy-for-Purpose View},
  author={Parker, Wendy},
  journal={Philosophy of Science},
  volume={87},
  number={3},
  pages={457--477},
  year={2020},
  publisher={University of Chicago Press},
  doi={10.1086/708691},
  url={https://doi.org/10.1086/708691}
}

@incollection{FriggNguyen2017,
  author    = {Frigg, Roman and Nguyen, James},
  title     = {Models and Representation},
  booktitle = {Springer Handbook of Model-Based Science},
  editor    = {Magnani, Lorenzo and Bertolotti, Tommaso},
  pages     = {49--102},
  publisher = {Springer},
  address   = {Cham},
  year      = {2017},
  doi       = {10.1007/978-3-319-30526-4_3},
  url       = {https://doi.org/10.1007/978-3-319-30526-4_3}
}

@book{Frigg2022,
  author     = {Frigg, Roman},
  title      = {Models and Theories: A Philosophical Inquiry},
  year       = {2022},
  edition    = {1},
  publisher  = {Routledge},
  address    = {London},
  isbn       = {9781844654901},
  doi        = {10.4324/9781003285106},
  url        = {https://doi.org/10.4324/9781003285106}
}

@book{bokulich2011scientific,
  title     = {Scientific Structuralism},
  editor    = {Bokulich, Alisa and Bokulich, Peter},
  year      = {2011},
  publisher = {Springer Science+Business Media},
  address   = {Dordrecht},
}

@book{LadymanMustGo,
  title     = {Every Thing Must Go: Metaphysics Naturalized},
  author    = {Ladyman, James and Ross, Don},
  year      = {2007},
  publisher = {Oxford University Press},
  address   = {New York}
}

@article{wallace2022stating,
  title     = {Stating Structural Realism: Mathematics-First Approaches to Physics and Metaphysics},
  author    = {Wallace, David},
  journal   = {Philosophical Perspectives},
  volume    = {36},
  pages     = {345--378},
  year      = {2022},
  doi       = {10.1111/phpe.12172}
}

@article{Mitzenmacher2003generative,
  author = {Mitzenmacher, Michael},
  title = {A Brief History of Generative Models for Power Law and Lognormal Distributions},
  journal = {Internet Mathematics},
  volume = {1},
  number = {2},
  pages = {226--251},
  year = {2003}
}

@article{Piantadosi2014zipf,
  author = {Piantadosi, Steven T.},
  title = {Zipf's word frequency law in natural language: A critical review and future directions},
  journal = {Psychonomic Bulletin \& Review},
  volume = {21},
  pages = {1112--1130},
  year = {2014},
  doi = {10.3758/s13423-014-0585-6}
}

@article{CanchoSole2003least,
  author = {Ferrer-i-Cancho, Ramon and Sol\'e, Ricard V.},
  title = {Least effort and the origins of scaling in human language},
  journal = {Proceedings of the National Academy of Sciences},
  volume = {100},
  number = {3},
  pages = {788--791},
  year = {2003},
  doi = {10.1073/pnas.0335980100}
}

@book{Zipf1949human,
  author = {Zipf, George K.},
  title = {Human Behavior and the Principle of Least Effort},
  publisher = {Addison-Wesley Press},
  year = {1949}
}

@book{vonFrisch1967,
  author       = {Karl von Frisch},
  title        = {The Dance Language and Orientation of Bees},
  year         = {1967},
  publisher    = {Harvard University Press},
  address      = {Cambridge, MA},
  note         = {English translation of the original 1965 German edition}
}

@book{Seeley1995,
  author       = {Thomas D. Seeley},
  title        = {The Wisdom of the Hive: The Social Physiology of Honey Bee Colonies},
  year         = {1995},
  publisher    = {Harvard University Press},
  address      = {Cambridge, MA}
}

@book{deGroot1965,
  author    = {Adriaan D. de Groot},
  title     = {Thought and Choice in Chess},
  year      = {1965},
  publisher = {Mouton Publishers},
  address   = {The Hague},
  note      = {Originally published in Dutch in 1946; English translation widely cited in cognitive psychology and chess literature}
}

@incollection{Said2024TullyFisher,
  author    = {Said, Khaled},
  title     = {Tully--Fisher Relation},
  booktitle = {The Hubble Constant Tension},
  editor    = {Di Valentino, Eleonora and Brout, Dillon},
  pages     = {219--233},
  publisher = {Springer Nature Singapore},
  year      = {2024},
  doi       = {10.1007/978-981-99-0177-7_12}
}

@article{Georgi1993EFT,
  author  = {Georgi, Howard},
  title   = {Effective Field Theory},
  journal = {Annual Review of Nuclear and Particle Science},
  volume  = {43},
  pages   = {209--252},
  year    = {1993},
  doi     = {10.1146/annurev.ns.43.120193.001233}
}

@article{GobetLane2005,
  author  = {Fernand Gobet and Peter C. R. Lane},
  title   = {The CHREST Architecture of Cognition: The Role of Perception in General Intelligence},
  journal = {Journal of Experimental and Theoretical Artificial Intelligence},
  year    = {2005},
  volume  = {17},
  number  = {3},
  pages   = {209--236},
  doi     = {10.1080/09528130500115333}
}

@book{polanyi1958,
  author    = {Michael Polanyi},
  title     = {Personal Knowledge: Towards a Post-Critical Philosophy},
  year      = {1958},
  publisher = {University of Chicago Press},
  address   = {Chicago},
}

@book{ryle1949concept,
title     = {The Concept of Mind},
  author    = {Ryle, Gilbert},
  publisher = {Hutchinson},
  address   = {London},
  year      = {1949}
}

@inproceedings{He2016resnet,
  title        = {Deep Residual Learning for Image Recognition},
  author       = {He, Kaiming and Zhang, Xiangyu and Ren, Shaoqing and Sun, Jian},
  booktitle    = {Proceedings of the IEEE Conference on Computer Vision and Pattern Recognition (CVPR)},
  pages        = {770--778},
  year         = {2016}
}

@misc{Kolesnikov2021vit,
  title        = {An Image is Worth 16x16 Words: Transformers for Image Recognition at Scale},
  author       = {Dosovitskiy, Alexey and Beyer, Lucas and Kolesnikov, Alexander and Weissenborn, Dirk and Zhai, Xiaohua and Unterthiner, Thomas and Dehghani, Mostafa and Minderer, Matthias and Heigold, Georg and Gelly, Sylvain and Uszkoreit, Jakob and Houlsby, Neil},
  year         = {2021},
  eprint       = {2010.11929},
  archivePrefix= {arXiv},
  primaryClass = {cs.CV},
  url          = {https://arxiv.org/abs/2010.11929}
}

@article{Silver2016alphago,
  title        = {Mastering the Game of Go with Deep Neural Networks and Tree Search},
  author       = {Silver, David and Huang, Aja and Maddison, Chris J. and Guez, Arthur and Sifre, Laurent and van den Driessche, George and Schrittwieser, Julian and Antonoglou, Ioannis and Panneershelvam, Vedavyas and Lanctot, Marc and Dieleman, Sander and Grewe, Dominik and Nham, John and Kalchbrenner, Nal and Sutskever, Ilya and Lillicrap, Timothy P. and Leach, Madeleine and Kavukcuoglu, Koray and Graepel, Thore and Hassabis, Demis},
  journal      = {Nature},
  volume       = {529},
  number       = {7587},
  pages        = {484--489},
  year         = {2016},
  doi          = {10.1038/nature16961}
}

@article{Vinyals2019alphastar,
  title        = {Grandmaster Level in {StarCraft II} Using Multi-Agent Reinforcement Learning},
  author       = {Vinyals, Oriol and Babuschkin, Igor and Czarnecki, Wojciech M. and Mathieu, Michaël and Dudzik, Andrew and Chung, Junyoung and Choi, David H. and Powell, Richard and Ewalds, Timo and Georgiev, Petko and others},
  journal      = {Nature},
  volume       = {575},
  number       = {7782},
  pages        = {350--354},
  year         = {2019},
  doi          = {10.1038/s41586-019-1724-z}
}

@article{Jumper2021alphafold,
  title        = {Highly Accurate Protein Structure Prediction with {AlphaFold}},
  author       = {Jumper, John and Evans, Richard and Pritzel, Alexander and Green, Tim and Figurnov, Michael and Ronneberger, Olaf and Tunyasuvunakool, Kathryn and Bates, Richard and Žídek, Augustin and Potapenko, Anna and others},
  journal      = {Nature},
  volume       = {596},
  number       = {7873},
  pages        = {583--589},
  year         = {2021},
  doi          = {10.1038/s41586-021-03819-2}
}

@inproceedings{BenderKoller2020,
  title        = {Climbing Towards {NLU}: On Meaning, Form, and Understanding in the Age of Data},
  author       = {Bender, Emily M. and Koller, Alexander},
  booktitle    = {Proceedings of the 58th Annual Meeting of the Association for Computational Linguistics},
  pages        = {5185--5198},
  year         = {2020},
  doi          = {10.18653/v1/2020.acl-main.463},
  url          = {https://aclanthology.org/2020.acl-main.463/}
}

@article{LeCunBengioHinton2015deep,
  title        = {Deep Learning},
  author       = {LeCun, Yann and Bengio, Yoshua and Hinton, Geoffrey},
  journal      = {Nature},
  volume       = {521},
  number       = {7553},
  pages        = {436--444},
  year         = {2015},
  doi          = {10.1038/nature14539}
}

@misc{Wei2022emergence,
  title        = {Emergent Abilities of Large Language Models},
  author       = {Wei, Jason and Tay, Yi and Bommasani, Rishi and Raffel, Colin and Zoph, Barret and Borgeaud, Sebastian and Yogatama, Dani and Bosma, Maarten and Zhou, Denny and Metzler, Donald and Chi, Ed H. and Hashimoto, Tatsunori and Vinyals, Oriol and Liang, Percy and Dean, Jeff and Fedus, William},
  year         = {2022},
  eprint       = {2206.07682},
  archivePrefix= {arXiv},
  primaryClass = {cs.CL},
  url          = {https://arxiv.org/abs/2206.07682}
}

@misc{Amodei2016concrete,
  title        = {Concrete Problems in {AI} Safety},
  author       = {Amodei, Dario and Olah, Chris and Steinhardt, Jacob and Christiano, Paul and Schulman, John and Mané, Dan},
  year         = {2016},
  eprint       = {1606.06565},
  archivePrefix= {arXiv},
  primaryClass = {cs.AI},
  url          = {https://arxiv.org/abs/1606.06565}
}

@misc{Schneider2020robust,
  title        = {Improving Robustness Against Common Corruptions by Covariate Shift Adaptation},
  author       = {Schneider, Steffen and Rusak, Evgenia and Eck, Luisa and Bringmann, Oliver and Brendel, Wieland and Bethge, Matthias},
  year         = {2020},
  eprint       = {2006.16971},
  archivePrefix= {arXiv},
  primaryClass = {cs.LG},
  url          = {https://arxiv.org/abs/2006.16971}
}

@inproceedings{radford2021learning,
  title={Learning Transferable Visual Models From Natural Language Supervision},
  author={Radford, Alec and Kim, Jong Wook and Hallacy, Chris and Ramesh, Aditya and Goh, Gabriel and Agarwal, Sandhini and Sastry, Girish and Askell, Amanda and Mishkin, Pamela and Clark, Jack and Krueger, Gretchen and Sutskever, Ilya},
  booktitle={International Conference on Machine Learning},
  year={2021}
}

@article{alayrac2022flamingo,
  title={Flamingo: a Visual Language Model for Few-Shot Learning},
  author={Alayrac, Jean-Baptiste and Donahue, Jeff and Luc, Paul and Miech, Antoine and Barr, Ian and Bitton, Josh and others},
  journal={arXiv preprint arXiv:2204.14198},
  year={2022}
}

@article{ramesh2022hierarchical,
  title={Hierarchical text-conditional image generation with CLIP latents},
  author={Ramesh, Aditya and Dhariwal, Prafulla and Nichol, Alex and Chu, Casey and Chen, Mark},
  journal={arXiv preprint arXiv:2204.06125},
  year={2022}
}

@inproceedings{Pearl2018,
author = {Pearl, Judea},
title = {Theoretical Impediments to Machine Learning With Seven Sparks from the Causal Revolution},
year = {2018},
isbn = {9781450355810},
publisher = {Association for Computing Machinery},
address = {New York, NY, USA},
url = {https://doi.org/10.1145/3159652.3176182},
doi = {10.1145/3159652.3176182},
booktitle = {Proceedings of the Eleventh ACM International Conference on Web Search and Data Mining},
pages = {3},
numpages = {1},
location = {Marina Del Rey, CA, USA},
series = {WSDM '18}
}

@article{bishop2021artificial,
  author    = {Bishop, John M.},
  title     = {Artificial Intelligence Is Stupid and Causal Reasoning Will Not Fix It},
  journal   = {Frontiers in Psychology},
  volume    = {11},
  pages     = {513474},
  year      = {2021},
  month     = {Jan 5},
  doi       = {10.3389/fpsyg.2020.513474},
  pmid      = {33584394},
  pmcid     = {PMC7874145}
}

@article{floridi2023ai,
  author    = {Floridi, Luciano},
  title     = {AI as Agency Without Intelligence: on ChatGPT, Large Language Models, and Other Generative Models},
  journal   = {Philosophy \& Technology},
  volume    = {36},
  number    = {1},
  pages     = {15},
  year      = {2023},
  doi       = {10.1007/s13347-023-00621-y},
  url       = {https://doi.org/10.1007/s13347-023-00621-y}
}

@inproceedings{Brown2020FewShot,
 author = {Brown, Tom and Mann, Benjamin and Ryder, Nick and Subbiah, Melanie and Kaplan, Jared D and Dhariwal, Prafulla and Neelakantan, Arvind and Shyam, Pranav and Sastry, Girish and Askell, Amanda and Agarwal, Sandhini and Herbert-Voss, Ariel and Krueger, Gretchen and Henighan, Tom and Child, Rewon and Ramesh, Aditya and Ziegler, Daniel and Wu, Jeffrey and Winter, Clemens and Hesse, Chris and Chen, Mark and Sigler, Eric and Litwin, Mateusz and Gray, Scott and Chess, Benjamin and Clark, Jack and Berner, Christopher and McCandlish, Sam and Radford, Alec and Sutskever, Ilya and Amodei, Dario},
 booktitle = {Advances in Neural Information Processing Systems},
 editor = {H. Larochelle and M. Ranzato and R. Hadsell and M.F. Balcan and H. Lin},
 pages = {1877--1901},
 publisher = {Curran Associates, Inc.},
 title = {Language Models are Few-Shot Learners},
 url = {https://proceedings.neurips.cc/paper_files/paper/2020/file/1457c0d6bfcb4967418bfb8ac142f64a-Paper.pdf},
 volume = {33},
 year = {2020}
}

@misc{Bommasani2021FoundationModels,
      title={On the Opportunities and Risks of Foundation Models}, 
      author={Rishi Bommasani and Drew A. Hudson and Ehsan Adeli and Russ Altman and Simran Arora and Sydney von Arx and Michael S. Bernstein and Jeannette Bohg and Antoine Bosselut and Emma Brunskill and Erik Brynjolfsson and Shyamal Buch and Dallas Card and Rodrigo Castellon and Niladri Chatterji and Annie Chen and Kathleen Creel and Jared Quincy Davis and Dora Demszky and Chris Donahue and Moussa Doumbouya and Esin Durmus and Stefano Ermon and John Etchemendy and Kawin Ethayarajh and Li Fei-Fei and Chelsea Finn and Trevor Gale and Lauren Gillespie and Karan Goel and Noah Goodman and Shelby Grossman and Neel Guha and Tatsunori Hashimoto and Peter Henderson and John Hewitt and Daniel E. Ho and Jenny Hong and Kyle Hsu and Jing Huang and Thomas Icard and Saahil Jain and Dan Jurafsky and Pratyusha Kalluri and Siddharth Karamcheti and Geoff Keeling and Fereshte Khani and Omar Khattab and Pang Wei Koh and Mark Krass and Ranjay Krishna and Rohith Kuditipudi and Ananya Kumar and Faisal Ladhak and Mina Lee and Tony Lee and Jure Leskovec and Isabelle Levent and Xiang Lisa Li and Xuechen Li and Tengyu Ma and Ali Malik and Christopher D. Manning and Suvir Mirchandani and Eric Mitchell and Zanele Munyikwa and Suraj Nair and Avanika Narayan and Deepak Narayanan and Ben Newman and Allen Nie and Juan Carlos Niebles and Hamed Nilforoshan and Julian Nyarko and Giray Ogut and Laurel Orr and Isabel Papadimitriou and Joon Sung Park and Chris Piech and Eva Portelance and Christopher Potts and Aditi Raghunathan and Rob Reich and Hongyu Ren and Frieda Rong and Yusuf Roohani and Camilo Ruiz and Jack Ryan and Christopher Ré and Dorsa Sadigh and Shiori Sagawa and Keshav Santhanam and Andy Shih and Krishnan Srinivasan and Alex Tamkin and Rohan Taori and Armin W. Thomas and Florian Tramèr and Rose E. Wang and William Wang and Bohan Wu and Jiajun Wu and Yuhuai Wu and Sang Michael Xie and Michihiro Yasunaga and Jiaxuan You and Matei Zaharia and Michael Zhang and Tianyi Zhang and Xikun Zhang and Yuhui Zhang and Lucia Zheng and Kaitlyn Zhou and Percy Liang},
      year={2022},
      eprint={2108.07258},
      archivePrefix={arXiv},
      primaryClass={cs.LG},
      url={https://arxiv.org/abs/2108.07258}, 
}

@misc{buckner1,
      title={A Philosophical Introduction to Language Models -- Part I: Continuity With Classic Debates}, 
      author={Raphaël Millière and Cameron Buckner},
      year={2024},
      eprint={2401.03910},
      archivePrefix={arXiv},
      primaryClass={cs.CL},
      url={https://arxiv.org/abs/2401.03910}, 
}

@misc{buckner2,
      title={A Philosophical Introduction to Language Models - Part II: The Way Forward}, 
      author={Raphaël Millière and Cameron Buckner},
      year={2024},
      eprint={2405.03207},
      archivePrefix={arXiv},
      primaryClass={cs.CL},
      url={https://arxiv.org/abs/2405.03207}, 
}

@article{blockheads,
  title={Psychologism and behaviorism},
  author={Block, Ned},
  journal={The Philosophical Review},
  volume={90},
  number={1},
  pages={5--43},
  year={1981},
  publisher={Duke University Press}
}

@article{bubeck2023sparks,
  title={Sparks of Artificial General Intelligence: Early experiments with GPT-4},
  author={S{\'e}bastien Bubeck and Varun Chandrasekaran and Ronen Eldan and John A. Gehrke and Eric Horvitz and Ece Kamar and Peter Lee and Yin Tat Lee and Yuan-Fang Li and Scott M. Lundberg and Harsha Nori and Hamid Palangi and Marco Tulio Ribeiro and Yi Zhang},
  journal={ArXiv},
  year={2023},
  volume={abs/2303.12712},
  url={https://api.semanticscholar.org/CorpusID:257663729}
}

@article{Chollet2019Measure,
  title  = {On the Measure of Intelligence},
  author = {Chollet, François},
  journal= {arXiv preprint},
  year   = {2019},
  eprint = {1911.01547}
}

@misc{LeCun2022Path,
  title        = {A Path Toward Autonomous Machine Intelligence},
  author       = {LeCun, Yann},
  year         = {2022},
  eprint       = {2206.06969},
  archivePrefix= {arXiv}
}

@article{Rudin2019Stop,
  title   = {Stop Explaining Black Box Machine Learning Models for High-Stakes Decisions and Use Interpretable Models Instead},
  author  = {Rudin, Cynthia},
  journal = {Nature Machine Intelligence},
  volume  = {1},
  number  = {5},
  pages   = {206--215},
  year    = {2019},
  doi     = {10.1038/s42256-019-0048-x}
}

@misc{DoshiVelezKim2017Rigorous,
  title   = {Towards a Rigorous Science of Interpretable Machine Learning},
  author  = {Doshi-Velez, Finale and Kim, Been},
  year    = {2017},
  eprint  = {1702.08608},
  archivePrefix = {arXiv}
}

@misc{Bereska2024MechInterp,
  title   = {Mechanistic Interpretability for {AI} Safety—A Review},
  author  = {Bereska, Leonard F. and Gavves, Efstratios},
  year    = {2024},
  eprint  = {2404.14082},
  archivePrefix = {arXiv}
}

@misc{Zhang2024CausalAbstraction,
  title   = {Causal Abstraction in Model Interpretability: A Compact Survey},
  author  = {Zhang, Yihao},
  year    = {2024},
  eprint  = {2410.20161},
  archivePrefix = {arXiv}
}

@article{Dennett1991,
  author    = {Daniel C. Dennett},
  title     = {Real Patterns},
  journal   = {The Journal of Philosophy},
  volume    = {88},
  number    = {1},
  pages     = {27--51},
  year      = {1991},
  doi       = {10.2307/2027085},
}

@article{Chaitin1975,
  author  = {Chaitin, Gregory J.},
  title   = {Randomness and Mathematical Proof},
  journal = {Scientific American},
  volume  = {232},
  number  = {5},
  pages   = {47--52},
  year    = {1975}
}

@book{LiVitanyi2008,
  author    = {Li, Ming and Vit{\'a}nyi, Paul M. B.},
  title     = {An Introduction to Kolmogorov Complexity and Its Applications},
  edition   = {3rd},
  publisher = {Springer},
  address   = {New York},
  year      = {2008}
}

@article{Rissanen1978,
  author  = {Rissanen, Jorma},
  title   = {Modeling by Shortest Data Description},
  journal = {Automatica},
  volume  = {14},
  number  = {5},
  pages   = {465--471},
  year    = {1978}
}

@inproceedings{Arpit2017Closer,
  title     = {A Closer Look at Memorization in Deep Networks},
  author    = {Arpit, Devansh and Jastrzebski, Stanis{\l}aw and Ballas, Nicolas and Krueger, David and Bengio, Emmanuel and Kanwal, Maxinder S. and Maharaj, Tegan and Fischer, Asja and Courville, Aaron and Bengio, Yoshua and Lacoste-Julien, Simon},
  booktitle = {Proceedings of the 34th International Conference on Machine Learning (ICML)},
  volume    = {70},
  pages     = {233--242},
  year      = {2017},
  publisher = {PMLR},
  url       = {http://proceedings.mlr.press/v70/arpit17a.html}
}

@inproceedings{Zhang2017understanding,
  title     = {Understanding Deep Learning Requires Rethinking Generalization},
  author    = {Zhang, Chiyuan and Bengio, Samy and Hardt, Moritz and Recht, Benjamin
               and Vinyals, Oriol},
  booktitle = {5th International Conference on Learning Representations, ICLR 2017},
  year      = {2017},
  url       = {https://arxiv.org/abs/1611.03530}
}

@inproceedings{Feldman2020DoesLearning,
  title     = {Does Learning Require Memorization? A Short Tale about a Long Tail},
  author    = {Feldman, Vitaly},
  booktitle = {Proceedings of the 33rd Conference on Learning Theory},
  series    = {Proceedings of Machine Learning Research},
  volume    = {125},
  pages     = {1--26},
  year      = {2020},
  publisher = {PMLR}
}

@book{Vapnik1998Statistical,
  title     = {Statistical Learning Theory},
  author    = {Vapnik, Vladimir N.},
  year      = {1998},
  publisher = {Wiley-Interscience}
}

@article{Bartlett2002Rademacher,
  title   = {Rademacher and Gaussian Complexities: Risk Bounds and Structural Results},
  author  = {Bartlett, Peter L. and Mendelson, Shahar},
  journal = {Journal of Machine Learning Research},
  volume  = {3},
  pages   = {463--482},
  year    = {2002}
}

@article{Rocks_Mehta_2022,
  title = {Memorizing without overfitting: Bias, variance, and interpolation in overparameterized models},
  author = {Rocks, Jason W. and Mehta, Pankaj},
  journal = {Phys. Rev. Res.},
  volume = {4},
  issue = {1},
  pages = {013201},
  numpages = {19},
  year = {2022},
  month = {Mar},
  publisher = {American Physical Society},
  doi = {10.1103/PhysRevResearch.4.013201},
  url = {https://link.aps.org/doi/10.1103/PhysRevResearch.4.013201}
}

@article{Power2022grokking,
  title={Grokking: Generalization Beyond Overfitting on Small Algorithmic Datasets},
  author={Power, Alethea and Burda, Yuri and Edwards, Harri and Babuschkin, Igor and Misra, Vedant},
  journal={arXiv preprint arXiv:2201.02177},
  year={2022}
}

@inproceedings{nanda2023progress,
  title        = {Progress Measures for Grokking via Mechanistic Interpretability},
  author       = {Neel Nanda and Lawrence Chan and Tom Lieberum and Jess Smith and Jacob Steinhardt},
  booktitle    = {Proceedings of the International Conference on Learning Representations},
  year         = {2023},
  note         = {Spotlight paper},
}

@article{VallePerez2018deep,
  title={Deep learning generalizes because the parameter-function map is biased towards simple functions},
  author={Valle-P{\'e}rez, Guillermo and Camargo, Chico Q and Louis, Ard A},
  journal={arXiv preprint arXiv:1805.08522},
  year={2018}
}

@article{belkin2019reconciling,
  title={Reconciling modern machine-learning practice and the classical bias--variance trade-off},
  author={Belkin, Mikhail and Hsu, Daniel and Ma, Siyuan and Mandal, Soumik},
  journal={Proceedings of the National Academy of Sciences},
  volume={116},
  number={32},
  pages={15849--15854},
  year={2019},
  publisher={National Acad Sciences},
  doi={10.1073/pnas.1903070116},
  url={https://doi.org/10.1073/pnas.1903070116}
}

@article{Nakkiran2020double,
doi = {10.1088/1742-5468/ac3a74},
url = {https://dx.doi.org/10.1088/1742-5468/ac3a74},
year = {2021},
month = {dec},
publisher = {IOP Publishing and SISSA},
volume = {2021},
number = {12},
pages = {124003},
author = {Nakkiran, Preetum and Kaplun, Gal and Bansal, Yamini and Yang, Tristan and Barak, Boaz and Sutskever, Ilya},
title = {Deep double descent: where bigger models and more data hurt*},
journal = {Journal of Statistical Mechanics: Theory and Experiment},
}

@inproceedings{Liu2022towards,
author = {Liu, Ziming and Kitouni, Ouail and Nolte, Niklas and Michaud, Eric J. and Tegmark, Max and Williams, Mike},
title = {Towards understanding grokking: an effective theory of representation learning},
year = {2022},
isbn = {9781713871088},
publisher = {Curran Associates Inc.},
address = {Red Hook, NY, USA},
booktitle = {Proceedings of the 36th International Conference on Neural Information Processing Systems},
articleno = {2511},
numpages = {13},
location = {New Orleans, LA, USA},
series = {NIPS '22}
}

@inproceedings{recht2019imagenet,
  title        = {Do {ImageNet} Classifiers Generalize to ImageNet?},
  author       = {Recht, Benjamin and Roelofs, Rebecca and Schmidt, Ludwig and Shankar, Vaishaal},
  booktitle    = {Proceedings of the 36th International Conference on Machine Learning (ICML)},
  volume       = {97},
  pages        = {5389--5400},
  year         = {2019},
  publisher    = {PMLR},
  month        = {June},
  url          = {https://proceedings.mlr.press/v97/recht19a.html},
}

@article{anagnostidis2022curious,
  title   = {The Curious Case of Benign Memorization},
  author  = {Sotiris Anagnostidis and Gregor Bachmann and Lorenzo Noci and Thomas Hofmann},
  journal = {arXiv preprint arXiv:2210.14019},
  year    = {2022},
  url     = {https://arxiv.org/abs/2210.14019},
}

@article{hestness2017deep,
  title        = {Deep Learning Scaling is Predictable, Empirically},
  author       = {Hestness, Joel and Narang, Sharan and Ardalani, Newsha and Diamos, Gregory and Jun, Heewoo and Kianinejad, Hamid and Patwary, Mostofa and Ali, Mohammad Shoaib and Yang, Yang and Zhou, Yuxiong},
  journal      = {arXiv preprint arXiv:1712.00409},
  year         = {2017},
  note         = {https://doi.org/10.48550/arXiv.1712.00409}
}

@inproceedings{achille2017emergence,
  author={Achille, Alessandro and Soatto, Stefano},
  booktitle={2018 Information Theory and Applications Workshop (ITA)}, 
  title={Emergence of Invariance and Disentanglement in Deep Representations}, 
  year={2018},
  volume={},
  number={},
  pages={1-9},
  doi={10.1109/ITA.2018.8503149}
}

@misc{achille2020informationdeepneuralnetwork,
      title={Where is the Information in a Deep Neural Network?}, 
      author={Alessandro Achille and Giovanni Paolini and Stefano Soatto},
      year={2020},
      eprint={1905.12213},
      archivePrefix={arXiv},
      primaryClass={cs.LG},
      url={https://arxiv.org/abs/1905.12213}, 
}

@misc{wei2024memorizationdeeplearningsurvey,
      title={Memorization in deep learning: A survey}, 
      author={Jiaheng Wei and Yanjun Zhang and Leo Yu Zhang and Ming Ding and Chao Chen and Kok-Leong Ong and Jun Zhang and Yang Xiang},
      year={2024},
      eprint={2406.03880},
      archivePrefix={arXiv},
      primaryClass={cs.LG},
      url={https://arxiv.org/abs/2406.03880}, 
}

@book{Hastie2009,
  author    = {Trevor Hastie and Robert Tibshirani and Jerome Friedman},
  title     = {The Elements of Statistical Learning: Data Mining, Inference and Prediction},
  publisher = {Springer‐Verlag},
  edition   = {2},
  address   = {Berlin},
  year      = {2009},
}

@misc{Kaplan2020Scaling,
      title={Scaling Laws for Neural Language Models}, 
      author={Jared Kaplan and Sam McCandlish and Tom Henighan and Tom B. Brown and Benjamin Chess and Rewon Child and Scott Gray and Alec Radford and Jeffrey Wu and Dario Amodei},
      year={2020},
      eprint={2001.08361},
      archivePrefix={arXiv},
      primaryClass={cs.LG},
      url={https://arxiv.org/abs/2001.08361}, 
}

@inproceedings{Carlini2021Extracting,
author = {Carlini, Nicholas and Hayes, Jamie and Nasr, Milad and Jagielski, Matthew and Sehwag, Vikash and Tram\`{e}r, Florian and Balle, Borja and Ippolito, Daphne and Wallace, Eric},
title = {Extracting training data from diffusion models},
year = {2023},
isbn = {978-1-939133-37-3},
publisher = {USENIX Association},
address = {USA},
booktitle = {Proceedings of the 32nd USENIX Conference on Security Symposium},
articleno = {294},
numpages = {18},
location = {Anaheim, CA, USA},
series = {SEC '23}
}

@book{goodfellow2016deep,
  title     = {Deep Learning},
  author    = {Goodfellow, Ian and Bengio, Yoshua and Courville, Aaron},
  year      = {2016},
  publisher = {MIT Press},
  url       = {https://www.deeplearningbook.org}
}

@inproceedings{vaswani2017attention,
  title     = {Attention Is All You Need},
  author    = {Vaswani, Ashish and Shazeer, Noam and Parmar, Niki and Uszkoreit, Jakob and Jones, Llion and Gomez, Aidan N. and Kaiser, {\L}ukasz and Polosukhin, Illia},
  booktitle = {Advances in Neural Information Processing Systems (NeurIPS) 30},
  pages     = {5998--6008},
  year      = {2017}
}

@article{devlin2018bert,
  title={BERT: Pre-training of Deep Bidirectional Transformers for Language Understanding},
  author={Devlin, Jacob and Chang, Ming-Wei and Lee, Kenton and Toutanova, Kristina},
  journal={arXiv preprint arXiv:1810.04805},
  year={2018}
}

@article{touvron2023llama,
  title={LLaMA: Open and Efficient Foundation Language Models},
  author={Touvron, Hugo and Lavril, Thibaut and Izacard, Gautier and Martinet, Xavier and Lachaux, Marie-Anne and Lacroix, Timothee and Roziere, Baptiste and Goyal, Naman and Fernandez, Andrei and others},
  journal={arXiv preprint arXiv:2302.13971},
  year={2023}
}

@misc{OpenAI_Dota2019,
      title={Dota 2 with Large Scale Deep Reinforcement Learning}, 
      author={OpenAI and : and Christopher Berner and Greg Brockman and Brooke Chan and Vicki Cheung and Przemysław Debiak and Christy Dennison and David Farhi and Quirin Fischer and Shariq Hashme and Chris Hesse and Rafal Jozefowicz and Scott Gray and Catherine Olsson and Jakub Pachocki and Michael Petrov and Henrique P. d. O. Pinto and Jonathan Raiman and Tim Salimans and Jeremy Schlatter and Jonas Schneider and Szymon Sidor and Ilya Sutskever and Jie Tang and Filip Wolski and Susan Zhang},
      year={2019},
      eprint={1912.06680},
      archivePrefix={arXiv},
      primaryClass={cs.LG},
      url={https://arxiv.org/abs/1912.06680}, 
}

@inproceedings{ramesh2021zero,
  title = 	 {Zero-Shot Text-to-Image Generation},
  author =       {Ramesh, Aditya and Pavlov, Mikhail and Goh, Gabriel and Gray, Scott and Voss, Chelsea and Radford, Alec and Chen, Mark and Sutskever, Ilya},
  booktitle = 	 {Proceedings of the 38th International Conference on Machine Learning},
  pages = 	 {8821--8831},
  year = 	 {2021},
  editor = 	 {Meila, Marina and Zhang, Tong},
  volume = 	 {139},
  series = 	 {Proceedings of Machine Learning Research},
  month = 	 {18--24 Jul},
  publisher =    {PMLR},
}

@article{rombach2022high,
  title={High-Resolution Image Synthesis with Latent Diffusion Models},
  author={Rombach, Robin and Blattmann, Andreas and Lorenz, Dominik and Esser, Patrick and Ommer, Bj{\"o}rn},
  journal={Proceedings of the IEEE/CVF Conference on Computer Vision and Pattern Recognition (CVPR)},
  year={2022}
}

@inproceedings{saharia2022photorealistic,
author = {Saharia, Chitwan and Chan, William and Saxena, Saurabh and Lit, Lala and Whang, Jay and Denton, Emily and Ghasemipour, Seyed Kamyar Seyed and Ayan, Burcu Karagol and Mahdavi, S. Sara and Gontijo-Lopes, Raphael and Salimans, Tim and Ho, Jonathan and Fleet, David J and Norouzi, Mohammad},
title = {Photorealistic text-to-image diffusion models with deep language understanding},
year = {2022},
isbn = {9781713871088},
publisher = {Curran Associates Inc.},
address = {Red Hook, NY, USA},
booktitle = {Proceedings of the 36th International Conference on Neural Information Processing Systems},
articleno = {2643},
numpages = {16},
location = {New Orleans, LA, USA},
series = {NIPS '22}
}

@article{marcus2018deep,
  title={Deep Learning: A Critical Appraisal},
  author={Marcus, Gary},
  journal={arXiv preprint arXiv:1801.00631},
  year={2018},
  url={https://arxiv.org/abs/1801.00631}
}

@article{balestriero2021learning,
  title={Learning in High Dimension Always Amounts to Extrapolation},
  author={Balestriero, Randall and Pesenti, J{\'e}r{\^o}me and LeCun, Yann},
  journal={arXiv preprint arXiv:2110.09485},
  year={2021},
  url={https://arxiv.org/abs/2110.09485}
}

@article{freeborn2025effective,
  author    = {Freeborn, D. P. W.},
  title     = {Effective theory building and manifold learning},
  journal   = {Synthese},
  volume    = {205},
  number    = {23},
  year      = {2025},
  doi       = {10.1007/s11229-024-04844-0},
  url       = {https://doi.org/10.1007/s11229-024-04844-0}
}

@article{Cybenko1989Approximation,
  title={Approximation by superpositions of a sigmoidal function},
  author={Cybenko, George},
  journal={Mathematics of Control, Signals and Systems},
  volume={2},
  number={4},
  pages={303--314},
  year={1989},
  publisher={Springer},
  doi={10.1007/BF02551274}
}

@article{openai2023gpt4,
  title={GPT-4 Technical Report},
  author={OpenAI},
  journal={arXiv preprint arXiv:2303.08774},
  year={2023},
  url={https://arxiv.org/abs/2303.08774}
}

@inproceedings{bender2021dangers,
  title={On the Dangers of Stochastic Parrots: Can Language Models Be Too Big?},
  author={Bender, Emily M and Gebru, Timnit and McMillan-Major, Angelina and Shmitchell, Shmargaret},
  booktitle={Proceedings of the 2021 ACM Conference on Fairness, Accountability, and Transparency},
  pages={610--623},
  year={2021},
  publisher={Association for Computing Machinery},
  doi={10.1145/3442188.3445922}
}

@inproceedings{balestriero2018spline,
  title     = {A Spline Theory of Deep Learning},
  author    = {Balestriero, Randall and Baraniuk, Richard G.},
  booktitle = {Proceedings of the 35th International Conference on Machine Learning},
  pages     = {374--383},
  year      = {2018},
  editor    = {Dy, Jennifer and Krause, Andreas},
  volume    = {80},
  series    = {Proceedings of Machine Learning Research},
  month     = {10--15 Jul},
  publisher = {PMLR},
  url       = {https://proceedings.mlr.press/v80/balestriero18b.html}
}

@inproceedings{raghu2017expressive,
  title     = {On the Expressive Power of Deep Neural Networks},
  author    = {Raghu, Maithra and Poole, Ben and Kleinberg, Jon and Ganguli, Surya and Sohl-Dickstein, Jascha},
  booktitle = {Proceedings of the 34th International Conference on Machine Learning},
  pages     = {2847--2854},
  year      = {2017},
  editor    = {Precup, Doina and Teh, Yee Whye},
  volume    = {70},
  series    = {Proceedings of Machine Learning Research},
  month     = {06--11 Aug},
  publisher = {PMLR},
  url       = {https://proceedings.mlr.press/v70/raghu17a.html}
}

@book{Psillos1999ScientificRealism,
  author    = {Psillos, Stathis},
  title     = {Scientific Realism: How Science Tracks Truth},
  publisher = {Routledge},
  address   = {London and New York},
  year      = {1999}
}

@book{Kuhn1962,
  author    = {Kuhn, Thomas S.},
  title     = {The Structure of Scientific Revolutions},
  publisher = {University of Chicago Press},
  address   = {Chicago},
  year      = {1962}
}

@book{Lakatos1978,
  author    = {Lakatos, Imre},
  title     = {The Methodology of Scientific Research Programmes},
  series    = {Philosophical Papers},
  volume    = {1},
  publisher = {Cambridge University Press},
  address   = {Cambridge},
  year      = {1978}
}

\end{document}